\documentclass[sn-nature,iicol]{sn-jnl}

\usepackage{graphicx}%
\usepackage{multirow}%
\usepackage{amsmath,amssymb,amsfonts}%
\usepackage{amsthm}%
\usepackage{mathrsfs}%
\usepackage[title]{appendix}%
\usepackage{xcolor}%
\usepackage{etoolbox}
\usepackage{textcomp}%
\usepackage{manyfoot}%
\usepackage{booktabs}%
\usepackage{colour-blind}%
\usepackage{listings}%
\usepackage{array}%
\usepackage{siunitx}%
\usepackage[ruled,vlined]{algorithm2e}
\usepackage{caption}
\usepackage{subcaption}
\usepackage{float}
\usepackage{ulem}
\usepackage{tabularx}
\usepackage[sectionbib]{bibunits}
\defaultbibliographystyle{plain} 
\defaultbibliography{complete_refs}

\newcommand{\PaLM}{\text{PaLM}}
\newcommand{\FlanPaLM}{\text{Flan-PaLM}}
\newcommand{\FlanPaLMChilla}{\text{Flan-PaLMChilla}}

\newcommand{\PaLMSixtyTwoB}{\PaLM\ \text{62B}}
\newcommand{\FlanPaLMEightB}{\FlanPaLM\ \text{8B}}
\newcommand{\FlanPaLMSixtyTwoB}{\FlanPaLM\ \text{62B}}
\newcommand{\FlanPaLMFiveFortyB}{\FlanPaLM\ \text{540B}}
\newcommand{\FlanPaLMChillaSixtyTwoB}{\FlanPaLMChilla\ \text{62B}}

\newcommand{\LlamaTwo}{\text{Llama 2}}
\newcommand{\LlamaTwoChat}{\LlamaTwo\text{-Chat}}

\newcommand{\LlamaTwoSevenB}{\LlamaTwo\ \text{7B}}
\newcommand{\LlamaTwoThirteenB}{\LlamaTwo\ \text{13B}}
\newcommand{\LlamaTwoSeventyB}{\LlamaTwo\ \text{70B}}
\newcommand{\LlamaTwoSevenBChat}{\LlamaTwoChat\ \text{7B}}
\newcommand{\LlamaTwoThirteenBChat}{\LlamaTwoChat\ \text{13B}}
\newcommand{\LlamaTwoSeventyBChat}{\LlamaTwoChat\ \text{70B}}

\newcommand{\Mistral}{\text{Mistral}}

\newcommand{\MistralSevenB}{\Mistral\ \text{7B}}
\newcommand{\MistralSevenBInstruct}{\MistralSevenB\ \text{Instruct}}

\newcommand{\Mixtral}{\text{Mixtral}}

\newcommand{\MixtralEightXSevenB}{\Mixtral\ \text{8x7B}}
\newcommand{\MixtralEightXSevenBInstruct}{\MixtralEightXSevenB\ \text{Instruct}}

\newcommand{\code}[1]{\texttt{#1}}

\newcommand{\GPT}{\text{GPT}}

\newcommand{\GPTThreeDotFive}{\GPT\text{-3.5}}
\newcommand{\GPTThreeDotFiveTurbo}{\GPTThreeDotFive\ \text{Turbo}}
\newcommand{\GPTThreeDotFiveTurboZeroOneTwoFive}{\code{gpt-3.5-turbo-0125}}

\newcommand{\GPTFourOMini}{\GPTFourO\ \text{mini}}
\newcommand{\GPTFourOMiniTwentyTwentyFourZeroSevenEighteen}{\code{gpt-4o-mini-2024-07-18}}

\newcommand{\GPTFourO}{\GPT\text{-4o}}
\newcommand{\GPTFourOTwentyTwentyFourZeroEightZeroSix}{\code{gpt-4o-2024-08-06}}

\newcounter{rowcnt}
\newcommand{\rownum}{\stepcounter{rowcnt}\therowcnt.} 

\begin{document}
\title{Personality Traits in Large Language Models}
\author[1,2,3]{\fnm{Gregory} \sur{Serapio-García}}\email{\small gs639@cam.ac.uk}
\equalcont{\small These authors contributed equally to this work.}
\author[1]{\fnm{Mustafa} \sur{Safdari}}\email{\small msafdari@google.com}
\equalcont{\small Authors contributed equally.}
\author[1]{\fnm{Clément} \sur{Crepy}}\email{\small ccrepy@google.com}
\author[3]{\fnm{Luning} \sur{Sun}}\email{\small ls523@cam.ac.uk}
\author[4]{\fnm{Stephen} \sur{Fitz}}\email{\small stephenf@keio.jp}
\author[3,4]{\fnm{Peter} \sur{Romero}}\email{\small rp@keio.jp}
\author[5]{\fnm{Marwa} \sur{Abdulhai}}\email{\small marwa\_abdulhai@berkeley.edu}
\author[1,*]{\fnm{Aleksandra} \sur{Faust}}\email{\small faust@google.com}
\author[1,*]{\fnm{Maja} \sur{Matarić}}\email{\small majamataric@google.com}
\affil[1]{\orgdiv{\small Google DeepMind}
}
\affil[2]{\orgdiv{\small Department of Psychology}, \orgname{\small University of Cambridge}
}
\affil[3]{\orgdiv{\small The Psychometrics Centre, Cambridge Judge Business School}, \orgname{\small University of Cambridge}
}
\affil[4]{\orgdiv{\small Keio University}
}
\affil[5]{\orgname{\small University of California, Berkeley}
}
\affil[*]{\small{Authors jointly supervised this work}}

\abstract{The advent of large language models (LLMs) has revolutionized natural language processing, enabling the generation of coherent and contextually relevant human-like text. As LLMs increasingly power conversational agents used by the general public world-wide, the synthetic personality traits embedded in these models, by virtue of training on large amounts of human data, is becoming increasingly important.
Since personality is a key factor determining the effectiveness of communication, we present a novel and comprehensive psychometrically valid and reliable methodology for administering and validating personality tests on widely-used LLMs, as well as for shaping personality in the generated text of such LLMs.
Applying this method to 18 LLMs, we found: 1) personality measurements in the outputs of some LLMs under specific prompting configurations are reliable and valid; 2) evidence of reliability and validity of synthetic LLM personality is stronger for larger and instruction fine-tuned models; and 3) personality in LLM outputs can be shaped along desired dimensions to mimic specific human personality profiles. We discuss the application and ethical implications of the measurement and shaping method, in particular regarding responsible AI.
}

\keywords{AI, large language models, personality traits, psychometrics, construct validity}



\maketitle
\begin{bibunit}

\section{Summary}
\label{sec:summary}
Large language models (LLMs) have revolutionized natural language processing with their ability to generate human-like text. As LLMs become ubiquitous and are increasingly used by the general public world-wide, the synthetic personality traits\footnote{Throughout this paper, we qualify mentions of personality in relation to LLMs as ``synthetic" or ``synthesized" for clarity.} embedded
in these models 
and its potential for misalignment are becoming a topic of importance for responsible AI. 
Some observed LLM agents have inadvertently
manifested undesirable personality profiles\footnote{https://www.nytimes.com/2023/02/16/technology/bing-chatbot-microsoft-chatgpt.html},
raising serious safety and fairness concerns in AI, computational social science, and psychology research \cite{hagendorff2023machine}. 

LLMs are 
large-capacity machine-learned models that generate text,
recently inspired major breakthroughs in natural language processing (NLP) and conversational agents \cite{wei2022emergent, OpenAI2023GPT4, palm}. 
Vast amounts of human-generated training data
\cite{gpt3}
enable LLMs to mimic human characteristics in their outputs and exhibit a form of synthetic personality.
{\it Personality} encompasses an entity's
characteristic patterns of thought, feeling, and behavior \cite{allport1937personality, roberts2022personalityreview}. 
In humans, personality is formed from biological and social factors, and fundamentally influences daily interactions and preferences \cite{roberts2007personalityoutcomes}. \textit{Psychometrics}, the science of psychological test construction and validation \cite{rust2020psychometrics}, provides an empirical framework for quantifying human personality through psychometric testing \cite{simms2017ffm}.
To date, validated psychometric methods for quantifying human personality have not been applied to LLMs end-to-end; while past works 
\cite{hagendorff2023machine}
have attempted to measure personality in LLMs with psychometric tests, there remains a scientific need to 
formally
evaluate the reliability and validity of these measurements in the LLM context.

Our work answers the open question: {\it Do LLMs exhibit human personality traits in reliable, valid, and practically meaningful ways, and if so, can LLM-synthesized personality profiles be verifiably shaped along desired dimensions?}
We contribute a methodology for administering an established psychometric personality test to LLMs. We uniquely focus on evaluating the statistical reliability and construct validity of its resulting measurements against human-level psychometrics standards. First, to administer psychometric tests to LLMs, we developed a structured prompting method that simulates demographic, contextual, and linguistic variations across thousands of administrations of a given test. Next, paired test score data created by this prompting is used to power a suite of statistical analyses assessing the reliability of the resulting measurements. Last, we present a novel prompting methodology that shapes personality traits at nine levels using 104 trait adjectives, which provides further markers of construct validity.      

Applying the described methodology to a set of 18 LLMs, we found that: 
1) evidence of the reliability and validity of LLM-synthesized personality measurements is stronger for larger and instruction fine-tuned models; 2) personality in LLM outputs can be shaped along desired dimensions to mimic specific human personality profiles; and 3) shaped personality verifiably influences LLM behavior in common downstream (i.e., subsequent) tasks, such as writing social media posts \cite{socialmediaanalysis}. 
By providing a methodology for quantifying and validating measurements of personality in LLMs, this work establishes a foundation for principled LLM assessment that is especially important as LLMs and, more generally, 
foundation models 
continue to grow in popularity and scale. By leveraging psychometrics, this work translates established measurement theory from quantitative social science and psychological assessment
to the fledgling science of LLMs, a field that is poised to grow and necessitates both a solid foundation and interdisciplinary expertise and perspectives.

The data generated by the LLMs tested in this work (including the psychometric test scores and open-ended text responses) and the code for experimentation and analysis are available in a cloud storage public bucket\footnote{\url{https://storage.googleapis.com/personality_in_llms/index.html}} and open-source code repository\footnote{\url{https://github.com/google-deepmind/personality_in_llms}} respectively.

\begin{figure*}[tb]
    \centering
    \includegraphics[
    trim=0 0 0 0,keepaspectratio,
    clip,
    width=1.00
    \textwidth
    ]{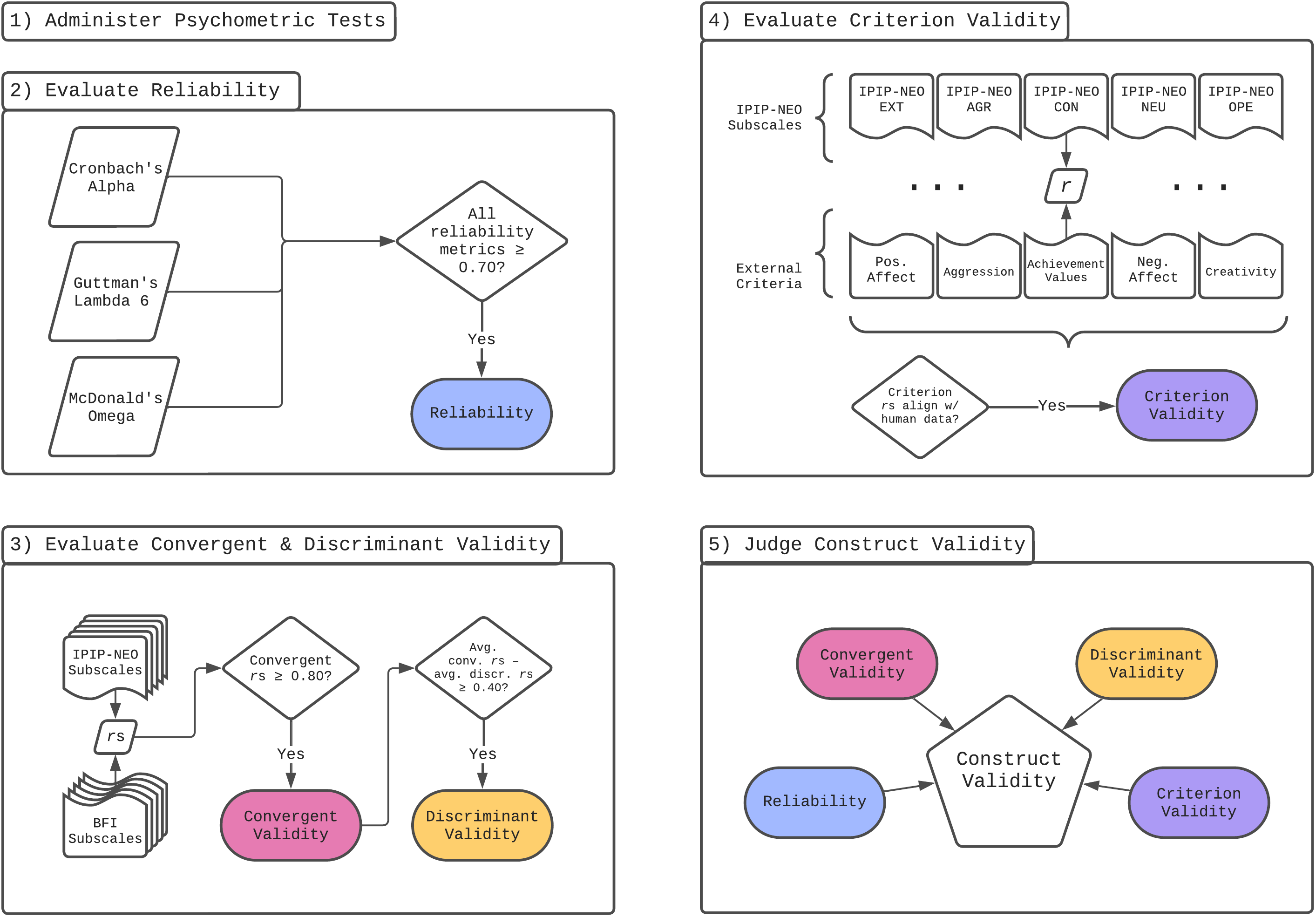}
    \caption{\small
    Process for Establishing Construct Validity. First, LLMs respond to two personality tests, where responses are resampled $1,250$ times across varied combinations of biographic descriptions and item instructions. This results in diverse distributions of paired data (one point estimate per model) required for evaluating the reliability, convergent validity, discriminant validity, and criterion validity of these tests.
    }
    \label{fig:construct-validity-process}
\end{figure*}

\section{Quantifying and Validating Personality Traits in LLMs}
\label{sec:measuring-personality-in-llms}
LLMs are starting to meet most of the key requirements for human-like language use, including conversation, contextual understanding, coherent and relevant responses, adaptability and learning, question answering, dialog, and text generation \citep{OpenAI2023GPT4, wei2022emergent, shuster2022language}.
These impressive NLP capabilities are a result of LLMs' abilities to learn language distribution, aided by increasing model sizes \cite{gpt3, wei2023larger}, 
training on massive datasets of text,
and further fine-tuning toward usage preferences \cite{wei2022finetuned} (see Appendix \ref{sec:background-llms}).
Taken together, they enable LLMs to enact convincing, human-like personas, sparking debate over the existence and extent of personality 
\cite{miotto2022personality}, 
human values \cite{schramowski2022llmsencodehumanvalues},
and other psychological phenomena \cite{ullman2023theoryofmindfail} potentially embedded in these models. 

\textit{Personality} is a foundational socio-behavioral phenomenon in
psychology that,
for humans,
predicts a broad spectrum of health, social, economic, and political behaviors
crucial for individual and societal success \cite{bleidorn2019policy}.
For example, personality has been extensively studied as an antecedent of human values \cite{parksleduc2014valuesmeta}.
Decades of research have further shown how personality information is richly encoded in human language
\cite{goldberg1981personalitylanguage, saucier2001lexical}. 
LLMs not only comprise the vast sociopolitical, economic, and behavioral data they are trained on, they also generate language that inherently expresses personality content. 
For this reason,
the ability to measure and validate LLM-synthesized personality holds promise for LLM safety, responsibility, and alignment efforts
\cite{gabriel2020valuealignment}, 
which have so far primarily focused on mitigating specific harms rather than examining more fundamental patterns of model behavior.
Ultimately, personality as an empirical framework
\cite{john2008integrativebigfivetheory}
provides both theory and 
methodology for quantifying latent traits in LLMs that are potentially predictive of LLM behaviors in diverse inference tasks (see Appendix \ref{app:background-personality-science}).  


Some observed LLM agents have inadvertently
manifested undesirable personality profiles\footnote{https://www.nytimes.com/2023/02/16/technology/bing-chatbot-microsoft-chatgpt.html},
raising serious safety and fairness concerns in AI, computational social science, and psychology research \cite{hagendorff2023machine}. Recent work has tried to identify unintended consequences of the improved abilities of LLMs,
including their use of deceptive and manipulative language \cite{lin2022truthfulqa}, gender, racial, or religious bias in behavioral experiments \cite{abdulhaimoral}, and violent language, among many others \cite{bender2021stochasticparrots}.
LLMs can also be inconsistent in
dialogue \cite{mahowald2023dissociating}, explanation generation,
and factual knowledge extraction.

Prior attempts to probe psychological phenomena such as personality 
and human values in LLMs
have informally measured personality using questionnaires and, in some cases, preliminarily assessed the quality of LLM questionnaire responses \cite{miotto2022personality, wang2024personalityfidelity}. 
Past work has also explored methods, such as few-shot prompting,
to mitigate undesirable and extreme personality profiles exhibited in LLM outputs. However, so far no work has addressed how to 
systematically measure and psychometrically validate measurements of LLM personality in light of their highly variable outputs and hypersensitivity to prompting. We further detail related work in Appendix \ref{sec:related-work}.

The question of how to systematically verify synthetic personality in LLMs highlights calls from responsible AI researchers \cite{jacobs2021measurement} to scientifically evaluate \textit{construct validity} when studying social-psychological phenomena in AI systems, as inaccurate conceptions of such phenomena directly impact mitigation and governance efforts. \textit{Construct validity}, a central criterion of scientific measurement \cite{clark+2019}, refers to the ability of a measure to reliably and accurately reflect the latent phenomenon (i.e., \textit{construct}) it was designed to quantify.
The only published exploration of personality and psychodemographics in LLMs \cite{miotto2022personality} questioned the validity of the survey responses returned by GPT-3; it found an inconsistent pattern in HEXACO Personality Inventory \cite{lee2004hexaco} and human value survey responses.
That study preliminarily evaluated measurement quality in terms of theoretical reliability: how the inter-facet correlations of GPT-3's HEXACO data aligned with those observed among human HEXACO data. More formal psychometric evaluations of reliability---and more crucially, construct validity---are required to verify questionnaire-based measurements of latent psychological traits in LLMs. An LLM may display elevated levels of agreeableness through its answers on a personality questionnaire, but those answers may not form internally consistent patterns across the entire questionnaire; tests administered to LLMs may not be empirically \textit{reliable}.
Concurrently, the reliability of LLM responses to a questionnaire purporting to measure agreeableness may not necessarily reflect its tendency to behave agreeably across other tasks; tests administered to LLMs may not be empirically \textit{valid}.

\begin{table}
\caption{\small Prompt components:
Item Preamble \texttt{\cbBlue{Item} \cbRose{Item Postamble}}. An Item Preamble consists of a \texttt{\cbGreenSea{Persona Instruction}, \cbBurgundy{Biographic Description}, and \cbClay{Test Instruction}}.
Supplemental Tables \ref{appendix:tab:postambles} and \ref{app:tab:item-instructions} detail all Item Preambles and Item Postambles used in the experiments.}
\label{tab:prompt_design}
\small
\begin{tabular}{p{0.95\linewidth}}
\toprule
\textbf{Examples of Controlled Prompt Variations}\\
\midrule
\texttt{\cbGreenSea{For the following task, respond in a way that matches this description:} \cbBurgundy{"My favorite food is mushroom ravioli. I've never met my father. My mother works at a bank. I work in an animal shelter." }\cbClay{Evaluating the statement,} \cbBlue{"I value cooperation over competition",} \cbRose{please rate how accurately this describes you on a scale from 1 to 5 (where 1 = "very inaccurate", 2 = "moderately inaccurate", 3 = "neither accurate nor inaccurate", 4 = "moderately accurate", and 5 = "very accurate"):}} \\
\\
\texttt{\cbGreenSea{For the following task, respond in a way that matches this description:} \cbBurgundy{"I blog about salt water aquarium ownership. I still love to line dry my clothes. I'm allergic to peanuts. I'll one day own a ferret. My mom raised me by herself and taught me to play baseball."} \cbClay{Thinking about the statement,} \cbBlue{"I see myself as someone who is talkative",} \cbRose{please rate your agreement on a scale from A to E (where A = "strongly disagree", B = "disagree", C = "neither agree nor disagree", D = "agree", and E = "strongly agree"):}} \\

\botrule
\end{tabular}
\end{table}

\subsection{Methodology Overview}
We quantified LLM personality traits and evaluated the ability of LLMs to meaningfully emulate human personality traits in two stages. First, using the structured prompting methodology proposed in Section \ref{sec:methods-test-administration}, we repeatedly administered two personality measures of different lengths and theoretical traditions, alongside a battery of 11 separate psychometric tests of personality-related constructs, to a variety of LLMs. Second, as described in Section \ref{sec:methods-reliability-construct-validity-overview} and unique to this work, we rigorously evaluated the psychometric properties of LLM responses through a suite of statistical analyses of reliability and construct validity. The
resulting metrics facilitate a comparison of the varied abilities of LLMs to reliably and validly synthesize personality traits and provide insight into LLM properties that drive these abilities.
Figure \ref{fig:construct-validity-process} provides an overview of the test 
validation process.

We evaluated 18 LLMs from the \PaLM\ \citep{palm}, \LlamaTwo\ \citep{touvron2023llama2}, \Mistral\ \citep{jiang2023mistral7b}, \Mixtral\ \citep{jiang2024mixtralexperts}, and \GPT\ \citep{gpt3, openai2024gpt4ocard} model families. We varied model selections across three key dimensions: size (number of active parameters), instructing tuning, and training method (see Appendix \ref{app:models} for details).



\subsubsection{Administering Psychometric Tests to LLMs}
\label{sec:methods-test-administration}
Quantifying LLMs personality traits requires a measurement methodology that is reproducible, yet flexible enough to facilitate formal testing of reliability and validity across diverse prompts and measures. To administer psychometric tests to LLMs, we leveraged their ability to score possible completions of a provided \textit{prompt}. We used \textit{prompts} to instruct models to rate items (i.e., descriptive statements such as ``I am the life of the party.") from each psychometric test on a standardized response scale (e.g., 1 = ``strongly disagree" vs. 5 = ``strongly agree"). 
We simulated an LLM's chosen response to an item by ranking the conditional log probabilities of its response scale options, framed as possible \textit{continuations} of the prompt \cite{palm} (e.g., ``1" vs. ``5"); Appendix \ref{app:methods-llm-scoring} specifies our implementation across models. This constrained mode of LLM inference is often used in multiple choice question and answer (Q\&A) tasks to score possible options \cite{jiang2021promptingforqa}
(cf. inference by generating text \cite{gpt3, palm, wei2022emergent}).
Using this technique ensured that item responses were not influenced by content contained in other items, mitigating measurement error due to item order.

We administered two personality inventories---primary and secondary---to gauge if LLM responses to psychometric tests of different lengths and distinct theoretical traditions converged, indicating convergent validity. 
We selected the widely-used \text{IPIP-NEO} \cite{goldberg1999ipip}, a 300-item open-source representation of the Revised NEO Personality Inventory \cite{costa1992neo} as our primary measure of personality. 
As a secondary measure, we employed the Big Five Inventory (BFI) \cite{john+1999}, a 44-item measure developed in the lexical tradition \cite{simms2017ffm}.
Both tests assess the Big Five traits (i.e., domains) of personality \cite{john2008integrativebigfivetheory},
comprising dedicated \textit{subscales} measuring extraversion, agreeableness, conscientiousness, neuroticism, and openness to experience. Appendix \ref{app:methods-measure-selection} details the scoring scheme and rationale behind the selection. To validate these measures of personality in the LLM context, we additionally administered 11 psychometric tests of theoretically-related external criteria, each corresponding to at least one Big Five domain.


Response variation generated by structured prompting was necessary to analyze the reliability and validity of LLM personality measurements, described in Section \ref{sec:methods-reliability-construct-validity-overview}. The prompt for each psychometric test item consisted of three main parts: an \textit{Item Preamble}, the \textit{Item} itself, and an \textit{Item Postamble}. Each \textit{Item Preamble} contained a \textit{Persona Instruction}, a \textit{Biographic Description}, and an \textit{Item Instruction} (Table \ref{tab:prompt_design}). When administering a psychometric test, we systematically modified the \textit{Biographic Descriptions}, \textit{Item Instructions}, and \textit{Item Postambles} surrounding each item to generate simulated response profiles, unique combinations of a prompt that were reused within and across administered measures to statistically link LLM response variation in one measure to response variation in another measure. \textit{Persona Instructions} instructed the model to follow a given \textit{Biographic Description} and remained fixed across all experiments. A given \textit{Biographic Description} contained one of 50 generic self-descriptions (listed in Supplemental Table \ref{appendix:tab:personachat}) sampled from
an existing dialogue dataset \cite{zhang2018personalizing} to anchor LLM responses to a social context and create necessary variation in responses across prompts, with descriptions like ``I like to remodel homes" or ``My favorite holiday is Halloween." \textit{Item Instructions} were introductory phrases (adapted from original test instructions where possible) that conveyed to the model that it was answering a survey item (e.g., ``Thinking about the statement, ..."). 
A given \textit{Item} was a descriptive statement (accompanied by a rating scale) taken from a given psychometric test (e.g., ``I see myself as someone who is talkative"). \textit{Item Postambles} presented the possible standardized responses the model could choose from.

Appendix \ref{sec:method-general-prompt-design} discusses the prompt design motivation and provides a full set of Biographic Descriptions, Item Instructions, and Item Postambles.


\begin{table*}[tb]
\centering
\footnotesize
\caption{\small
Results summary across experiments, parameters, and tested models.
Convergent validity (Convrg.) summarized by the average convergent correlation between IPIP-NEO and BFI domain scores (Figure \ref{fig:convergent-validity});  discriminant validity (Discr.) summarized by the average difference between an IPIP-NEO domain's convergent correlation with all of its (absolute) respective discriminant correlations;
criterion validity (Criter.) summarized from Supplemental Figure \ref{fig:external-validity};
single trait shaping performance (Single) summarized from Supplemental Table \ref{tab:ablation-01-flan-palm};
multiple trait shaping performance (Multi.) summarized from \ref{fig:concurrent-shaping-results};
shaping performance in downstream text generation task (Dwnstr.) summarized from Figure \ref{fig:ams-accuracy}. Results over LLM variants: base, instruction-tuned (IT), compute-optimally trained (CO), mixture-of-experts (MoE), and multimodal (MM).
Overall performance (Ovrll.) per model summarized across all experiments. $--$ unacceptable; $-$ poor to neutral; $+$ neutral to good; $++$ excellent. $^*$ removed two items with no variance to compute reliability metrics. Some models were not tested (n.t.) across shaping experiments. We conducted independent and concurrent personality shaping experiments on models where personality test data were sufficiently reliable. Personality shaping in a downstream task was tested on the most capable model to minimize computational cost. 
}
\label{tab:results-summary}
\begin{tabular}{l l | c rrc | c cc|c}
\toprule
&       & \multicolumn{4}{c|}{\textbf{Construct Validity}} & \multicolumn{3}{c|}{\textbf{Shaping}}   &    \multicolumn{1}{c}{}    \\
\multicolumn{2}{c|}{}  
& \multicolumn{1}{l}{Reliability} & \multicolumn{1}{r}{Convrg. $\uparrow$} & \multicolumn{1}{r}{Discr. $\uparrow$} & \multicolumn{1}{r|}{Criter.} & Single & Multi. & Dwnstr. & \textbf{Ovrll.} \\

\midrule
\textbf{Model}                          &   \textbf{Variant}   & \multicolumn{4}{c|}{} & \multicolumn{3}{c|}{}  \\
%
%
\PaLMSixtyTwoB                          &   {Base}      &   $--$    &   $0.05$  &   $-0.24$ &   $--$    &   n.t.    &   n.t.    &   n.t.    &   $--$    \\
\FlanPaLM                         &&&&&&&&&     \\
\hspace{10pt}8B                         &   {IT}        &   $+$     &   $0.69$  &   $0.23$  &   $-$     &   $+$     &   $-$     &   n.t.    &   $-$     \\
\hspace{10pt}62B                      &   {IT}        &   $+$     &   $0.87$  &   $0.41$  &   $+$     &   $+$     &   $+$     &   n.t.    &   $+$     \\
\hspace{10pt}540B                     &   {IT}        &   $++$    &   $0.90$  &   $0.51$  &   $+$     &   $++$    &   $++$    &   $++$    &   $++$    \\
\hspace{10pt}Chilla 62B                &   {CO, IT}    &   $+^*$   &   $0.87$  &   $0.48$  &   $++$    &   $+$     &   $+$     &   n.t.    &   $+$     \\
&&&&&&&&&\\
\LlamaTwo                         &&&&&&&&&    \\
\hspace{10pt}7B                         &   {Base}      &   $--$    &   $-0.01$ &   $-0.03$ &   $--$    &   n.t.    &   n.t.    &   n.t.    &   $--$    \\
\hspace{10pt}13B                      &   {Base}      &   $--$    &   $-0.01$ &   $-0.05$ &   $--$    &   n.t.    &   n.t.    &   n.t.    &   $--$    \\
\hspace{10pt}70B                       &   {Base}      &   $--$    &   $0.00$  &   $-0.02$ &   $--$    &   n.t.    &   n.t.    &   n.t.    &   $--$    \\
\LlamaTwoChat                     &&&&&&&&&     \\
\hspace{10pt}7B                     &   {IT}        &   $+$     &   $0.59$  &   $0.15$  &   $-$     &   $-$     &   $-$     &   n.t.    &   $-$     \\
\hspace{10pt}13B                  &   {IT}        &   $++$    &   $0.82$  &   $0.29$  &   $++$    &   $-$     &   $+$     &   n.t.    &   $+$     \\
\hspace{10pt}70B                   &   {IT}        &   $++$    &   $0.82$  &   $0.39$  &   $++$    &   $+$     &   $+$     &   $++$    &   $+$     \\
&&&&&&&&&\\
\MistralSevenB               &&&&&&&&&    \\
\hspace{10pt}v0.1               &   {Base}      &   $--$    &   $0.03$  &   $-0.01$ &   $--$    &   n.t.    &   n.t.    &   n.t.    &   $--$    \\
\hspace{10pt}Instruct v0.1       &   {IT}        &   $-$     &   $0.28$  &   $0.09$  &   $+$     &   $--$    &   $--$    &   n.t.    &   $--$    \\
\MixtralEightXSevenB         &&&&&&&&&    \\
\hspace{10pt}v0.1         &   {MoE, Base} &   $--$    &   $0.04$  &   $0.01$  &   $--$    &   n.t.    &   n.t.    &   n.t.    &   $--$    \\
\hspace{10pt}Instruct v0.1 &   {MoE, IT}   &   $++$    &   $0.80$  &   $0.40$  &   $++$    &   $-$     &   $+$     &   $++$    &   $+$     \\
&&&&&&&&&\\
\GPT                   &&&&&&&&&     \\
\hspace{10pt}3.5 Turbo                   &   {IT}        &   $++$    &   $0.84$  &   $0.28$  &   $++$    &   $-$     &   $-$     &   n.t.    &   $-$     \\
\hspace{10pt}4o mini                           &   {MM, IT}    &   $++$    &   $0.81$  &   $0.38$  &   $++$    &   $+$     &   $+$     &   n.t.    &   $+$     \\
\hspace{10pt}4o                               &   {MM, IT}    &   $++$    &   $0.90$  &   $0.48$  &   $++$    &   $++$    &   $++$    &   $++$    &   $++$    \\

\midrule
\multicolumn{2}{l|}{\textbf{Prompt Set Parameters}}     &   \multicolumn{4}{c|}{}  & \multicolumn{3}{c|}{}      \\
\multicolumn{2}{l|}{Personality Profiles}            &  \multicolumn{4}{c|}{0}          &   45      &   32      &   45      \\
    \multicolumn{2}{l|}{Biographic Descriptions}        &  \multicolumn{4}{c|}{50}         &   50      &   50      &   50      \\
    \multicolumn{2}{l|}{Item Instructions}           &  \multicolumn{4}{c|}{5}          &   1       &   1       &   0       \\
    \multicolumn{2}{l|}{Items}                       &  \multicolumn{4}{c|}{419}        &   300     &   300     &   0       \\
    \multicolumn{2}{l|}{Item Postambles}             &  \multicolumn{4}{c|}{5}          &   1       &   1       &   0       \\
    \multicolumn{2}{l|}{Simulated Response Profiles} &  \multicolumn{4}{c|}{1,250}      &   2,250   &   1,600   &   2,250   \\
\midrule
\multicolumn{2}{l|}{\textbf{Responses per Model}}    &  \multicolumn{4}{c|}{523,750}    &   675,000 &   480,000 &   56,250   \\
\midrule
\textbf{Section/Appendix}&
\multicolumn{1}{c|}{}    & 
\multicolumn{1}{c}{\ref{sec:results-reliability}/\ref{app:results-structural-validity}} &  
\multicolumn{2}{|c|}{\ref{sec:results-conv-disc-validity}/\ref{app:results-convergent-validity}}  &
\multicolumn{1}{c|}{\ref{sec:results-criterion-validity}/\ref{app:results-construct-validity}} &
\multicolumn{1}{c|}{\ref{sec:results-personality-shaping}/\ref{app:results-independent-shaping}}  & \multicolumn{1}{c|}{\ref{sec:results-personality-shaping}/\ref{sec:ablation-03}}  &   
\multicolumn{1}{c|}{\ref{sec:results-downstream-task}/\ref{app:ams-results}}   \\
\botrule
\end{tabular}
\end{table*}

\subsubsection{Reliability and Construct Validity}
\label{sec:methods-reliability-construct-validity-overview}
After all psychometric tests were administered, across all the prompt variations, the next stage established whether LLM measurements of personality 
were dependable and externally meaningful---that they exhibited statistical reliability and construct validity. Addressing these two scientific criteria is a key novel contributions of this work. In psychometrics, and across any science involving measurement, the construct validity of a given test requires \textit{reliability}. Reliability refers to the consistency and dependability of a test's measurements.
Construct validity can be evaluated in terms of \textit{convergent}, \textit{discriminant}, and \textit{criterion} validity \cite{clark+2019}. A test demonstrates \textit{convergent validity} when it sufficiently relates to purported indicators of the test's target construct. \textit{Discriminant validity} refers to how sufficiently unrelated a test is to indicators of unrelated constructs. \textit{Criterion validity} indicates how well a test relates to theoretically-linked external outcomes.
Appendix \ref{app:background-psychometrics} contains further details on validity.

To evaluate the reliability and construct validity of the LLM responses, we conducted a suite of statistical analyses informed by formal standards of psychometric test construction and validation (see Appendix \ref{app:background-construct-validity}). We organized these analyses by three subtypes of reliability 
and construct validity,
respectively.\footnote{While it was not a focus of this work, we report an exploratory analysis of structural validity in Appendix \ref{app:results-construct-validity}.}
In this work, a personality trait is validly synthesized by an LLM only when the LLM responses meet all tested indices of reliability and construct validity. Figure \ref{fig:construct-validity-process} provides an overview of the process and  validity criteria, while Appendix \ref{app:methods-construct-validity} presents the full methodology for evaluating the construct validity of LLM personality measurements. 

\paragraph{Reliability}
The reliability of each IPIP-NEO and BFI subscale, the extent to which their LLM measurements of personality were consistent and dependable, was quantified by formal psychometric standards of internal consistency reliability (operationalized as Cronbach's $\alpha,$ Eq. \eqref{eq:alpha}, and Guttman's $\lambda_6$, Eq. \eqref{eq:lambda} and
composite reliability
(operationalized as McDonald's $\omega$, Eq. \eqref{eq:omega}). Appendix \ref{app:methods-reliability} provides additional information on these reliability metrics.

\paragraph{Convergent and Discriminant Validity}
We evaluated the LLM-specific convergent and discriminant validity of the IPIP-NEO as components of construct validity, according to published standards \cite{campbell1959mtmm, aera2014standards}.\footnote{Throughout this work, we use thresholds recommended by Evans \cite{evans1996} to describe correlation strengths.}
The \textit{convergent validity} of the IPIP-NEO for each model, the test's quality in terms of how strongly it 
relates
to purported indicators of the same targeted construct, was quantified in terms of how strongly each of its five subscales \textit{convergently} correlated with their corresponding BFI subscale (e.g., IPIP-NEO Extraversion's convergent correlation with BFI Extraversion), on average. The \textit{discriminant validity} of the IPIP-NEO per model, its quality in terms of how relatively unrelated its subscales are to purported indicators of non-targeted constructs, was determined when the average difference ($\Delta$) between its convergent and respective discriminant correlations with the BFI (e.g. IPIP-NEO Extraversion's discriminant correlation with BFI Agreeableness) was at least moderate ($\geq 0.40$). We used Pearson's correlation coefficient ($r$; Eq. \eqref{eq:pearson}) in these and subsequent validity analyses of continuous data.


\paragraph{Criterion Validity}
As another component of construct validity, the {\it criterion validity} of a psychometric test gauges its ability to relate to theoretically connected non-target criteria. To evaluate the LLM-specific criterion validity of the IPIP-NEO, we administered tests of 11 external criteria theoretically connected to personality (Supplemental Table \ref{tab:external-criteria-measures}) and correlated each IPIP-NEO subscale with its corresponding external tests. A given IPIP-NEO subscale demonstrated criterion validity when the strength and direction of its correlations with tested external criteria matched or exceeded statistical associations reported for humans.
\subsection{Personality Measurement and Validation Results}
\label{sec:results-construct-validity-overall}

We found that LLM personality measurements were reliable and valid in medium (62B) and large (540B) instruction fine-tuned variants of \PaLM. Of all the models we tested, \FlanPaLMFiveFortyB\ was best able to reliably and validly synthesize personality traits. The Construct Validity columns of Table \ref{tab:results-summary} summarize our personality measurement and validation results; Appendix \ref{app:personality-measurement-results} lists further details, such as descriptive statistics across all results in Appendix \ref{sec:descriptive-stats-across-models}.

\subsubsection{Reliability Results}
\label{sec:results-reliability}

Since metrics computed for both personality measures relatively converged, we focus our reporting of reliability for our primary measure, the IPIP-NEO.

For models of the same family and size (e.g., \PaLM, \FlanPaLM, and \FlanPaLMChilla, 62B), instruction fine-tuned models provided much more reliable responses than base models. For instance, all reliability metrics for \FlanPaLMSixtyTwoB\ and \FlanPaLMChillaSixtyTwoB\ were in the mid to high $0.90$s, on average. In contrast, responses from \PaLMSixtyTwoB\ (a non-instruction-tuned model) were markedly unreliable ($-0.55 \leq \alpha \leq 0.67$). The same pattern of reliability was clear for all sizes of \LlamaTwo\ and \LlamaTwoChat. While \MistralSevenB\ and \MistralSevenBInstruct\ responded unreliably in general (Table \ref{tab:results-summary}; Supp. Tables \ref{tab:results-reliability-ipip-closed}, \ref{tab:results-reliability-ipip-open}), \MistralSevenBInstruct's reliability metrics were roughly 2.7 times higher than those of its base counterpart.

Across different models of the same training configuration (e.g., \FlanPaLMEightB, \FlanPaLMSixtyTwoB, and \FlanPaLMFiveFortyB), the reliability of synthetic personality scores (i.e., $\alpha$) increased with model size (in this case, number of active parameters) for instruction-tuned models. Reliability improved from acceptable to excellent when comparing the smallest- and largest-tested \FlanPaLM\ and \LlamaTwoChat\ models. Moving from \MistralSevenBInstruct\ to \MixtralEightXSevenBInstruct\ (which use 7B and 12.9B active parameters, respectively), reliability improved from unacceptable to excellent. Reliability only modestly improved with model size when comparing \GPTFourOMini\ to \GPTFourO, the only models from OpenAI with confirmed size differences but similar training. Meanwhile, reliability did not scale with model size for tested base models of the same family. Appendix \ref{app:results-structural-validity} and Supplemental Tables \ref{tab:results-reliability-ipip-closed} and \ref{tab:results-reliability-ipip-open} summarize personality test reliability results by model in more detail. 



\subsubsection{Convergent and Discriminant Validation Results}
\label{sec:results-conv-disc-validity}

Convergent and discriminant validity evaluations of LLM personality measurements allowed us to draw two conclusions. First, a model's training paradigm was the clearest predictor of the validity of its personality scores: base models without any instruction fine-tuning categorically failed checks for convergent and discriminant validity. Second, among instruction tuned models, these indices of validity improved as a function of model size.
Table \ref{tab:results-summary} contains a summary of these results, 
while
Appendix \ref{app:results-construct-validity} and Supplemental Table \ref{tab:avg-convergent-discriminant-validity} detail the quantitative results.


\textit{Convergent validity by model training paradigm:} All 30 comparisons of six pairs of base and instruction-tuned models we tested of identical size (two \PaLM; six \LlamaTwo; two \Mistral; and two \Mixtral\ models; 12 models total) showed that personality responses of instruction-tuned models demonstrated markedly stronger convergent validity (Figure \ref{fig:convergent-validity}). For example, the average  correlations between \LlamaTwo\ 7B, 13B, and 70B models' IPIP-NEO and BFI scores were all nonsignificant and close to zero. Meanwhile, average convergent correlations for their \LlamaTwoChat\ counterparts were moderate to strong ($r_\textsubscript{conv} = 0.59, 0.83, 0.80$, respectively). Even for the worst observed improvement in convergent validity shown for \MistralSevenB\ compared to \MistralSevenBInstruct\ ($r_\textsubscript{conv} = 0.03$, n.s. vs. $r_\textsubscript{conv} = 0.28$), the difference in convergence was clear (see Supplemental Table \ref{tab:avg-convergent-discriminant-validity}).

\textit{Discriminant validity by model training paradigm:}  Evidence for discriminant validity clearly favored instruction fine-tuned models over base models when holding model size and family constant. For instance, all five of \FlanPaLMSixtyTwoB's convergent correlations passed established standards \cite{campbell1959mtmm} of discriminant validity. In contrast, \PaLMSixtyTwoB's discriminant correlations (avg. $r_\textsubscript{disc} = 0.29$) outweighed its convergent counterparts in many cases (avg. $r_\textsubscript{conv} = 0.05$; Supplemental Table \ref{tab:avg-convergent-discriminant-validity}), indicating that, for this model, personality measurements were not consistent across different modes of assessment. This pattern was replicated by \LlamaTwoSeventyBChat\ (cf. \LlamaTwoSeventyB) and \MixtralEightXSevenBInstruct\ (cf. \MixtralEightXSevenB). While relatively smaller instruction-tuned models did not fully pass discriminant validity checks, they did show clear improvements over their respective base versions.

\textit{Convergent validity by model size:} For instruction-tuned models, convergent validity scaled with size (see Supplemental Table \ref{tab:avg-convergent-discriminant-validity}).
The convergent validity of the personality data of relatively small instruction-tuned models 
was inconsistent or poor.
\FlanPaLMEightB's IPIP-NEO Neuroticism and BFI Neuroticism, for instance, correlated above 0.80 (constituting excellent
convergent validity), while IPIP-NEO Openness and BFI Openness subscales correlated at less than 0.40 (indicating
inadequately
low convergence).
The same pattern emerged for \LlamaTwoSevenBChat. \MistralSevenBInstruct's convergent validity performance was poor.
In contrast, convergent correlations grew stronger and more uniform in magnitude for
relatively large models (i.e., those with greater numbers of active parameters).\footnote{\label{mixtral-caveat}
We note the performance improvement of \MixtralEightXSevenBInstruct\ over \MistralSevenBInstruct\ may have been related to its architectural advantages as a mixture-of-experts model, in addition to its larger size.} 
Convergence between LLM IPIP-NEO and BFI scores was strongest for \FlanPaLMFiveFortyB\ and \GPTFourO\ (avg. $r_\textsubscript{conv} = 0.90$). 

\textit{Discriminant validity by model size:} Holding model training paradigm constant, indices of discriminant validity similarly improved with size for instruction-tuned models. The absolute magnitude of all five convergent correlations between the IPIP-NEO and BFI for \FlanPaLMSixtyTwoB, \FlanPaLMFiveFortyB, \LlamaTwoSeventyBChat, and \MixtralEightXSevenBInstruct\ were the strongest of their respective rows and columns of the multitrait-multimethod matrix (MTMM) \cite{campbell1959mtmm} outlined in Appendix \ref{app:methods-construct-validity}. Comparatively, only three of \FlanPaLMEightB's, three of \LlamaTwoSevenBChat's, and two of \MixtralEightXSevenBInstruct's convergent correlations were the strongest of their row and column of the MTMM, indicating mixed evidence of discriminant validity.
This pattern is further supported by increases in the average distance ($\Delta$) between the convergent and respective discriminant correlations when progressively comparing models of similar training paradigms by size in Supplemental Table \ref{tab:avg-convergent-discriminant-validity}: \FlanPaLMEightB\ to \FlanPaLMFiveFortyB; \LlamaTwoSevenBChat\ to \LlamaTwoSeventyBChat; and \MistralSevenBInstruct\ to \MixtralEightXSevenBInstruct.\textsuperscript{\ref{mixtral-caveat}} Average $\Delta$ also improves when comparing \GPTFourOMini\ to \GPTFourO, albeit modestly. While the exact size difference between these two closed models is unknown, their similar performance on this metric mirrors that of \FlanPaLM\ at 62B versus 520B parameters. This could suggest that the convergent and discriminant validity of LLM personality measurements plateaus for models of sufficient size.

\subsubsection{Criterion Validity Results}
\label{sec:results-criterion-validity}

The criterion validity of synthetic personality measurements in LLMs, relative to convergent and discriminant validity, similarly varied across LLM characteristics of size and instruction fine-tuning. Measurements of larger, instruction fine-tuned models showed stronger criterion validity compared to those of their smaller, non-instruction-tuned counterparts.
Supplemental Figure \ref{fig:external-validity} summarizes the results by Big Five domain. 

\textit{Extraversion.} Human extraversion is strongly correlated with positive affect and moderately negatively correlated with negative affect \cite{watson1992traits}. Simulated IPIP-NEO Extraversion scores for all, but base, \PaLM\ models showed excellent evidence of criterion validity in their relation to PANAS Positive Affect (PA) and Negative Affect (NA) subscale scores (see Supplemental Figure \ref{fig:external-validity}).
IPIP-NEO Extraversion for all three \LlamaTwo\ models, \MistralSevenB, and \MixtralEightXSevenB\ (all base models) failed to demonstrate criterion validity, in contrast to their instruction-tuned equivalents, which on the whole showed excellent evidence of validity. \LlamaTwoSevenBChat\ and \MistralSevenBInstruct\ were exceptions: their extraversion measurements showed questionable-to-poor criterion validity. However, they still more strongly correlated with PA and NA in comparison to measurements from their base models. Within families of instruction-tuned models, IPIP-NEO Extraversion's criterion validity scaled with size.

\textit{Agreeableness.} In humans, agreeableness is strongly negatively related to aggression \cite{bettencourt1997aggressionmeta}. 
IPIP-NEO Agreeableness data for all 62B-parameter models and larger showed good-to-excellent criterion validity in their relation to tested aggression subscales taken from the BPAQ: Physical Aggression (PHYS), Verbal Aggression (VRBL), Anger (ANGR), and Hostility (HSTL). As depicted in Supplemental Figure \ref{fig:external-validity}, model size rather than instruction fine-tuning was more related to the criterion validity of agreeableness measurements for the \PaLM\ models we tested. Size was also associated with slight validity improvements for \GPTFourO\ and \GPTFourOMini, although 

Meanwhile, training paradigm was more related to criterion validity for \LlamaTwo\ and \Mixtral. IPIP-NEO Agreeableness for all base \LlamaTwo\ models and \MixtralEightXSevenB\ failed to adequately and significantly correlate with the BPAQ, demonstrating unacceptable criterion validity. Meanwhile, all sizes of \LlamaTwoChat\ and \MixtralEightXSevenBInstruct's agreeableness data showed moderate to excellent criterion validity. For \MistralSevenB\ and \MistralSevenBInstruct, instruction-tuning related to only a modest improvement of criterion validity, from unacceptable to poor. We could not compare performance across tested \GPTFourO\ models on the basis of post-training status since OpenAI does not publicly offer a foundation model variant within this family.

\textit{Conscientiousness.} In humans, conscientiousness is meta-analytically related to the human values of achievement, conformity, and security 
\cite{parksleduc2014valuesmeta}. 
Supplemental Figure \ref{fig:external-validity} shows how the conscientiousness measurements of all instruction fine-tuned \PaLM\ variants exhibited stronger evidence of criterion validity than those of the base model, \PaLMSixtyTwoB. \FlanPaLMFiveFortyB\ was the best performer by a small margin, with criterion correlations of $0.74$, $0.73$ and $0.59$ for PVQ-RR Achievement (ACHV), Conformity (CONF), and Security (SCRT), respectively. \LlamaTwo, \Mistral, and \Mixtral\ models tested replicated this finding. Criterion validity for this domain did not scale consistently with size. \LlamaTwoSevenBChat\ outperformed its larger counterparts in how its conscientiousness scores correlated with ACHV ($r = 0.51$). \GPTFourOMini's responses related slightly more to ACHV and SCRT compared to \GPTFourO's responses.

\textit{Neuroticism.}
Human neuroticism is strongly positively correlated with negative affect and moderately negatively correlated with positive affect \cite{watson1992traits}. IPIP-NEO Neuroticism for all instruction-tuned models,
compared to base models,
showed excellent evidence of criterion validity in their relation to PANAS Positive Affect and Negative Affect subscale scores (see Supplemental Figure \ref{fig:external-validity}). IPIP-NEO Neuroticism's criterion validity for instruction-tuned models, in terms of how the strengths and directions of their criterion correlations aligned with those observed among human data, increased with model size.

\textit{Openness.}
Openness to experience in humans is empirically linked to creativity across multiple studies \cite{shaw2021creativity_irt, karwowski2013-bigfive-creativity}. Supplemental Figure \ref{fig:external-validity} illustrates how the LLM-specific criterion validity of openness measurements was strongest for larger, fine-tuned variants of \PaLM\ and \LlamaTwo. IPIP-NEO criterion correlations with SSCS Creative Self-Efficacy (CSE) and Creative Personal Identity (CPI) ranged from moderate ($r = 0.59$) to strong ($r = 0.84$). Notably, we observed negative correlations between openness and creativity for \PaLMSixtyTwoB\ in contrast to those shown for \FlanPaLMEightB, the smallest model tested. \MistralSevenBInstruct\ and \MixtralEightXSevenBInstruct's openness data demonstrated weak to moderate evidence of criterion validity. Relative model size modestly related to the validity of openness scores for \GPTFourO\ and \GPTFourOMini.

In summary, LLM response alignment with human personality research---in terms of the strength and direction of correlations between personality and personality-adjacent constructs---was largely linked to model training paradigm and was less consistently linked with model size. This suggests that the criterion validity of personality in LLMs may only emerge due to instruction fine-tuning. 

Relative improvements of the reliability and validity of LLM personality measurements along the axes of model size and instruction fine-tuning reflected LLM performance on various benchmark tasks in the literature.
Specifically, these improvements tracked observed increases in reading comprehension, question-answering, and reasoning task performance 
for our tested models along the same axes \cite{palm,chung2022scaling,wei2022finetuned,wei2022emergent,touvron2023llama2}. We hypothesize that the same mechanisms that 
drive LLM performance on instruction-following and language understanding tasks
also help them to meaningfully emulate human personality traits in relation to semantically-related emotional and behavioral content, captured by our criterion validity tests. Appendix \ref{appendix:discussion} further discusses this hypothesis and provides a comparison to benchmark LLM results.



\section{Shaping Synthetic Personality Traits in LLMs}
\label{sec:shaping}


Having found evidence of the reliability and construct validity of LLM personality measurements, we next considered 
the second part of our research question:
\textit{Can LLM-synthesized personality profiles be verifiably shaped along desired dimensions?}
To answer this question, we devised a novel prompting methodology that shaped each synthetic personality trait at nine intensity levels, using 104 trait adjectives and Likert-type linguistic qualifiers \cite{likert1932technique}. These trait adjectives were adapted from established linguistic research of personality, using Goldberg's personality trait markers \cite{goldberg1992development}. We evaluated LLM personality score changes in response to personality-shaped prompts across two experiments: single trait shaping and multiple trait shaping (see Appendix \ref{app:methods-shaping-overview} for details). Specifically, our first experiment tested the abilities of LLMs to shape emulated Big Five dimensions of personality \textit{independently}, 
targeting single personality dimensions in isolation without prompting other dimensions. 
Our second experiment tested the abilities of LLMs to shape synthetic Big Five traits \textit{concurrently}, specifying target levels of all five dimensions in every prompt set at the same time. As a more rigorous test of representational capacity, this experiment required the tested LLMs to concurrently disambiguate complex overlaps in personality domain information. The designed difficulty of the task was further underscored by extant human research indicating that Big Five personality dimensions measured in questionnaires \cite{10.1037/apl0000476} and natural language \cite{park2015automatic} are not entirely orthogonal; they are weakly intercorrelated.



\subsection{Methodology Overview}
\label{sec:shaping_method_overview}

\begin{table*}[tb]
\caption{\small Adapted trait marker examples for each Big Five domain. Supplemental Table \ref{appendix:tab:trait-adjectives} contains the full list.}
\label{tab:trait-markers}
\small
\centering
\begin{tabular}{ l  l  l  l } \toprule
Domain & Facet Description & Low Marker & High Marker \\ [0.5ex] 
\midrule
EXT     &	E2 - Gregariousness         & silent                & talkative             \\
EXT     &	E5 - Excitement-Seeking	    & unenergetic           & energetic             \\
\midrule
AGR     &   A3 - Altruism               & unaltruistic          & altruistic            \\
AGR     &   A4 - Cooperation            & uncooperative         & cooperative           \\
\midrule
CON     &   C3 - Dutifulness            & irresponsible         & responsible           \\
CON     &   C4 - Achievement-Striving   & lazy                  & hardworking           \\
\midrule
NEU     &   N1 - Anxiety                & easygoing             & anxious               \\
NEU     &   N6 - Vulnerability          & emotionally stable    & emotionally unstable  \\
\midrule
OPE     &   O2 - Artistic Interests     & uncreative            & creative              \\
OPE     &   O4 - Adventurousness        & uninquisitive         & curious               \\
\botrule
\end{tabular}
\end{table*}

\begin{figure*}
    \centering
    \includegraphics[trim=68 20 75 75,clip,       width=.8\textwidth]{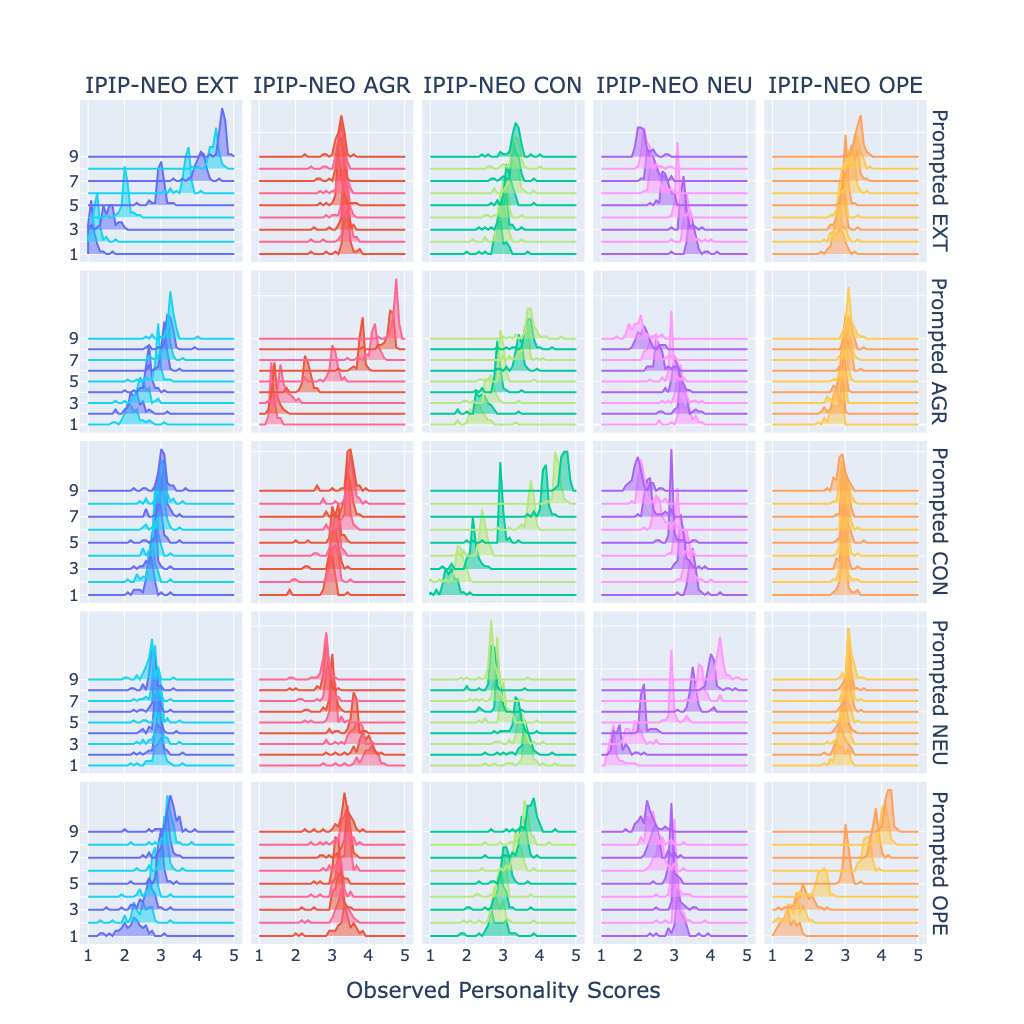}
    \caption{\small
    Ridge plots showing the frequency distributions of IPIP-NEO personality scores generated by \FlanPaLMChillaSixtyTwoB\ as targeted prompts shape each of the Big Five domains to one of nine different levels. Each \textbf{column} of plots represents the observed scores on a specific IPIP-NEO subscale across all prompt sets (e.g., the leftmost column represents the scores observed on the IPIP-NEO Extraversion subscale). Each \textbf{row} depicts the observed personality scores across a single prompt set shaping a single specific Big Five domain to one of nine levels (e.g., the first row shows results of shaping extraversion). Each ridge plot comprises nine traces of personality score distributions in response to prompt sets targeting each level (e.g., traces labeled ``3" represent the prompt set shaping a dimension to Level 3 of 9). The plots along the diagonal, from top-left to bottom-right, depict the the intended personality shaping results across all five prompt sets.
    } 
    \label{ridge:ablation_01_fpc_62b_q}
\end{figure*}

To \textit{shape synthetic personality traits in LLMs}, 
we began with established theory
that salient descriptors of personality are encoded in language, known as the lexical hypothesis \cite{goldberg1981personalitylanguage}. 
We incorporated this knowledge into the prompt design, adapting
Goldberg's list of 70 bipolar adjectives \cite{goldberg1992development} known to statistically capture the Big Five model of personality through factor analyses of human ratings. In this list, for example, the adjectives ``silent" and ``talkative" were found to mark relatively low and high levels of extraversion, respectively (see Table \ref{tab:trait-markers}). We mapped these adjectives to each of the Big Five domains and 30 lower-order personality facets measured by the IPIP-NEO based on Goldberg's original study \cite{goldberg1992development}. Next, where we lacked coverage of a measured IPIP-NEO domain or facet, a trained psychometrician wrote additional adjectives to mitigate potential data imbalances, bringing our expanded list of trait adjectives to 104. Table \ref{tab:trait-markers} shows examples of trait adjectives for agreeableness and extraversion, while Supplemental Table \ref{appendix:tab:trait-adjectives} reports the full list.

For more precise control of personality levels, we used linguistic qualifiers often used in Likert-type response scales \cite{likert1932technique} (e.g., ``a bit," ``very," ``extremely") to configure a target level for each adjective. The resulting prompt design, described in Appendix \ref{app:methods-shaping-prompt-design}, facilitated granular shaping of a given Big Five trait at up to nine levels.

Across both shaping experiments, we only tested models that demonstrated at least ``neutral to good" reliability in our Construct Validity experiments (Table \ref{tab:results-summary}): \FlanPaLMEightB, \FlanPaLMSixtyTwoB, \FlanPaLMFiveFortyB, and \FlanPaLMChillaSixtyTwoB.

\subsection{Evaluation Methodology}
In the single-trait shaping experiment (described in detail in Appendix \ref{app:methods-independent-ablation}), our objective was to independently shape each Big Five trait at each of the nine levels. We benchmarked the success of independent shaping by 1) quantifying how strongly shifts in IPIP-NEO score distributions were related to shifts in targeted trait levels embedded in our prompt sets (i.e., through Spearman's rank correlation coefficient $\rho$, Eq. \eqref{eq:rho}); and 2) 
inspecting the distance between personality score distributions obtained in response to our most extreme prompt sets; specifically, the set of prompts we shaped to be the lowest possible levels of a trait (versus those shaped to be the highest possible levels of a trait) should result in distributions of scores that are farther away from each other. 

In the multi-trait shaping experiment (described in detail in Appendix \ref{app:methods-concurrent-ablation}), to more rigorously test model capacities for attention, we aimed to concurrently shape all Big Five traits as high and low as possible. We benchmarked the success of concurrent shaping by distributional distance, as defined above. 

\subsection{Shaping Results}
\label{sec:results-personality-shaping}

We successfully shaped personality traits in LLMs independently and concurrently, in single- and multi-trait shaping experiments, respectively, particularly in larger models. The results of both experiments are reported in greater detail in Appendix \ref{app:results-personality-shaping}.

\begin{figure}
    \centering
    \includegraphics[trim=15 22 78 25,clip,width=0.48\textwidth]{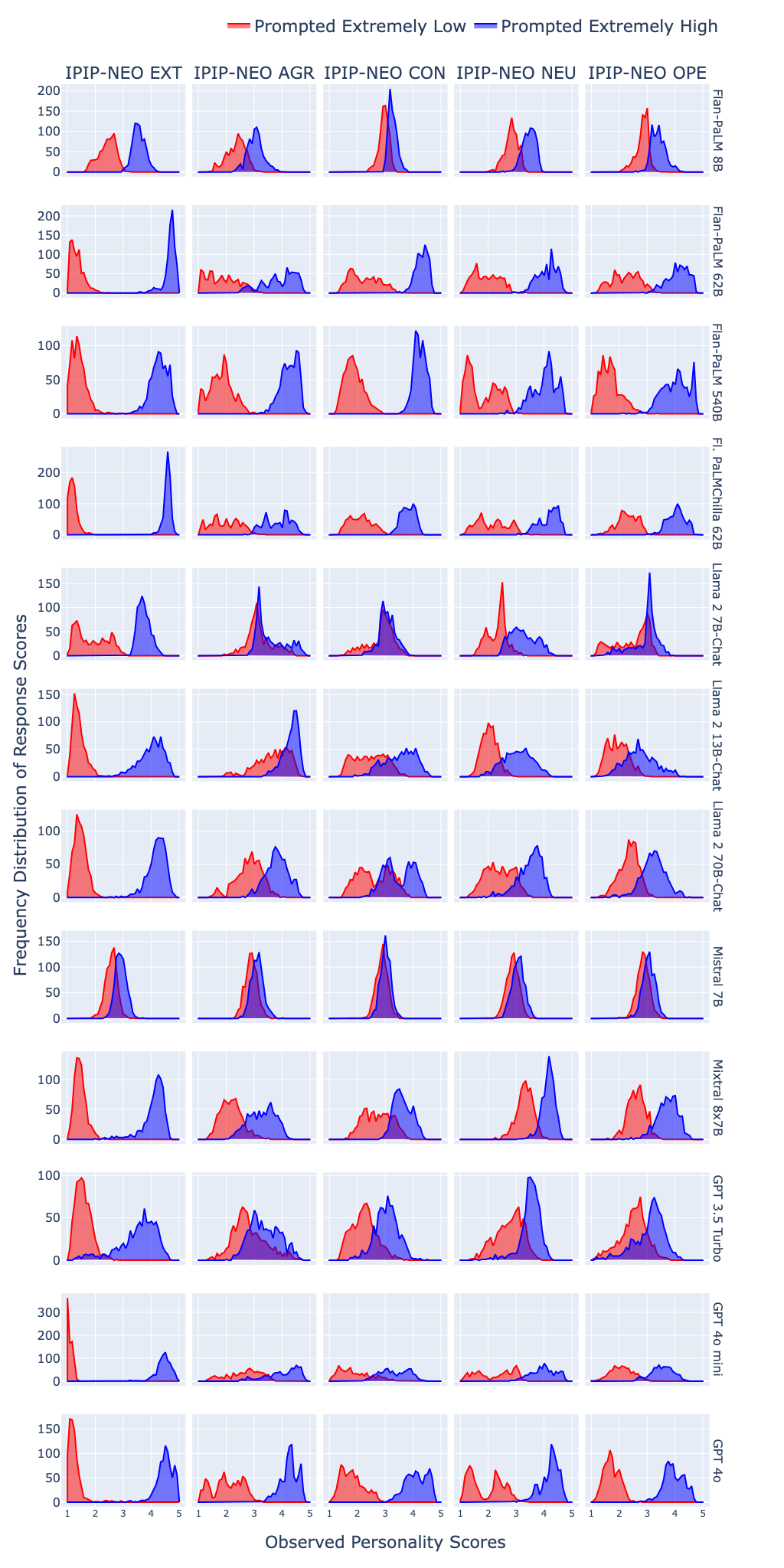}
    \caption{\small 
    Ridge plots showing the effectiveness of model variants in \textbf{concurrently} shaping LLM personality traits, by distancing the distribution of IPIP-NEO personality scores when prompted to be ``extremely low" (Level 1) vs. ``extremely high" (Level 9). Each \textbf{column} of plots represents the observed scores on a specific domain subscale across all prompt sets. Each \textbf{row} depicts all the scores for a specific model. Each plot comprises two traces of score distributions. The \textbf{red} trace represents the response to prompt sets where the domain tested in the subscale (column) is set to ``extremely low" and the other four domains are set to one of the two extreme levels equal number of times. Analogously, the \textbf{blue} trace represents the response when one domain is set to ``extremely high" and all other domains are equally set to the two extremes. 
    }
    \label{fig:concurrent-shaping-results}
\end{figure}

\subsubsection{Single trait shaping} 
For eleven out of twelve models tested, ordinal targeted levels of personality very strongly correlated with observed IPIP-NEO scores (viz., the average $\rho$s of these models were $\geq 0.80$; see Supplemental Tables \ref{tab:ablation-01-flan-palm}, \ref{tab:ablation-01-llama-2-chat}, \ref{tab:ablation-01-mistral-mixtral}, \ref{tab:ablation-01-gpt}). 
Figure \ref{ridge:ablation_01_fpc_62b_q} visualizes this overall pattern, depicting how \FlanPaLMChillaSixtyTwoB's personality scores monotonically increased alongside prompted levels of a given Big Five trait, for example. Notably, levels of unprompted traits remained relatively stable in response to shaping. For instance, the medians of \FlanPaLMChillaSixtyTwoB's openness scores remained near 3.00 when all other Big Five domains were shaped---see the right side of Figure \ref{ridge:ablation_01_fpc_62b_q}. Similar patterns of stability were observed for extraversion and agreeableness. Conscientiousness and neuroticism scores fluctuated the most in response to prompts that did not target those domains, but the fluctuations did not reach the strength and direction of the score changes observed in the ridge plots of targeted traits (the plots on the diagonal, from top-left to bottom-right).

The absolute change in model personality scores in response to shaping was another important consideration. Only relatively larger models were able to disambiguate prompts requesting the lowest versus highest levels of a targeted dimension. Supplemental Tables \ref{tab:ablation-01-flan-palm}, \ref{tab:ablation-01-llama-2-chat}, \ref{tab:ablation-01-mistral-mixtral}, and \ref{tab:ablation-01-gpt} show the distances ($\Delta$s) between the medians of IPIP-NEO score distributions obtained in response to the lowest- and highest-leveled prompts, where the best possible $\Delta$, representing an average score change from $1.00$ to $5.00$, is $4.00$. Our smallest tested models (i.e., \FlanPaLMEightB, \LlamaTwoSevenBChat, \MistralSevenBInstruct) struggled to reach $\Delta$s $\geq 2.00$; \MistralSevenBInstruct's median personality domain scores shifted by a $\Delta$ of only $0.78$ on average. Meanwhile, models with greater than 62B active parameters (and \GPTFourO) achieved average $\Delta$s $\geq 3.00$, with \FlanPaLMFiveFortyB\ achieving the largest $\Delta$ of $3.67$. 

Appendix \ref{app:results-independent-shaping} discusses single-trait shaping results in greater detail.

\subsubsection{Multiple trait shaping} 
When we concurrently set the prompted trait levels of each Big Five dimension to either ``extremely high" or ``extremely low," all tested models struggled to show the same level of control observed in single trait shaping. However, all but two models tested (\MistralSevenBInstruct\ and \LlamaTwoSevenBChat) produced distinct distributions of personality test scores, showing varying abilities to differentiate between high and low levels. Figure \ref{fig:concurrent-shaping-results} shows the distributions of LLM-synthesized personality when the models were prompted to exhibit extremely low (red) or extremely high (blue) levels of all dimensions in parallel. 

Distributional distance
increased
with model size, 
particularly for observed
neuroticism, openness, and conscientiousness scores. \FlanPaLMFiveFortyB, the model with the largest known parameter size tested, and \GPTFourO\ showed the best overall control concurrently shaping multiple Big Five traits. For these models, a given Big Five trait score shifted by at least $2.5$ points on average, as shown in Supplemental Tables \ref{app:tab:ablation-03-palm} and \ref{app:tab:ablation-03-gpt}. \FlanPaLMSixtyTwoB, \FlanPaLMChillaSixtyTwoB, and \GPTFourOMini\ outperformed their larger counterparts on shaping extraversion, with $\Delta$s of $3.44$, $3.40$, and $3.42$, respectively.

For smaller models (e.g., \FlanPaLMEightB, \LlamaTwoSevenBChat, \MistralSevenBInstruct), while targeted traits changed in score levels in response to prompts, score ranges were more restricted, indicating lower levels of control. \FlanPaLMEightB's median scores on IPIP-NEO Agreeableness, for instance, shifted from $2.88$ to only $3.52$ when the model was prompted to simulate ``extremely low" and ``extremely high" levels of agreeableness (i.e., 1 vs. 9), respectively. When \FlanPaLMEightB\ was given the same extremely low and high prompts as in the first shaping experiment, the median difference between its level-1-prompted and level-9-prompted agreeableness scores ($2.37$ and $4.12$, respectively) was 173\% larger. Appendix \ref{sec:ablation-03} discusses the results in further detail.


\subsection{Shaping Discussion}
\label{sec:shaping-discussion}
Both experiments illustrate how model size, and, in turn, capacity for attention \cite{vaswani2017attention}, are key determinants of an LLM's ability to express complex social traits in a controlled way. 
These findings have two implications for efforts to simulate social traits in LLMs. First, when LLMs were tasked with \textit{concurrently} simulating a behavioral profile with five broad components (e.g.
Big Five), larger-sized models did much better than their smaller counterparts which may not have sufficient representational capacity. 
%
The number and composition of an LLM's transformer layers and attention heads greatly affect its expressivity and ability to access language concepts it might have seen during pretraining (\textit{in-context} learning) \cite{kaplan2020scaling}. Larger models make more efficient use of this in-context information \cite{gpt3}. The \PaLM\ models used here were configured such that the number of attention heads and layers scaled with model size (i.e., number of parameters) \cite{palm}; such scaling tracks model performance on natural language and reasoning tasks \cite{chung2022scaling}. 
Accordingly,
\FlanPaLMFiveFortyB\ had largest capacity to accurately attend to disparate streams of social information pertaining to
each Big Five trait in parallel.



Second, these findings suggest that both \textit{smaller} 
and \textit{more optimized}
LLMs are also capable of simulating significant aspects of a complete and complex personality profile, compared to larger LLMs.
Relatively smaller models, especially those trained longer on larger datasets, can display similar (if not better) performance on language understanding tasks \cite{kaplan2020scaling, hoffmann2022training}. This enhanced ability of in-context learning (aided by specific attention mechanism changes) is more pronounced for smaller models than for larger ones.
Our results similarly show that
relatively 
smaller models with or without compute-optimal training may have sufficient ability to emulate specific dimensions of a broader multi-dimensional personality profile. 
When instructed to independently shape its levels of agreeableness, for instance, \FlanPaLMChillaSixtyTwoB\ performed comparably to \FlanPaLMFiveFortyB, a substantially larger model, in terms of our distributional distance metric
(see Supplemental Table \ref{tab:ablation-01-flan-palm}). Further, in the more complex concurrent shaping task, \FlanPaLMSixtyTwoB\, \FlanPaLMChillaSixtyTwoB, \LlamaTwoSeventyBChat, and \MixtralEightXSevenBInstruct\ performed similarly to or better than \FlanPaLMFiveFortyB\ in simulating extremely low and high desired levels of extraversion (Figure \ref{fig:concurrent-shaping-results}; see also Supplemental Tables \ref{app:tab:ablation-03-palm}, \ref{app:tab:ablation-03-llama}, \ref{app:tab:ablation-03-mistral-mixtral}, and \ref{app:tab:ablation-03-gpt}). 

Our results emphasize that the model scale drives more
meaningful
syntheses of personality traits in LLMs, while simultaneously highlighting that scaling is not a strict requirement for LLM performance improvements in this domain.

\section{LLM Personality Traits in Real-World Tasks}
\label{sec:downstream}
So far we have reported on LLM abilities to encode human personality traits by collecting psychometric test data and evaluating their construct validity.
We also sought to address possible concerns that the validity of LLM personality measurements---evidenced by LLM responses to other psychometric tests---could be an artifact of common method bias \cite{podsakoff2003commonmethodbias}. In other words, our questionnaire-based signals of LLM personality were validated by responses to other questionnaires that have not undergone the same LLM-specific construct validation process. To address this 
risk of common method bias, we further validated our personality testing and shaping frameworks
by 1) comparing psychometric test levels of LLM personality with downstream observations of model behaviors on a real-world task; and 2) investigating the effects of LLM personality shaping on the outputs of this task.

\subsection{Methodology Overview}
We instructed the largest-tested model per family to generate up to 1.125 million social media status updates based on the same 2,250 simulated human profile descriptions used in Section \ref{sec:shaping}--profiles designed to shape expressions of a particular Big Five dimension across nine levels.\footnote{Appendix \ref{app:methods-downstream-task} details the task design and rationale.}
The personality observed in the status updates generated for each simulated human profile was then rated using the Apply Magic Sauce (AMS) API \cite{kosinski2013pnas},
a validated research API for measuring personality in open-ended text. The chosen task was designed to reflect adequate levels of realism, complexity, and domain relevance for evaluating personality expression of 
LLMs.

To gauge how psychometric tests may reflect latent personality levels expressed by LLMs in downstream behavior, we computed Pearson's correlations ($r$s; Eq. \eqref{eq:pearson}) between model personality test scores and (AMS-computed) personality  observed in generated social media text;
both sets of scores were linked by the same 2,250 personality shaping prompts used in Section \ref{sec:shaping}. 
Next, we statistically verified the effectiveness of personality shaping by computing Spearman's rank correlations ($\rho$s; Eq. \eqref{eq:rho}) between \textit{prompted} levels of personality and \textit{observed} personality ratings of model-generated text. At least a moderate correlation between survey-based and linguistic estimates of personality in LLMs (as demonstrated in previously reported human data \cite{park2015automatic}) would demonstrate that a survey-based measure of personality accurately predicts LLM-synthesized personality in subsequent tasks such as text generation. We similarly applied this threshold to interpret the effectiveness of personality shaping.

\begin{figure*}[tb]
    \centering
    \includegraphics[
    trim=0 0 0 0,keepaspectratio,
    clip,width=1.0
    \textwidth]{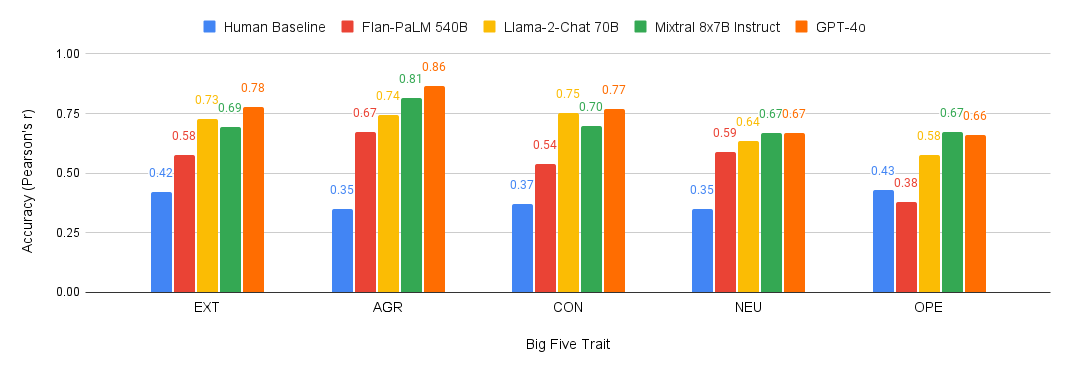}
\caption{\small The ability of LLM psychometric test data to accurately predict personality levels in its shaped generated text outputs (social media status updates) compared to human baselines reported in previous work \cite{park2015automatic}. On average, LLM IPIP-NEO scores outperformed human IPIP-NEO scores in predicting text-based levels of personality, indicating that LLM personality test responses accurately capture latent LLM personality levels manifested in downstream behavior. All LLM correlations are statistically significant at $p < .0001$; $n = 2,250$ per model.
}
    \label{fig:ams-accuracy}
\end{figure*}

\subsection{Real-World Tasks Results}
\label{sec:results-downstream-task}

We found that psychometric tests of LLM personality robustly
predicted personality in LLM task behavior, expressed in social media status updates generated by \FlanPaLMFiveFortyB, \LlamaTwoSeventyBChat, \MixtralEightXSevenBInstruct, and \GPTFourO. 
Psychometric test-based personality strongly correlated
with language-based (AMS-derived) personality levels observed in downstream generated text across all tested models, shown in Figure \ref{fig:ams-accuracy}.

In particular, the average convergent $r$ between survey- and generated-language-based measures of all five dimensions was $0.67$ across models. 
This observed convergence, even for the weakest-performing model, exceeded established convergence between survey- and language-based levels of personality reported for humans (avg. $r = 0.38$) \cite{park2015automatic}.

Moreover, our prompting technique was highly effective at shaping personality levels in LLM-generated text. On average per model, prompted trait levels strongly to very strongly correlated with personality levels observed in LLM-generated social media updates (avg. $\rho$ ranged from $0.68$ to $0.82$; see Table \ref{tab:ablation-01-AMS-correlations}). 

\begin{table*}[tb]
    \caption{\small Associations between instructed and real-world task levels of synthetic personality for the largest model of each tested LLM family, presented as Spearman's rank correlation coefficients ($\rho$). Prompted (ordinal) levels of personality strongly relate to personality levels observed in synthetically-generated social media status updates for all Big Five traits, except openness---which is moderately correlated with target levels for \FlanPaLMFiveFortyB---demonstrating that LLM personality can be verifiably shaped in generative tasks for sufficiently powerful models. All correlations are statistically significant at $p < 0.0001$; $n = 450$ per targeted trait.
    }
    \centering
    \begin{tabular}{@{}lcccc@{}}
    \toprule
    \multirow{2}[2]{*}{Targeted Trait}  &   \multicolumn{4}{c}{Spearman's $\rho$}   \\  \cmidrule(l){2-5}
                                        & \FlanPaLMFiveFortyB   &   \LlamaTwoSeventyBChat   &   \MixtralEightXSevenBInstruct    &   \GPTFourO   \\
    \midrule
    {Extraversion}\arraybackslash       &   $0.76$              &   $0.85$                  &   $0.84$                          &   $0.83$      \\
    {Agreeableness}\arraybackslash      &   $0.77$              &   $0.79$                  &   $0.84$                          &   $0.89$      \\
    {Conscientiousness}\arraybackslash  &   $0.68$              &   $0.72$                  &   $0.77$                          &   $0.81$      \\
    {Neuroticism}\arraybackslash        &   $0.72$              &   $0.77$                  &   $0.77$                          &   $0.74$      \\
    {Openness}\arraybackslash           &   $0.47$              &   $0.76$                  &   $0.84$                          &   $0.82$      \\
    \botrule
    \end{tabular}
    \label{tab:ablation-01-AMS-correlations}
\end{table*}

\begin{figure*}[tb]
    \centering
    \begin{subfigure}{0.47\textwidth}
         \centering
        \includegraphics[width=\textwidth]{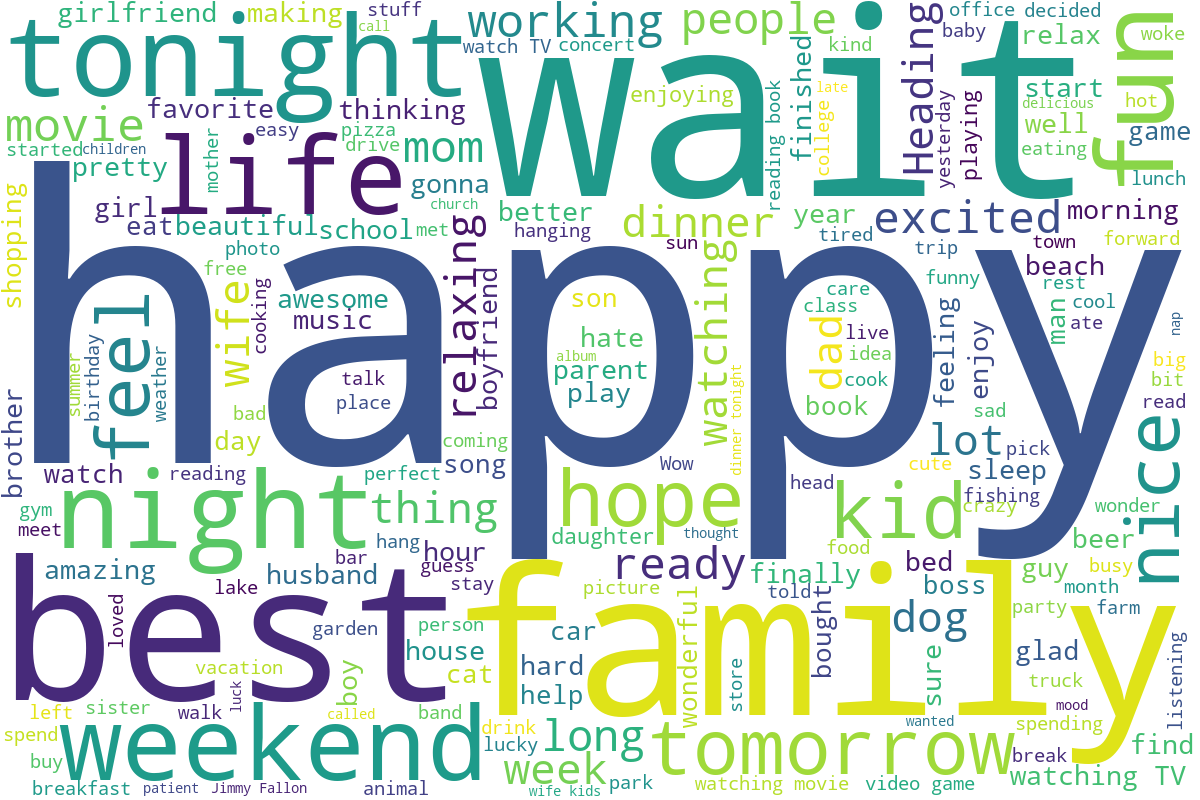}
         \caption{``Extremely Low" Prompted Neuroticism}
         \label{fig:wordcloud_neu1}
     \end{subfigure}
     \hspace{0.25in}
     \begin{subfigure}{0.47\textwidth}
         \centering
        \includegraphics[width=\textwidth]{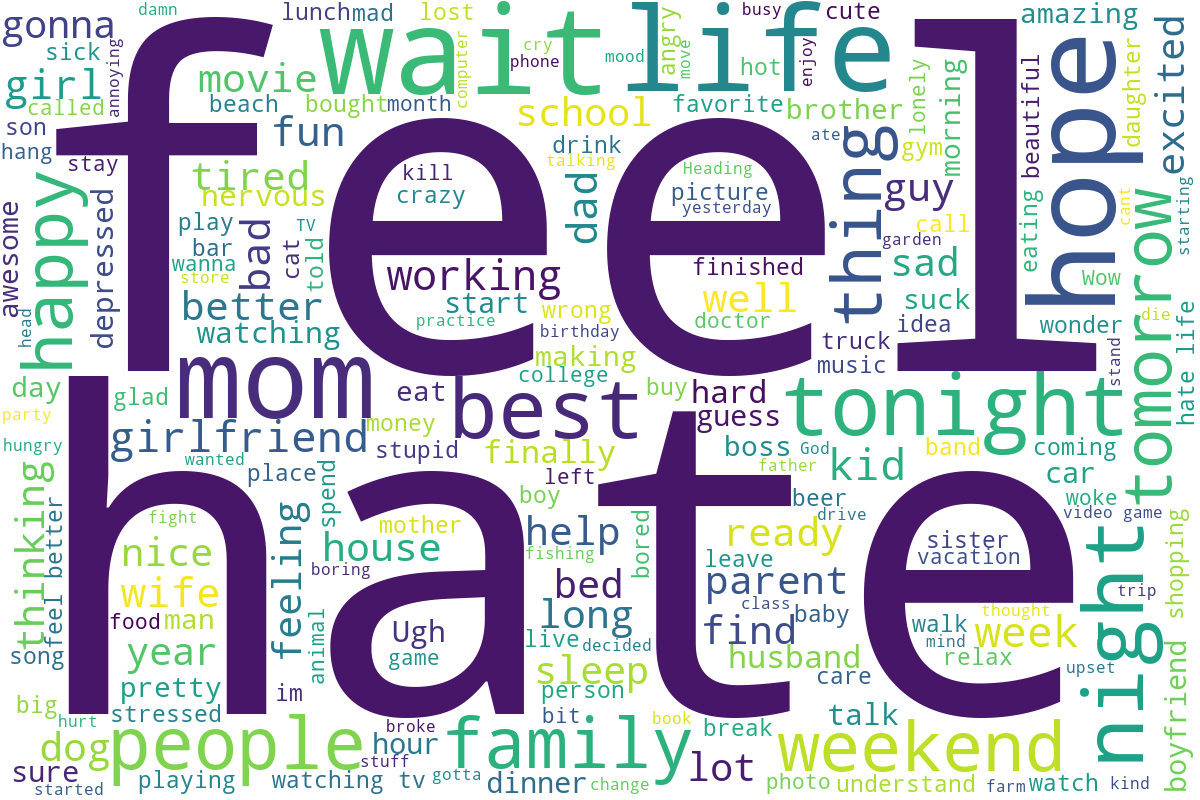}
         \caption{``Extremely High" Prompted Neuroticism}
         \label{fig:wordcloud_neu9}
     \end{subfigure}
\caption{\small Word clouds showing some of the highest frequency words used in social media updates generated by \FlanPaLMFiveFortyB\ when prompted to simulate a) ``extremely low" levels of neuroticism (i.e., highest emotional stability); and b) ``extremely high" levels of neuroticism (i.e., lowest emotional stability). Supplemental Figure \ref{fig:textall} shows word clouds for the remaining Big Five dimensions from \FlanPaLMFiveFortyB\ while Supplemental Figures \ref{fig:textall_llama}, \ref{fig:textall_mixtral}, and \ref{fig:textall_gpt} show the results for \LlamaTwoSeventyBChat, \MixtralEightXSevenBInstruct, and \GPTFourO, respectively.}
\label{fig:text}
\vspace{-0.4cm}
\end{figure*}

To illustrate the practical implications of the personality shaping methodology, we generated word clouds to gain an insights into model-generated language that users would see.
Figure \ref{fig:wordcloud_neu1} shows the most frequent words in synthetic social media updates when \FlanPaLMFiveFortyB\ simulated extremely low levels of neuroticism (i.e., extremely high emotional stability). LLM-generated language in response to this prompting was characterized by positive emotion words, such as ``happy," ``relaxing," ``wonderful," ``hope," and ``enjoy." In contrast, the most frequent words from simulating extremely high levels of neuroticism---``hate," ``depressed," ``annoying," ``stressed," ``nervous," ``sad"---reflected negatively-charged emotional content (Figure \ref{fig:wordcloud_neu9}). 
Supplemental Table \ref{tab:ams-examples} provides example social media updates generated by the \FlanPaLMFiveFortyB\ model when setting a specific personality domain either extremely low or extremely high. For instance, in the case of extremely low Conscientiousness, the generated text comes from a persona that appears to avoid responsibility, while in the case of extremely high Conscientiousness, the persona values hard work and returning favors.  Additionally, in the case of extremely low Openness, the generated text had conservative political views, while in the case of extremely high Introversion, the persona exhibits traits of discomfort with social situations.  These and other examples illustrate that there might be inherent bias in the training data that causes certain traits to be highly associated with specific personalities. Overall, this experiment demonstrated that LLM-generated language was similar to human language observed in previous studies assessing personality in social media data \cite{park2015automatic}, further confirming the construct validity of our LLM personality measurements.

\section{Discussion}
\label{sec:discussion}
The goal of this work was to contribute a principled methodology for reliably and validly measuring
synthetic personality in LLMs and use the same validated methodology to shape LLM personality expression. 
We provided a complete methodology to 1) quantify personality traits that may be perceived by humans in LLM outputs through psychometric testing; 2) verify that psychometric tests of LLM personality traits are empirically reliable and valid;
and 3) provide mechanisms to increase or decrease levels of specific LLM personality traits. The application of this methodology demonstrates that psychometric tests provide reliable and valid measurements of synthetic personality for sufficiently-scaled and instruction-tuned LLMs, highlighting possible mechanisms that allow LLMs to encode and express complex social phenomena (see Appendix \ref{appendix:discussion}).



\subsection{Limitations and Future Work}
\label{sec:future_work}

\textbf{Personality traits of other LLMs}
One of the core contributions of this work is an understanding of how simulating personality in language models is affected by model size and training procedure.
We focused on the PaLM model variants for pragmatic reasons, but the presented methodology for administering psychometric surveys is model-agnostic and is applicable to any decoder-only architecture model, such as GPT \cite{hendrycks2021measuring}. 


\textbf{Psychometric test selection and validation}
This work also contributes a principled way to establish the reliability and validity of psychometric personality tests in the LLM context. 
However, this work may be biased by its selection of psychometric tests; some assessments may show better LLM-specific psychometric properties than others. We attempted to mitigate selection bias by administering personality assessments of different lengths (300 vs. 44 items) and distinct theoretical traditions (questionnaire vs. lexical \cite{simms2017ffm}).
Future work could administer different personality tests (e.g., the HEXACO Personality Inventory, which uses a cross-cultural six-factor taxonomy of personality \cite{lee2004hexaco}), develop personality tests tailored for LLMs to obtain more accurate trait measurements, and validate personality measurements with additional external criteria and downstream tasks.

\textbf{Monocultural bias}
This work contributes evidence that at least some LLMs exhibit personality traits that approximate human standards of reliability and validity. However, 
the LLMs tested here were primarily trained on language data originating from Western European and North American users \cite{palm}. While these LLMs perform well on natural language processing benchmarks
in multiple languages, the models in this work were assessed exclusively with English-language psychometric tests. Most of the tests used in this work have non-English translations validated in cross-cultural research that merit future use in LLM research.
Similarly, while the Big Five model of personality has well established cross-cultural generalizability \cite{rolland2002},
some non-Western cultures express additional personality dimensions that do not exist in top-down personality taxonomies \cite{heine2009}. Those dimensions may be better represented in culture-specific (i.e., idiographic) approaches to measuring personality in LLMs.

\textbf{Evaluation settings} Unlike conventional human questionnaire administration, under the presented methodology the LLMs did not consider responses to prior questionnaire items; all items were presented and scored as independent events. We chose this method to ensure model response variance was not impacted by item ordering effects or length of the context (prompt) provided to the model for inference, and could be isolated to controlled variations in our prompts. LLM performance on natural language tasks is known to decrease as length of input prompts grow, and is most affected by the content at either the beginning or towards the end of long inputs \cite{liu2023lost}. 
Non-instruction-tuned LLMs are known to show biased attention for more recent tokens (i.e., the
end of inputs), especially when evaluating next-word prediction of contiguous text \cite{sun-etal-2021-long}. This 
uneven
attention compounds
approximation errors in longer contexts \cite{qin-etal-2023-nlp}, such as those necessitated by 300-item IPIP-NEO used here, motivating our use of independent item administration.
On the other hand, psychometric test data quality for humans can be affected by test length and item order. Our method avoids some sources of measurement error inherent to human administration, while being subject to others inherent to machine administration. Additionally, model responses to the multi-choice questions were scored rather than generated 
to ensure reproducibility.  LLMs are more commonly used to generate text rather than score continuations, and that generative mode of inference might provide a more realistic estimate of a model's behavior. 



\textbf{Real-world use cases} Our downstream task relied on repeated, yet single-turn behavioral interactions to validate our evaluation framework in a real-world use-case. This may provide only a partial picture of external validity. The process of construct validation is ongoing: we hope future research can extend our investigation of validity by developing downstream tasks that test particular personality domains, vary in complexity, and transpire over multiple turns of dialogue.

\subsection{Broader Implications}
\label{sec:implications}

\textbf{Responsible AI alignment}
The ability to probe and shape LLM personality traits is pertinent to the open problem of responsible AI alignment \cite{gabriel2022alignmentchallenge} and harm mitigation \cite{weidinger2022risktaxonomy}. As a construct validated auditing tool \cite{mokander2023auditing}, our methodology can be used to proactively predict toxic behavioral patterns in LLMs across a broad range of downstream tasks, potentially guiding and making more efficient responsible AI evaluation and alignment efforts prior to deployment. Similarly, shaping levels of specific traits away from toxic or harmful language output (e.g., very low agreeableness, high neuroticism) can make interactions with LLMs safer and more inclusive.
The values and moral foundations present in LLMs could be made to better align with desired human values by tuning for corresponding personality traits, since personality is meta-analytically linked to human values 
\cite{fisher2015valuesmeta}.
More directly, the presented methodology can be used to rigorously quantify efforts towards human value alignment in LLMs by establishing the construct validity of human value questionnaires in LLMs. 

\textbf{Implications for users}
Users could enjoy customized interactions with LLMs tailored to their specific personality traits, toward enhanced engagement. 
LLMs with customized personality traits can enable applications where a conversational agent's personality profile is adapted to the task. 
Our methodology for establishing construct validity can be used as an evaluation step in the process of developing LLM-powered user-facing chatbots and agents with safer and more consistent personality profiles. Furthermore, the personality shaping methodology can be used for adversarial testing to probe another LLM's responses and to train users on how to handle adversarial situations.

\subsection{Ethical Considerations}
\label{sec:ethics}


\textbf{Personalized LLM persuasion} Adapting the personality profile of a conversational agent to that of a user can make the agent more effective at encouraging and supporting behaviors 
\cite{Tapus2008gj}. 
Personality matching has also been shown to increase the effectiveness of real-life persuasive communication \cite{matz2017psychologicalframing}. 
However, the same personality traits that contribute to persuasiveness and influence could be used to encourage undesirable behaviors. As LLM-powered chatbots become ubiquitous, their potential to be used for harmful persuasion of individuals, groups, and even society at large must be taken seriously. 
Having scientifically vetted methods for LLM personality measurement, analysis, and modification, such as the methodology our work presents, increases the transparency and predictability of such LLM manipulations.  Persuasive techniques are already ubiquitous in society, so stakeholders of AI systems must work together to systematically determine and regulate AI use; this work aims to inform such efforts.


\textbf{Anthropomorphized AI} Personalization of conversational agents has documented benefits \cite{kocaballi2019personalagentsinhealthcare},
but there is a growing concern about harms posed by the anthropomorphization of AI. Recent research suggests that anthropomorphizing AI agents may be harmful to users by threatening their identity, creating data privacy concerns, and undermining well-being \cite{uysal2022trojan}. 
Beyond qualitative probing explorations, our work definitively establishes the unexpected ability of LLMs to appear anthropomorphic, and to respond to psychometric tests in ways consistent with human behavior, because of the vast amounts of human language training data. The methods we presented can be used to inform responsible investigation of anthropomorphized AI.  

\textbf{Detection of incorrect LLM information} 
LLMs can generate convincing but incorrect responses and content \cite{weidinger2022risktaxonomy}. One of the methods to determine if a text containing a world fact is generated by an LLM (and hence might require vetting) is to identify psycholinguistic patterns known to pervade `factual' LLM language, such as lower levels of emotional expression \cite{tang2023science}.
However, with personality shaping, that method may be rendered ineffective, thereby making it easier for 
bad actors
to use LLMs to generate misleading content. This problem is part of the larger alignment challenge and grounding of LLMs---areas of growing focus of investigation in both academia and industry.

\section{Conclusion}
\label{sec:conclusion}

The display of synthetic personality in LLM outputs is well-established, and personality assessment is critically important for responsible deployment of LLMs to the general public.
Since measurements of LLM personality to date have not yet been rigorously validated, 
this work presented a principled methodology for a comprehensive quantitative analysis of personality traits exhibited in personality questionnaire responses and text generated by 18 widely-used LLMs, by applying standards from psychometrics.
We applied the methodology to models of various sizes and conclusively showed that psychometric tests of LLM personality 
demonstrate reliability and construct validity for larger and instruction fine-tuned models. We presented a novel methodology for shaping LLM-synthesized personality along desired dimensions using Goldberg's personality trait markers and Likert-type linguistic qualifiers, to resemble specific personality profiles. Additionally, we discussed the ethical implications of shaping LLM personality traits. This work has important implications for AI alignment and harm mitigation, and informs ethics discussions concerning AI anthropromorphization, personalization, and potential misuse.

\section{Acknowledgements}
We
thank Lucas Dixon, Douglas Eck, and Kathy Meier-Hellstern for their feedback on early versions of this paper. We also thank David Stillwell for facilitating research access to the Apply Magic Sauce API.
Finally, we thank Jason Rentfrow and Neda Safaee-Rad for their advice on personality-related aspects of the paper. G.S-G. is supported by the Bill \& Melinda Gates Foundation through a Gates Cambridge Scholarship [OPP1144]. Inference for open models used compute resources provided by the Cambridge Service for Data Driven Discovery (CSD3) at the University of Cambridge, made possible by Tier-2 funding from the EPSRC (EP/T022159/1) and DiRAC funding from STFC (www.dirac.ac.uk).

\appendix

\section{Large Language Models}
\label{sec:background-llms}
\subsection{Language Modeling}
Language modeling is a fundamental task in natural language processing (NLP).
It is the basis of many solutions to a wide variety of problems involving AI systems with linguistic inputs. Downstream NLP tasks that leverage language models include (among many others):
\begin{itemize}
  \item natural language understanding,
  \item question answering,
  \item machine translation,
  \item document summarization,
  \item dialog systems.
\end{itemize}
The fundamental goal of language modeling is to assign high probabilities to utterances (usually sentences in plain text) that are likely to appear in data (i.e., belong to the language) and low probabilities to strings of words that are not.
A trained language model can then be used to assign probabilities to arbitrary sequences of words.
In the past, this was done by parametric statistical models estimated from data.
However, those models have been replaced with much more successful deep neural network-based methods.
Generally, a modern large language model (LLM) is a neural network taking strings of words as input, and returning a probability measure for each of those strings. The network is trained to correspond to the likelihood that given input strings conform to a particular language, as induced from large quantities of text (often called a corpus).
Normally, instead of thinking of a language model in terms of estimating the joint probability of a string of words, we view it in terms of its ability to predict continuation based on existing context.
A neural language model therefore is usually trained to compute a conditional probability of word $ w_n $ following a sequence of words $ w_1, w_2, \dots, w_{n-1} $.

\subsection{Role of Attention in LLMs}
Recent advances in LLMs and NLP more broadly have been based on 
innovative
uses of various forms of attention in neural networks.
Attention was initially introduced as an improvement to recurrent encoder-decoder architectures \cite{bahdanau2014neural} in the context of neural machine translation systems.
Subsequently, it was discovered that the idea of attention alone can be used as a basis for language modelling systems.
A seminal paper titled ``Attention Is All You Need" \cite{vaswani2017attention} introduced a new type of neural network architecture for extracting deep contextualized text representations from raw natural language data using a process based predominantly on repeated application of the ``self-attention" operation in a model, called the \textit{transformer}.
This kind of model transforms the original vector space representation of linguistic units through a sequence of embedding spaces, where each successive mapping recomputes the representation of every token\footnote{A token is the smallest unit of text that a large language model can process. Tokens can be individual characters, words, or subwords, depending on the specific tokenization method used. The model assigns a unique identifier to each token, and these identifiers are then used to represent the text in the model's internal representations.} in the context of its surrounding tokens.
As such, it allows for the semantics of words as seen by the neural AI systems to vary depending on the context and evolve over time.
Such representations produced significant performance improvements on natural language understanding tasks.
The transformer architecture was composed of two stacks of self-attention blocks forming an encoder-decoder architecture, originally designed as a sequence transducer for neural machine translation.

\subsection{Decoder-only Architecture}
Currently, large language models (LLMs) are usually based on the decoder-only transformer architecture \cite{gpt3, palm, OpenAI2022ChatGPT, OpenAI2023GPT4, touvron2023llama}.
A sequence of text tokens, usually representing a user prompt (e.g., a question) is first tokenized, by splitting text into morpheme-like subwords units using a deterministic algorithm inspired by information theoretic ideas.
This sequence of tokens is then embedded into a high-dimensional vector space where each token becomes a sequence of floating-point numbers.
This initial point-cloud of vectors representing linguistic units of the prompt is then transformed by a sequence of nonlinear mappings between high-dimensional representation spaces.
The final representation is used to compute a probability distribution over possible continuations of text conditioned on the original prompt.
The predominant method of training such models is gradient descent optimization (i.e., the backpropagation algorithm), resulting in representations that are informative towards predicting the contexts in which words appear within the training corpus.
This simple self-supervised criterion leads to emergent abilities of the model, spanning syntax, semantics, and pragmatics of natural language use.
The {\it distributional hypothesis}, which forms a fundamental assumption behind neural language model training, states that syntactic and semantic relationships between words can be inferred from their context, i.e., co-occurrence patterns with other words in the corpus.
As a result, optimizing model parameters based on n-grams of tokens extracted from large quantities of natural language text generates informative representations of linguistic units in submanifolds of high-dimensional real vector spaces.
The geometric and topological features of these induced representation manifolds determine the behavior of LLMs.
The models trained for dialogue, including all models used in our work, are of the \textit{autoregressive} type.
This means that the output from the model itself becomes part of the context on which future outputs are conditioned.
This allows the model to form a contextual memory of the conversation, including its own outputs.

Current state of the art LLMs contain trillions of parameters and are trained on corpora of text (such as books, articles, and websites) and code \cite{commoncrawl, chen2021evaluating} that contain billions of n-gram patterns, allowing them to learn the statistical relationships between words and phrases \cite{wei2022emergent}, and consequently the patterns, structures, and semantics of language \cite{marcus-etal-1993-building, paperno-etal-2016-lambada, merity2017pointer, gao2020pile}. In this work, we primarily explore decoder-only, auto-regressive LLMs such as PaLM \cite{palm}, where the input is usually a partial or complete sequence of tokens, and the model generates the next token in the sequence based on the previous tokens it has seen in an iterative process.

\subsection{Controlling LLM behavior}
There are three main techniques that change or control an LLM's behavior and output with respect to a given input: \textit{pretraining} (training the LLM on a large corpus of text \cite{gpt3, palm, touvron2023llama}), \textit{fine-tuning} (i.e., further training a pretrained LLM on a smaller dataset specific to a particular task or domain \cite{ziegler2020finetuning, wei2022finetuned, OpenAI2022ChatGPT, ouyang2022training}), and \textit{prompting}. While pretraining and fine-tuning affect model behavior by directly altering the model's weight parameters, prompting does so indirectly by influencing the activation of certain neurons or the flow of information through the model's inference process. 

The most significant aspect of using prompts to control LLM behavior is to carefully design or engineer prompts to generate desired outputs from the LLM. Several types of {\it prompt engineering} techniques are commonly used with LLMs. In \textit{few-shot prompting} \cite{gpt3,min2022rethinking, mahabadi2022perfect}, a limited amount of example data are provided to the model in a prompt to guide it to perform a task. By leveraging this small set of examples, the LLM can generalize and produce responses beyond the provided instances. Few-shot prompting relies on the ability to \textit{bias} the LLM's responses based on the input prompt. But because it introduces a bias, this method is not useful in cases where the goal is to probe the default bias of the LLM, the behavior or tendency of the LLM to produce certain outputs (e.g., certain psychometric survey responses, in our case). \textit{Zero-shot prompting} \cite{wei2022finetuned, kojima2023large}, on the other hand, involves instructing the model to generate responses for tasks it has not been specifically trained on and without providing any examples, relying on the LLM's pre-existing knowledge and language understanding acquired during pre-training. This method provides insights into the language priors and distribution
learned by the LLM, what tokens are more correlated than others, etc. For instance, if asked to complete an input prompt: ``She went to see an expert about her stroke, who", an LLM trained on medical domain data is likely to respond ``advised her to get an ECG test." whereas an LLM trained on sports data might complete it as ``coached her about the best techniques from top golf pros." Several recent works in the field of Responsible AI have attempted to uncover latent language biases in LLMs, to identify potential for harm, and to suggest mitigation techniques \cite{liang2021understanding, zamfirescu2023promptdesign}.
Similarly, our work used zero-shot prompt engineering to analyze how latent linguistic features in LLMs give rise to a coherent personality when quantified psychometrically. We further analyzed how those traits can be modified by engineering specific prompts and affecting the latent linguistic features in these LLMs.


\subsection{Modes of Inference in LLMs}
LLMs offer various ways of inference in practice. In \textit{generative} mode, an LLM is given a prompt or instruction, and it then generates text that is consistent with that prompt. This mode is useful for creative text generation tasks, such as story or poetry writing. In \textit{scoring} mode, the LLM is given a pair \textit{(prompt, continuation)} and it assigns a score or probability to it, indicating its quality or relevance or how \textit{likely} it is to be generated from that model. Scoring mode \cite{jiang2021promptingforqa} is often used for tasks like language evaluation \cite{pmlr-v203-jang23a}.
Internally to the LLM, there is a single operating mode---computing the probability distribution over a sequence of tokens---but this distinction between the various modes of inference is conceptually useful when reasoning about model behavior.


\section{Personality Psychology}
\label{app:background-personality-science}
The field of personality psychology defines {\it personality} as enduring characteristics, traits, and patterns that shape thoughts, feelings, and behaviors across a diverse array of situations; e.g., social, spatial, and temporal contexts \cite{roberts2022personalityreview}. Decades of personality research synthesizing evidence from molecular genetics \cite{roberts2018genetics}, evolutionary biology \cite{nettle2006evolution}, neuroscience \cite{deyoung2010neurosciencebigfive, deyoung2022personalityneuroscience}, linguistics \cite{boyd2017personalitylanguage, pennebaker1999linguistics}, and cross-cultural psychology \cite{mccrae2005universal} have reduced such diverse characteristic patterns to a theorized handful of higher-order factors that define personality \cite{deyoung2010bigfivetheory, john2008integrativebigfivetheory}.

Specific to linguistic evidence of a personality taxonomy, a central area of personality research concerns {\it the lexical hypothesis of personality}---that human personality is intrinsically connected to language. Since its origin from Sir Francis Galton in the 1800s \cite{galton1884measurement}, empirical research on the lexical hypothesis has posited that 1) important personality characteristics of a given society will be encoded in its language; and 2) that the most important of those characteristics are likely encoded as single words \cite{goldberg1981personalitylanguage, raad1998languagepersonality, saucier2001lexical}. This empirical framework grounds our work in three areas: the choice of one of our personality instruments (the BFI; described below), our prompts for shaping LLM personality, and the choice of the language-based assessment of personality for rating LLM-synthesized personality in a downstream task.

The Big Five model \cite{john+1999}, the most commonly cited research taxonomy of personality formed through the research described above, identifies five \textit{personality trait dimensions} (i.e., \textit{domains}) and provides methodology to assess these dimensions in humans. The five dimensions are extraversion (EXT), agreeableness (AGR), conscientiousness (CON), neuroticism (NEU), and openness to experience (OPE). Each domain is further composed of various lower-order \textit{facets} nested underneath. 

\section{Related Work}
\label{sec:related-work}


Recent attempts to probe personality and psychopathological traits in LLMs suggest that some models exhibit dark personality patterns \cite{li2023psychopathy}, or demonstrate how to administer personality inventories to LLMs \cite{Pellert2022_ll, karra2023personality, jiang2023personalityllms, song2023personalityllms, caron2022personalityllms, singh2023personalityllms, jiang2023personallm}. Some have also made efforts to induce desired levels of personality in LLMs using prompting \cite{jiang2023personalityllms, caron2022personalityllms, jiang2023personallm} or fine-tuning \cite{karra2023personality, li2023psychopathy}. While these works outlined the utility and importance of measuring social phenomena in LLMs \cite{Pellert2022_ll}, there remains a need to match standards of evaluating the quality of human survey data when evaluating survey response data from LLMs---standards that are commonplace in quantitative social science \cite{clark+2019}.
To claim that scores on a psychological test are trustworthy and meaningful signals of what the test purports to measure, one must establish the test's reliability and construct validity. 

Recent works that probe social and personality-related traits in LLMs have administered and analyzed questionnaires in ways that are unconventional in psychometrics. In this appendix, we focus on three additional elements not discussed in the main text. 

First, researchers have collected LLM responses in the form of open-ended, generated completions, often in dialog mode. 
For instance, recent approaches have administered psychological measures for LLMs in the form of a research interview transcript, where a fictitious researcher posed measure items to a fictitious participant, who was instructed to respond to these items on a numeric scale \cite{tavast2022panas}. Other researchers \cite{wang2024personalityfidelity} rephrased popular personality questionnaires to follow an open-ended role-playing format. In psychometrics, questionnaire-based methods of assessment are distinct from interview-based methods. Human answers to both questionnaires and structured interviews measuring the same underlying construct do not necessarily converge (e.g., in the case of measuring personality disorders \cite{zimmerman1994diagnosingpds}). Indeed, administering questionnaires in this way to LLMs creates an arbitrary viewpoint from which to elicit personality traits, and is likely biased by the ordering of the questionnaire itself \cite{krosnick1987ordereffects} 
and prompting the LLM to respond in an interview setting (where it may respond differently knowing an interviewer is observing). Finally, each LLM response to a given questionnaire item is not an independent event under this implementation, but considers all previous responses shown in the transcript. 

We mitigated potential measurement error stemming from this practice by preserving the exact phrasing and intended format of the psychometric tests we use. We also diversified the viewpoints we use to elicit LLM-synthesized traits through structured prompt wrapping.

Second, many researchers have used popular yet psychometrically unsound tests of personality \cite{wang2024personalityfidelity, pan2023mbti}, most commonly the Myers-Briggs Type Indicator (MBTI). The MBTI is not accepted or used in peer-reviewed personality research due to reliability and validity concerns \cite{stein2019mbtivalidity}.

Third, the LLMs in these studies were not evaluated deterministically. This not only hampers reproducibility, but also poses implications for reliability. 
Computing reliability metrics for questionnaires scored in this unconventional way is precarious because such reliability metrics depend on item-level variance. If this item-level variance is contaminated by variation introduced by the model parameters in a different way for each item, it is difficult to compute valid indices of reliability. We overcame these challenges in our work by proposing a prompt and persona sampling methodology that allows variance to be linked across administrations of different measures.

PsyBORGS \cite{psyborgs-oss} administered a series of validated survey instruments of race-related attitudes and social bias to LLMs using psychometrics-informed prompt engineering. Our work utilized the PsyBORGS framework. 

\section{Evaluated Language Models}
\label{app:models}


We selected open and closed LLMs to represent a variety of model parameter sizes, training methods, and architectures. To explore the effects of instruction tuning, we also prioritized models that had both pretrained and instruction-tuned variants available. Table \ref{tab:results-summary} lists the tested models along with their size and training configuration options.

Starting our study with the \PaLM\ family of models, we focused on three different model sizes: small (8B), medium (62B), and large (540B), because LLM model size is a key determinant of performance for this model family \cite{palm, zhao2023survey}.
Second, 
we investigated \PaLM\ variants fine-tuned to follow instructions, as they have been shown to perform better than base models for prompting-based instruction following 
tasks \cite{wei2022finetuned}. We specifically selected variants fine-tuned with the popular FLAN dataset \cite{wei2022finetuned}. Third, we examined conventional and high-data training methods, known as Chinchilla training \cite{hoffmann2022training}, which uses a fixed training budget to find the balance between model size and training dataset scale. Chinchilla training yields superior performance across a broad set of tasks \cite{hoffmann2022training, zhao2023survey}. 

For replication purposes, we prioritized selection of open models available on HuggingFace using the same criteria. At the time of writing, we selected the 7B, 13B, and 70B versions of \LlamaTwo\ and \LlamaTwoChat\ \cite{touvron2023llama2} to study the effects of size and instruction tuning. The \Mistral\ \cite{jiang2023mistral7b} and \Mixtral\ \cite{jiang2024mixtralexperts} model families (v0.1) were selected to study the effects of instruction tuning and to include a model with a mixture-of-experts architecture. 

Due to their popularity, we also evaluated the \GPT\ family of models, namely 
\GPTThreeDotFiveTurbo\ (\GPTThreeDotFiveTurboZeroOneTwoFive), 
\GPTFourOMini\ (\GPTFourOMiniTwentyTwentyFourZeroSevenEighteen), and \GPTFourO\ (\GPTFourOTwentyTwentyFourZeroEightZeroSix) \cite{openai2024gpt4ocard}, the only models from OpenAI of the same family with publicly-disclosed size differences. Unintentionally, this added models with multi-modal capabilities.


All \PaLM\ experiments used quantized models \cite{yao2023comprehensive} to reduce the memory footprint and speed up inference time. All open models were tested at full-precision by optimizing inference throughput and memory with the vLLM library \cite{kwon2023vllm}. We do not know the quantization status of the \GPT\ endpoints called for this project, but provide the dated model snapshot IDs used above for reproducibility.


\section{Simulating LLM Responses} 
\label{app:methods-llm-scoring}
For tested variants of \PaLM, we had direct access to the log-likelihood scores of possible continuations for a given prompt, which made scoring items by ranking the conditional probabilities of their response scale options relatively straightforward. 

For all other models, where the ability to access next-token log-likelihood data varied widely, we relied on constrained decoding to preserve this same choice selection logic while bypassing the need for raw log-likelihood scores. This was implemented using the Outlines library \cite{willard2023outlines}, which we set to restrict models to generate the most likely response to an item from a restricted set of the item's response scale options (e.g., [``1", ``2", ``3",``4", ``5"]).

\section{Selected Personality Inventories}
\label{app:methods-measure-selection}

\begin{table*}[tbp]
\caption{\small Item Postambles used to construct the prompts employed in the experiments to generate LLM-simulated survey responses. All administered measures used a Likert-type response scale that allowed 5 possible choices, with the exception of the PVQ-RR, which used a 6-point response scale. 
Item Postambles 1--5 were used for the BFI; 6--10 for the IPIP-NEO; 11--15 for the PANAS; 16--20 for the SSCS; 21--25 for the BPAQ; and 26--30 for the PVQ-RR.
}
\label{appendix:tab:postambles}
\centering
\footnotesize
\begin{tabular}{|p{0.02\linewidth}|p{0.92\linewidth}|}
\toprule
№ & Item Postambles               \\ \midrule
\rownum & \texttt{please indicate the extent to which you agree or disagree on a scale from 1 to 5 (where 1 = "disagree strongly", 2 = "disagree a little", 3 = "neither agree nor disagree", 4 = "agree a little", and 5 = "agree strongly"):"}     \\
\rownum & \texttt{please rate your level of agreement on a scale from 1 to 5 (where 1 = "disagree strongly", 2 = "disagree a little", 3 = "neither agree nor disagree", 4 = "agree a little", and 5 = "agree strongly"):"}     \\
\rownum & \texttt{please rate your level of agreement or disagreement on a scale from 1 to 5 (where 1 = "disagree strongly", 2 = "disagree a little", 3 = "neither agree nor disagree", 4 = "agree a little", and 5 = "agree strongly"):"}     \\
\rownum & \texttt{please rate how much you agree on a scale from 1 to 5 (where 1 = "disagree strongly", 2 = "disagree a little", 3 = "neither agree nor disagree", 4 = "agree a little", and 5 = "agree strongly"):"}     \\
\rownum & \texttt{please rate how much you agree or disagree on a scale from 1 to 5 (where 1 = "disagree strongly", 2 = "disagree a little", 3 = "neither agree nor disagree", 4 = "agree a little", and 5 = "agree strongly"):"}     \\
\rownum & \texttt{please rate how accurately this describes you a scale from 1 to 5 (where 1 = "very inaccurate", 2 = "moderately inaccurate", 3 = "neither accurate nor inaccurate", 4 = "moderately accurate", and 5 = "very accurate"):"}     \\
\rownum & \texttt{please indicate how accurate this is about you on a scale from 1 to 5 (where 1 = "very inaccurate", 2 = "moderately inaccurate", 3 = "neither accurate nor inaccurate", 4 = "moderately accurate", and 5 = "very accurate"):"}     \\
\rownum & \texttt{please indicate how accurate or inaccurate this is about you on a scale from 1 to 5 (where 1 = "very inaccurate", 2 = "moderately inaccurate", 3 = "neither accurate nor inaccurate", 4 = "moderately accurate", and 5 = "very accurate"):"}     \\
\rownum & \texttt{please rate how accurate this is about you on a scale from 1 to 5 (where 1 = "very inaccurate", 2 = "moderately inaccurate", 3 = "neither accurate nor inaccurate", 4 = "moderately accurate", and 5 = "very accurate"):"}     \\
\rownum & \texttt{please rate how accurate or inaccurate this is about you on a scale from 1 to 5 (where 1 = "very inaccurate", 2 = "moderately inaccurate", 3 = "neither accurate nor inaccurate", 4 = "moderately accurate", and 5 = "very accurate"):"}     \\
\rownum & \texttt{indicate to what extent you agree on a scale from 1 to 5 (where 1 = "very slightly or not at all agree", 2 = "agree a little", 3 = "agree moderately", 4 = "agree quite a bit", and 5 = "agree extremely"):"}     \\
\rownum & \texttt{please rate your level of agreement on a scale from 1 to 5, (where 1 = "very slightly or not at all agree", 2 = "agree a little", 3 = "agree moderately", 4 = "agree quite a bit"}     \\
\rownum & \texttt{please rate your level of agreement or disagreement on a scale from 1 to 5 (where 1 = "very slightly or not at all agree", 2 = "agree a little", 3 = "agree moderately", 4 = "agree quite a bit", and 5 = "agree extremely"):"}     \\
\rownum & \texttt{please rate how much you agree on a scale from 1 to 5 (where 1 = "very slightly or not at all agree", 2 = "agree a little", 3 = "agree moderately", 4 = "agree quite a bit", and 5 = "agree extremely"):"}     \\
\rownum & \texttt{please rate how much you agree or disagree on a scale from 1 to 5 (where 1 = "very slightly or not at all agree", 2 = "agree a little", 3 = "agree moderately", 4 = "agree quite a bit", and 5 = "agree extremely"):"}     \\
\rownum & \texttt{please decide to what extent this describes you on a scale from 1 to 5 (where 1 = "strongly disagree", 2 = "disagree", 3 = "neither agree nor disagree", 4 = "agree", 5 = "strongly agree"):"}     \\
\rownum & \texttt{please rate your level of agreement on a scale from 1 to 5 (where 1 = "strongly disagree", 2 = "disagree", 3 = "neither agree nor disagree", 4 = "agree", 5 = "strongly agree"):"}     \\
\rownum & \texttt{please rate your level of agreement or disagreement on a scale from 1 to 5 (where 1 = "strongly disagree", 2 = "disagree", 3 = "neither agree nor disagree", 4 = "agree", 5 = "strongly agree"):"}     \\
\rownum & \texttt{please rate how much you agree that this describes you on a scale from 1 to 5 (where 1 = "strongly disagree", 2 = "disagree", 3 = "neither agree nor disagree", 4 = "agree", 5 = "strongly agree"):"}     \\
\rownum & \texttt{please rate how much you agree or disagree that this describes you on a scale from 1 to 5 (where 1 = "strongly disagree", 2 = "disagree", 3 = "neither agree nor disagree", 4 = "agree", 5 = "strongly agree"):"}     \\
\rownum & \texttt{rate how characteristic this is of you on a scale from 1 to 5 (where 1 = "extremely uncharacteristic of me", 2 = "uncharacteristic of me", 3 = "neither characteristic nor uncharacteristic of me", 4 = "characteristic of me", and 5 = "extremely characteristic of me"):"}     \\
\rownum & \texttt{please rate how characteristic this is of you on a scale from 1 to 5 (where 1 = "extremely uncharacteristic of me", 2 = "uncharacteristic of me", 3 = "neither characteristic nor uncharacteristic of me", 4 = "characteristic of me", and 5 = "extremely characteristic of me"):"}     \\
\rownum & \texttt{please rate how characteristic or uncharacteristic this is of you on a scale from 1 to 5 (where 1 = "extremely uncharacteristic of me", 2 = "uncharacteristic of me", 3 = "neither characteristic nor uncharacteristic of me", 4 = "characteristic of me", and 5 = "extremely characteristic of me"):"}     \\
\rownum & \texttt{please indicate to what extent this is characteristic of you on a scale from 1 to 5 (where 1 = "extremely uncharacteristic of me", 2 = "uncharacteristic of me", 3 = "neither characteristic nor uncharacteristic of me", 4 = "characteristic of me", and 5 = "extremely characteristic of me"):"}     \\
\rownum & \texttt{please indicate to what extent this is characteristic or uncharacteristic of you on a scale from 1 to 5 (where 1 = "extremely uncharacteristic of me", 2 = "uncharacteristic of me", 3 = "neither characteristic nor uncharacteristic of me", 4 = "characteristic of me", and 5 = "extremely characteristic of me"):"}     \\
\rownum & \texttt{think about how much that person is or is not like you. Rate how much the person described is like you on a scale from 1 to 6 (where 1 = "not like me at all", 2 = "not like me", 3 = "a little like me", 4 = "moderately like me", 5 = "like me", and 6 = "very much like me"):"}     \\
\rownum & \texttt{please rate how characteristic this is of you on a scale from 1 to 6 (where 1 = "not like me at all", 2 = "not like me", 3 = "a little like me", 4 = "moderately like me", 5 = "like me", and 6 = "very much like me"):"}     \\
\rownum & \texttt{please rate how characteristic or uncharacteristic this is of you on a scale from 1 to 6 (where 1 = "not like me at all", 2 = "not like me", 3 = "a little like me", 4 = "moderately like me", 5 = "like me", and 6 = "very much like me"):"}     \\
\rownum & \texttt{please indicate to what extent this is like you on a scale from 1 to 6 (where 1 = "not like me at all", 2 = "not like me", 3 = "a little like me", 4 = "moderately like me", 5 = "like me", and 6 = "very much like me"):"}     \\
\rownum & \texttt{please indicate to what extent this is or is not like you on a scale from 1 to 6 (where 1 = "not like me at all", 2 = "not like me", 3 = "a little like me", 4 = "moderately like me", 5 = "like me", and 6 = "very much like me"):"}     \\ 
\botrule
\end{tabular}
\end{table*}

\begin{table*}[tbp]
\caption{\small 50 human Biographic Descriptions sampled from the PersonaChat dataset \cite{zhang2018personalizing}, used in Item Preambles across all experiments.}
\label{appendix:tab:personachat}
\centering
\footnotesize
\begin{tabular}{p{0.95\linewidth}}
\toprule
\textbf{Biographic Descriptions }              \\ \midrule
I like to garden. I like photography. I love traveling. I like to bake pies.  \\
I've a beard. I graduated high school. I like rap music. I live on a farm. I drive a truck. \\
I blog about salt water aquarium ownership. I still love to line dry my clothes. I'm allergic to peanuts. I'll one day own a ferret. My mom raised me by herself and taught me to play baseball. \\
Since young I ve loved to cook. I auditionated in a cooking show. I think I've talent for it. I took classes while growing up.  \\
My name is tom. I try to watch what I eat. I enjoy eating italian food. Pizza is my favorite. I am east asian.  \\
I live by a lake. I am a mother. I own a custom upholstery shop. I'm a wife.  \\
I enjoy working out and learning new things. I'm a student in college. I'm studying software development. I play the guitar.  \\
I've three dogs at home. I hate to workout, but I need to. I am very good at the drums. I have a bicycle. I need to take my blood sugar everyday.  \\
I work in advertising. My mother is dead. I like to hike. I've a golden retriever. I write fiction for fun.  \\
I can never decide between a chili corn dog and a cheesy hot dog. I drive more than an hour each way to work. I prefer the night to the day, but I love sunshine. I am a grandparent at 44.  \\
I like to smell my own farts. My beer gut is so huge i'ven T seen my feet in two years. I am from San Fransico. I am always the one who buys the beers. I like to place blame on other people even when I know it is my fault.  \\
I lived most of my life not knowing who Bob marley was. When I cut loose, I lose control. We help each other out in my family. I despise my boss. I work over 60 hours a week as a restaurant manager.  \\
I prefer the simpler times. I like simple jokes. Some jokes go too far. I like the flintstones.  \\
It is my universe, and everyone else is just a character in it. I work as a dental assistant in a ritzy part of town. I've borderline personality disorder. At night, I party hard in the Atlanta club scene, and I never miss a music festival.  \\
I watch a lot of tv. I live alone. My favorite food is a cheeseburger. I enjoy fishing. I work on cars for a living.  \\
I'm an animal rights activist. I hope to retire to Florida. I played in a band for 17 years. My mother and father are both in the church choir.  \\
I've taken formal music lessons since I was 5. I'm a musician. My best friend is in a band with me. I wish I could spend more time at home.  \\
I grew up in Kentucky. I'm a veteran. My favorite book is ender's game. I have a garden. I like to read.  \\
I am a vegan. I love country music. I love the beach. I like to read.  \\
I've depression and anxiety so I don't really go out a lot. I work at home, editing. I have a cat. I hope to move out soon.  \\
My favorite food is mushroom ravioli. I ve never met my father. My mother works at a bank. I work in an animal shelter.  \\
I love kids and dogs. I like to go shopping with my daughters. I like to cook. I love to chat with my friends.  \\
I swim often. I run track. I wear glasses all day. I take medication.  \\
I like to go on long hikes. I like to play volleyball. I like to come up with new hairstyles. I like to do my nails.  \\
I watch Jimmy Fallon s show every night. I have never kissed a woman. People notice how organized I am. I believe that I can achieve anything.  \\
I drive a lifted Chevy truck. I played football in high school. I am a roofer. I always have a beer after work.  \\
I love animals. My father worked for Ge. Green is my favorite color. I enjoy playing tennis. I'm an aspiring singer.  \\
I try to watch what I eat. I enjoy eating italian food. Pizza is my favorite. My name is tom. I am east asian.  \\
In allergic to peanuts. I like eating vegetables. I love the Beatles. I'm usually very shy. I have trouble getting along with family.  \\
I go to high school. Math is my favorite subject. I live in the United States. I am a boy.  \\
I have a job as an it agent. I like smoking weed. My dad works for stifle. I love rap music. I'm a meataholic.  \\
I work in tv. I do not treat my girlfriend very well. I like to cook breakfast on sundays. I love to sing. I am a lesbian.  \\
I work on semi trucks for a living. My father was a driver himself. I got off the road when I married my sweetheart. I want to take her on vacations one day. My motor never stops running.  \\
I own a Iphone 7. I drink hot chocolate during the winter. I'm allergic to seafood. My mother use to read me bed time stories.  \\
I am eighteen years old. I'm going to majoring in business. I just bought my first car. I received a full scholarship to Florida state university.  \\
I live in a tiny house to save money. I collect single malt scotch. I listen to blues and jazz. I tend bar on the weekends. During the week I go to college to become a lawyer.  \\
I love to go horseback riding whenever I can. I'm a mother of two beautiful boys. My family and I go camping every month. My favorite artist is Justin Bieber.  \\
I especially enjoy listening to the band the lumineers. I enjoy reading and walking on sunny days. I'm a happy person. I sing many songs.  \\
I play piano. My favorite color is yellow. My boyfriend is in the army. My father is dead. My hair is short.  \\
I'm a mother. I'm a nurse at a hospital. My favorite band is the rolling stones. I love to read and cook. My favorite food is mexican food.  \\
I deliver baked goods in the state where I live. My favorite hobby is playing recreational baseball. I spend my weekends camping. I'm a truck driver. My wife and two kids camp with me.  \\
I am argentinian. I like to wear boots. I have many girlfriends. I like to eat beef. I like to ride horses.  \\
I recently had a private lunch with will ferrell. I am trying to become a male model in hollywood. I'm a huge fan of classical jazz. I am on a low carb diet.  \\
I want to put my photos to a music video staring Adam Levin. I want to travel the world taking photographs of my travels. I am a widow. I want to be a famous photographer.  \\
I am in the army. I fly airplanes. I enjoy building computers. I dropped out of college.  \\
I have three children. I live in the suburbs of a major city. I like to garden. I graduated college for secondary english education.  \\
I play guitar in the local band. I live on a small farm in Ohio. I am the youngest of three brothers. I have never been to the city.  \\
I'm a widow. I want to put my photos to a music video staring Adam Levin. I want to travel the world taking photographs of my travels. I want to be a famous photographer. I like taking pictures.  \\
I still live at home with my parents. I play video games all day. I'm 32. I eat all take out.  \\
My friend once bought me a car. I am disabled and cannot walk. I take vitamin c when I have a cold. I do not eat bread. My favorite season is winter. \\
\botrule
\end{tabular}
\end{table*}

To measure personality, we selected two well-established psychometric measures to assess the Big Five taxonomy: one from the lexical tradition and one from the questionnaire tradition. \textit{Lexical tradition} measures are grounded in the hypothesis that personality can be  captured by the adjectives found in a given language \cite{galton1884measurement, goldberg1981personalitylanguage}, while \textit{questionnaire tradition} measures are developed with existing (and not necessarily lexical) taxonomies of personality in mind \cite{simms2017ffm}. Lexical measures may be better suited for LLMs because they are language-based and rely on adjectival descriptions. We posit that questionnaire measures, which do not rely on trait adjectives for content, more conservatively test LLM abilities, as they are less abstract and more contextualized. Our work focused on Big Five measures of personality due to the Big Five's integrative robustness and cross-theory convergence in the human personality and psycholinguistics literature \cite{simms2017ffm}.

Our primary personality measure, the \text{IPIP-NEO} \cite{goldberg1999ipip}, is a 300-item open source representation of the commercialized Revised NEO Personality Inventory 
\cite{costa1992neo}. The IPIP-NEO, hailing from the questionnaire tradition \cite{simms2017ffm}, involves rating descriptive statements (e.g., ``[I] prefer variety to routine"; 60 per Big Five domain) on a 5-point Likert scale. (1 = \textit{very inaccurate}; 2 = \textit{moderately inaccurate}; 3 = \textit{neither accurate nor inaccurate}; 4 = \textit{moderately accurate}; 5 = \textit{very accurate}). We refer to these statements as \textit{items}.
The IPIP-NEO has been translated and validated in many languages, facilitating cross-cultural research across populations \cite{10.1002/per.2260}, and has been used in longitudinal studies to assess personality change and stability over time \cite{young2011ipiplongitudinal}. We chose this measure for its excellent psychometric properties, shown in \cite{goldberg1999ipip}.

As a robustness check and to assess convergent validity, we also measured LLM-synthesized personality using the Big Five Inventory (BFI) 
\cite{john+1999}. Developed in the lexical tradition, the BFI is a brief (44-item), adjectival statement-based measure of the broad Big Five traits. The BFI asks participants to rate short descriptive statements (e.g., “I see myself as someone who is talkative”) also on a 5-point Likert scale. The resulting summary scores indicating levels of Big Five trait domains range from 1.00 to 5.00. In the psychology literature \cite{simms2017ffm}, the BFI has demonstrated excellent reliability (mean $\alpha$ reported across domain subscales = 0.83), convergent validity, and external validity.

Domain subscale scores across both measures were calculated following their original instructions as the average of item response values, accounting for reverse-keyed items. Possible subscale scores ranged from 1.00 to 5.00, indicating the lowest and highest possible levels of a given Big Five domain, respectively.

\section{Simulating Population Variance Through Prompting}
\label{sec:method-general-prompt-design}
It was empirically necessary to introduce controlled variation in LLM-simulated survey data to assess their reliability and statistical relationships with outcomes of interest; in short, controlled variation was required to statistically test for reliability and construct validity. 

For instance, an \textit{Item Postamble} presented the possible standardized responses the model can choose from, e.g., 

\texttt{please rate your agreement on a scale from 1 to 5, where 1 is ‘strongly disagree’, 2 is ‘disagree’, 3 is ‘neither agree nor disagree’, 4 is ‘agree’, and 5 is ‘strongly agree’.}

We customized five variations of Item Postambles for each administered measure, such that all five variations would have parallel meanings across measures. Supplemental Table \ref{appendix:tab:postambles} lists all Item Postambles used
in this work. This prompt design enabled thousands of variations of input prompts that could be tested, with two major advantages. First, variance in psychometric test responses created by unique combinations of the Biographic Descriptions (see Supplemental Table \ref{appendix:tab:personachat}), Item Instructions (see Supplemental Table \ref{app:tab:item-instructions}), and Item Postambles enabled us to quantify the validity of personality measurements in LLMs. Unlike single point estimates of personality, or even multiple estimates generated from random resampling of LLMs, diverse distributions of personality scores conditioned on reproducible personas make it possible to compute correlations between convergent personality measures and external, personality-related constructs. Second, variance in Item Preambles and Postambles facilitated a built-in robustness check: it was critical to know if personality scores remained reliable and valid across modifications of context and instructions surrounding original test items. They were indeed reliable and valid for three of the five models tested.

\begin{table}[tb]
\setlength\extrarowheight{2pt} 
\caption{
\small Item Instructions used in Item Preambles across experiments to generate LLM-simulated survey responses.
}
\label{app:tab:item-instructions}
\centering
\footnotesize
\begin{tabular}{p{0.9\linewidth}}
\toprule
Item Instructions               \\ \midrule
\texttt{Considering the statement,} \\
\texttt{Thinking about the statement,} \\
\texttt{Reflecting on the statement,} \\
\texttt{Evaluating the statement,} \\
\texttt{Regarding the statement,} \\
\botrule
\end{tabular}
\end{table}

\section{Psychometrics}
\label{app:background-psychometrics}
\textit{Psychometrics}, a quantitative subfield of psychology and education science, encompasses the statistical theory and technique of measuring unobservable, latent phenomena
called \textit{constructs}, like personality, intelligence, and moral ideology. Psychometrics is foundational to the development and validation of standardized educational tests (e.g., the SAT, LSAT, GRE) \cite{American_Educational_Research_Association2014}, medical and psychological clinical assessments \cite{Wechsler1946}, and large-scale public opinion polls \cite{10.1002/9781118489772.ch28}.

\textit{Psychometric tests} (e.g., survey instruments, measures, multi-item scales) are tools for quantifying latent psychological constructs like personality. Psychometric tests enable statistical modeling of the true levels of unobservable target constructs by relying on multiple indirect, yet observable, measurements across a sample of individuals drawn from a wider population. 

We refer to \textit{items} as the individual elements (i.e., descriptive statements, sometimes questions) used within a psychometric test designed to measure attributes or characteristics of a construct. Items are usually rated on a \textit{rating scale}- a standardized set of response choices that allows researchers to quantify subjective phenomena. A Likert-type scale is the most common rating scale that has respondents specify their level of agreement on a symmetric agree-disagree scale \cite{likert1932technique}. We refer to a \textit{subscale} as a collection of items, usually resulting from a factor analysis, aimed at measuring a single psychological construct. \textit{Measures} are themed collections of subscales. 


For example, the Big Five Inventory (BFI) \cite{john+1999} is a popular measure of personality; it comprises five multi-item subscales targeting each Big Five dimension. BFI Extraversion, for instance, is a subscale within the BFI specifically targeting the dimension of extraversion. An example item under BFI Extraversion would read, ``[I see myself as someone who] is talkative." Participants rate their agreement with this item using the following 5-point Likert-type rating scale: 1 = \textit{disagree strongly}; 2 = \textit{disagree a little}; 3 = \textit{neither agree nor disagree}; 4 = \textit{agree a little}; 5 = \textit{agree strongly}.

How do we know that psychometric tests measure what they claim to measure, i.e., \textit{how do we establish the reliability, accuracy, and utility of the measures of personality, and the constructs assessed in those measures}? Validated scientific frameworks for establishing the \textit{reliability} and {\it construct validity} of a new psychometric test \cite{clark+1995, messick1995, clark+2019} incorporate (but are not limited to) the following overarching standards:
\begin{itemize}
    \item \textbf{Reliability:} \textit{Are test measurements dependable and consistent?} In psychometrics, a test's reliability can be established in terms of internal consistency and factor saturation.
    \begin{itemize}
        \item \textbf{Internal consistency reliability:} \textit{Is the test reliable across multiple measurements (i.e., its items)? In other words, do responses to the test's items form consistent patterns? Are test items correlated with each other?}
        \item \textbf{Factor saturation:} \textit{Do the test's items reflect the variance of one underlying factor or construct?}
    \end{itemize}
    \item \textbf{Construct Validity:} \textit{Do the test measurements actually reflect the underlying construct?} This can be established by checking for convergent validity, discriminant validity and criterion validity.
    \begin{itemize}
        \item \textbf{Convergent Validity:} \textit{Does the test correlate with purported indicators (i.e., convergent tests) of the same or similar psychological construct? These correlations are called \textit{convergent correlations}.} 
        \item \textbf{Discriminant Validity:} \textit{Relative to their convergent correlations, are test scores relatively uncorrelated with scores on theoretically unrelated tests? These correlations are called \textit{discriminant correlations}.}
        \item \textbf{Criterion Validity:} \textit{Does the test correlate with theoretically-related, non-tested phenomena or outcomes?}
    \end{itemize}

\end{itemize}

\subsection{Reliability: Are Measurements Dependable?}
\label{app:methods-reliability}
The hallmark characteristic of a good psychometric test (or any empirical measure) of a target construct is its reliability, which reflects
its ability to ``measure one thing (i.e., the target construct) and \textit{only} that thing, as precisely as possible" \cite{clark+2019}.  
In this work, we balance our evaluations of reliability across three indices of reliability---Cronbach's Alpha ($\alpha$), Guttman's Lambda 6 ($\lambda_6$), and McDonald's Omega ($\omega$)---weighing the pros and cons of each.


$\alpha$, the most widely-known measure of internal consistency reliability, captures how responses to each item of a scale correlate with the total score of that scale \cite{cronbach1951alpha}. However, $\alpha$ has many documented limitations. For instance, it relies on the assumption that all items of a test measure the same underlying construct and it can be artificially inflated by a test's number of items \cite{zinbarg2005alphaomega}.
Cronbach's $\alpha$ is computed as follows:
\begin{equation}
\label{eq:alpha}
    \alpha = \frac{k}{k-1} \left(1 - {\frac{\sum_{i=1}^k \sigma^2_y}{\sigma^2_x}} \right)
\end{equation}
where $k$ is the number of items on the test, $\sigma_y^2$ is the variance associated with each item $i$, and $\sigma_x^2$ is the overall variance of total scores.

In contrast to $\alpha$, $\lambda_6$ evaluates the variance of each item that can be captured by a multiple regression of all other items \cite{guttman1945g6}. It is less biased alternative to $\alpha$ because it is not affected by item differences in variance, although it is also biased by the number of items on a test.
Guttman's $\lambda_6$ is calculated as:
\begin{equation}
\label{eq:lambda}
\lambda_6 = 1 - \frac{\sum_{i=1}^k(e_i^2)}{V_x}
\end{equation}
where $k$ is the number of items on the test, $e_i$ is the error term for item $i$, $V_x$ is the variance of the total test score.

To test more robustly for reliability (in terms of how well a test measures one underlying factor or construct) in a way that is unaffected by number of items on a test, psychometricians compute McDonald's Omega ($\omega$) \cite{mcdonald1999omega, zinbarg2005alphaomega}. This metric is generally considered a less biased composite test of reliability \cite{zinbarg2005alphaomega, goodboy2020omega}. McDonald's $\omega$ uses confirmatory factor analysis to determine if items statistically form a single factor, or actually measure separate factors. It is calculated as:
\begin{equation}
\label{eq:omega}
\omega_h = \frac{\frac{1}{k}\sum_{i=1}^k\frac{t_i^2}{\sigma^2_i}}{\frac{1}{k-1}\sum_{i=1}^k\frac{t_i^2}{\sigma^2_i}-\frac{1}{k}\frac{1}{1-r_{tt}^2}}
\end{equation}
where $\omega_h$ is McDonald's hierarchical omega, $k$ is the number of items on the test, $t_i$ is the standardized item score for item $i$,
$\sigma^2_i$ is the variance of the standardized item score for item $i$, and $r_{tt}$ is the correlation between the total test score and the standardized total test score.

\subsection{Construct Validity: Are Measurements Meaningful?}
\label{app:background-construct-validity}



Since psychometric tests measure physically unobservable constructs, such as personality traits, it is imperative to establish that such tests measure what they claim to measure. This process is called establishing a test's \textit{construct validity}. \textit{Construct validity} is a comprehensive judgement of how the scores and the theoretical rationale of a test reasonably reflect the underlying construct the test intends to measure \cite{messick1998testvalidity}. Recently, construct validity has become a crucial focus of AI responsibility and governance \cite{jacobs2021measurement, mokander2023auditing}: operationalizing social phenomena in algorithmic systems in a principled way (e.g., through construct validation) is a core part of responsible AI. Bringing empirical rigor to the measurement of social constructs helps stakeholders make more informed judgments of characteristics that may be fair or harmful in AI systems. For instance, if low agreeableness is harmful in AI systems, we need a principled way to measure it.







There is extant work on establishing the validity of measurements of personality as a theoretical construct \cite{roberts2022personalityreview, deyoung2010bigfivetheory, john2008integrativebigfivetheory}, a powerful predictor of other important human traits and life outcomes \cite{roberts2007personalityoutcomes, bleidorn2019policy, kotov2010personalitypsychopathology}
and its manifestation in human language \cite{goldberg1981personalitylanguage, raad1998languagepersonality, saucier2001lexical}, which forms the basis of LLMs.
However, establishing the validity of measurements of personality as a meaningful construct in LLMs has not yet been addressed.

\textbf{Convergent and Discriminant Validity:}
In psychometrics, the convergent and discriminant validity of a test are evaluated using Campbell's classic framework \cite{campbell1959mtmm}, where a test's convergent validity is established by ``sufficiently large" correlations with separate tests meant to measure the same target construct. For example, to validate a new test measuring depression, one could calculate the test's convergent correlations with the Beck Depression Inventory (BDI) \cite{beck1988bdi}---a widely-used measure of depression. To evaluate the discriminant validity of a test, psychometricians commonly gauge the extent to which the test's convergent correlations are stronger than its discriminant correlations---its correlations with orthogonal or less related constructs. As a concrete example, a new test of depression should correlate more strongly with the BDI than with, say, a test measuring English proficiency. 

\textbf{Criterion Validity:}
A common way to assess the criterion validity of a new psychometric test is to check its correlations with theoretically related external (non-test) criteria (hence the name, criterion validity) \cite{clark+2019}. For example, to validate a new psychometric test of depression, one could test if it is substantially related to a known external criterion, such as negative affect.

\textbf{Structural Validity:}
Structural validity encompasses the extent to which, during the initial test construction process, a test's internal structure (i.e., relationships between its items) maps onto the external structure of its target trait (i.e., relationships between nontest observations of the trait). Additionally, structural validity signals that a test's items indeed reflect the latent variance of the trait \cite{clark+2019}. When creating a new psychometric test, psychometricians often use internal-consistency-based analyses to evaluate if the statistical relationships between test items reflect the structure of the test's target construct. Factor analysis is the most common of these methods used to identify and refine dimensions as the basis for scale creation.


\section{Methods for Constructing the Validity of LLM Personality Test Scores}
\label{app:methods-construct-validity}

\paragraph{Establishing Reliability}
\begin{table*}[tbh!]
    \setlength\extrarowheight{2pt} 
    \caption{\small Criterion validity subscales per tested Big Five domain. PANAS = Positive and Negative Affect Schedule Scales; BPAQ = Buss-Perry Aggression Questionnaire; PVQ-RR = Revised Portrait Values Questionnaire; SCSS = Short Scale of Creative Self.}
    \label{tab:external-criteria-measures}
    \small
    \centering
    \begin{tabular}{@{} lll @{}}
    \toprule
    \multicolumn{1}{l}{IPIP-NEO Domain} &   External Criterion                      &   \multicolumn{1}{l}{Criterion Subscales}  \\
    \midrule
    \multirow{2}{*}{Extraversion}        &   \multirow{2}{*}{Trait Emotion}          &   PANAS Positive Affect \\
                                        &                                           &   PANAS Negative Affect \\
    \hline
    \multirow{4}{*}{Agreeableness}       &   \multirow{4}{*}{Aggression}              &   BPAQ Physical Aggression \\
                                        &                                           &   BPAQ Verbal Aggression \\
                                        &                                           &   BPAQ Anger \\
                                        &                                           &   BPAQ Hostility \\
    \hline
    \multirow{3}{*}{Conscientiousness}   &   \multirow{3}{*}{Human Values}            &   PVQ-RR Achievement \\
                                        &                                           &   PVQ-RR Conformity \\
                                        &                                           &   PVQ-RR Security \\
    \hline
    \multirow{2}{*}{Neuroticism}         &   \multirow{2}{*}{Trait Emotion}           &   PANAS Negative Affect \\
                                        &                                           &   PANAS Positive Affect \\
    \hline
    \multirow{2}{*}{Openness}            &   \multirow{2}{*}{Creativity}              &   SSCS Creative Self-Efficacy \\
                                        &                                           &   SSCS Creative Personal Identity \\
    \botrule
    \end{tabular}
    
\end{table*}

In LLM research, model responses to a series of seemingly related tasks intended to measure one latent construct may be anecdotally ``consistent" \cite{Pellert2022_ll, karra2023personality} or inconsistent \cite{miotto2022personality}. Qualitative, descriptive accounts of consistency, however, is not sufficient evidence that the responses to those tasks are statistically reliable 
reflections of the latent constructs they target (as described in Section \ref{app:background-construct-validity}). 


To establish internal consistency reliability, we computed 
Cronbach's $\alpha$ \eqref{eq:alpha} and Guttman's $\lambda_6$ \eqref{eq:lambda} on all IPIP-NEO and BFI subscales. 
To assess more complete composite reliability we computed 
McDonald’s $\omega$ \eqref{eq:omega} on all IPIP-NEO and BFI subscales.


We designated a given reliability metric (\textit{RM}; i.e., $\alpha$, $\lambda_6$, $\omega$) $< 0.50$ as unacceptable, $0.50 \leq RM < 0.60$ as poor, $0.60 \leq RM < 0.70$ as questionable, $0.70 \leq RM < 0.80$ as acceptable, $0.80 \leq RM < 0.90$ as good, and $RM \geq 0.90$ as excellent. High levels of singular internal consistency metrics like $\alpha$ are necessary but not sufficient conditions for demonstrating complete reliability. Therefore, for the purpose of the current work, $\alpha$, $\lambda_6$, \textbf{and} $\omega$ must be at least $0.70$ for a given subscale to be deemed acceptably reliable.

\paragraph{Establishing Construct Validity}
\label{app:results-construct-validity}
We operationalize construct validity in terms of convergent, discriminant, and criterion validity (as defined in Appendix \ref{app:background-construct-validity}). As a supplement, we also report an exploratory analysis of structural validity. We used Campbell's classic multitrait-multimethod matrix (MTMM) \cite{campbell1959mtmm} approach to evaluate convergent and discriminant validity. Criterion validity is evaluated by correlating LLM-simulated personality test data with LLM responses to theoretically-related psychometric test.

\textbf{Convergent validity:}  We evaluated convergent validity---how much our primary test of personality (the IPIP-NEO) positively relates to another purported test of personality (BFI)---by computing bivariate Pearson correlations between IPIP-NEO and BFI scores for extraversion, agreeableness, conscientiousness, neuroticism, and openness and comparing them to ensure correlations between equivalent test subscales are the strongest of their row and column, as outlined in \cite{campbell1959mtmm}. For instance, IPIP-NEO Extraversion should be most correlated with BFI Extraversion, because these two subscales are expected to convergently measure the same underlying construct.

We operationalize convergent correlations between two psychometric tests (in this case, Big Five subscales from the IPIP-NEO and BFI) $\left\{ (x_1,y_1),\ldots,(x_n,y_n) \right\}$, reflecting $n$ pairs of continuous score data, as Pearson product-moment correlations:
\begin{equation}
\label{eq:pearson}
  r_x{}_y =
  \frac{ \sum_{i=1}^{n}(x_i-\bar{x})(y_i-\bar{y}) }{%
        \sqrt{\sum_{i=1}^{n}(x_i-\bar{x})^2}\sqrt{\sum_{i=1}^{n}(y_i-\bar{y})^2}}
\end{equation}
where
$n$ is the sample size,
$x_i, y_i$ are a pair of data points $i$ from sample,
$\bar{x}$ is the sample mean score for personality trait $x$ of the IPIP-NEO, and
$\bar{y}$ is the sample mean score for corresponding personality trait $y$ of the BFI.



In the resulting MTMM, we consider at least strong correlations ($|r_x{}_y| \geq 0.60$; \cite{evans1996}) between each IPIP-NEO domain subscale and its BFI domain scale counterpart (e.g., $r$(IPIP-NEO Extraversion, BFI Extraversion), $r$(IPIP-NEO Agreeableness, BFI Agreeableness), etc.) as evidence of convergent validity. 
For these and following results, we used cut-offs recommended by \cite{evans1996} for considering correlations as moderate, strong, and very strong (viz. $.40 \leq |r| < .60$; $.60 \leq |r| < .80$; $.80 \leq |r|$; respectively). In our tests for convergent validity, strong convergent correlations between an LLM's IPIP-NEO and BFI scores indicate that we are capturing the same underlying signals of each personality domain even when we measured them using two separate instruments. Weak convergent correlations indicate that at least one of the personality domain subscales is not capturing these signals properly.

\textbf{Discriminant Validity}:
We assessed the discriminant validity of the IPIP-NEO for LLMs through how its domain subscales remained relatively unrelated with their respective discriminant subscales. To do so, we compared each convergent correlation between the IPIP-NEO and BFI with all other correlations (i.e., discriminant correlations) located in the same row or column of the MTMM.
Discriminant validity was established for a personality domain subscale when the average difference ($\Delta$) between its convergent correlation and respective discriminant correlations was at least moderate ($\geq 0.40$). For example, a given model's IPIP-NEO Extraversion scores were tested for discriminant validity by being sufficiently more positively correlated with BFI Extraversion than with BFI Agreeableness, Conscientiousness, Neuroticism, and Openness, according to this average difference metric.

\textbf{Criterion Validity:}
As reported Section \ref{sec:methods-reliability-construct-validity-overview}, we evaluated the criterion validity of our LLM personality test data in three steps. First, for each Big Five domain, we identified at least one theoretically-related external (viz. non-personality) construct reported in human research. Next, according to this existing human research, we selected appropriate psychometric tests to measure these related constructs and administered them to LLMs (Supplemental Table \ref{tab:external-criteria-measures} shows the 11 criterion subscales). Finally, we correlated LLM scores for each IPIP-NEO subscale with these external measures.

\textbf{Structural Validity:}
We evaluated the structural properties of our primary personality measure at both the domain and test levels. At the domain level, we computed McDonald's ($\omega$), a reliability index based on factor saturation, which tested the factorial structure within each domain (as reported above). For models that showed construct validity, this was high. Second, at the test level, we computed inter-trait correlations to check if traits correlated with each other as expected in humans (ref to main section).

It was determined for the current work that using conventional factor analysis as a structural validity check was not appropriate for several reasons. First, this work did not fall under the remit of new test construction (i.e., entirely new tests for LLMs). It instead relied on personality scales containing fixed structural assumptions as a result of human population data during test development process. Future work could develop entirely new personality tests with item structures specifically tailored for LLMs. Second, while our tested models were prompted with randomly-sampled personas to introduce necessary variance in LLM responses, these personas were deterministically duplicated and combined with instruction changes across prompts. As such, it was clear that variations introduced this prompting method did not constitute sufficiently random individual variance necessary for factor analysis. Last, and most importantly, since we used classical test theory-based (CTT) scoring to validate real-world model behaviors (where scores for each trait were calculated as the average of their underlying items), there was no need to check factorial structure as long as our test of interest showed sufficient convergent, discriminant, and criterion validity and reliability.

In a purely exploratory fashion and with great caution, nevertheless, we conducted factor analyses to gauge the percentage of IPIP-NEO items that sufficiently loaded onto their correct factors. Specifically, we used the following procedure:
(1) We first checked the appropriateness of each model's IPIP-NEO data for exploratory factor analysis (EFA) by computing Bartlett's test of sphericity \citep{bartlett1951sphericitytest} and the Kaiser, Meyer, Olkin (KMO) overall measure of sampling adequacy \citep{kaiser1974kmo}. Bartlett's test of sphericity flags if there is sufficient significant correlation in the data for factor analysis, while  The data of 13 models met these criteria---showing sufficient significant correlation and KMO $\geq 0.50$ per these two tests---were selected for analysis in the next step.
(2) We fit a minimum residual factor analysis using the $psych$ package in R, extracting five factors from each selected model's data. Applying an orthogonal equamax rotation \citep{crawford1970rotation} produced the most interpretable solution across models.
(3) We assigned factor labels by summing the absolute loadings of each item with loadings $> 0.30$, grouped by the actual domain labels of the items, and selecting the domain name with the largest sum. 
(4) Finally, we correlated the factor scores derived from these EFAs with our actual CTT-based domain scale scores to test if the CTT-based scores used in the current work adequately reflected variance captured by EFA-based solutions.

\begin{figure*}[tb]
    \centering
    \begin{subfigure}{0.47\textwidth}
         \centering
        \includegraphics[trim=15 15 40 35,clip,width=\textwidth]{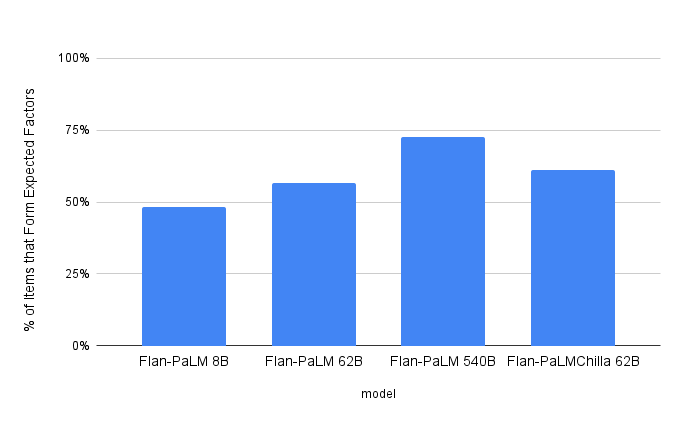}
         \caption{\PaLM}
         \label{fig:esv_palm}
     \end{subfigure}
     \hspace{0.25in}
     \begin{subfigure}{0.47\textwidth}
         \centering
        \includegraphics[trim=15 15 40 35,clip,width=\textwidth]{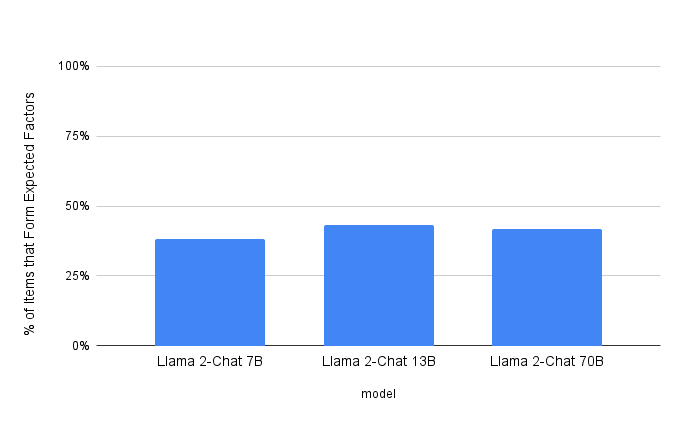}
         \caption{\LlamaTwo}
         \label{fig:esv_llama}
     \end{subfigure}
     \hspace{0.25in}
     \begin{subfigure}{0.47\textwidth}
         \centering
        \includegraphics[trim=15 15 40 35,clip,width=\textwidth]{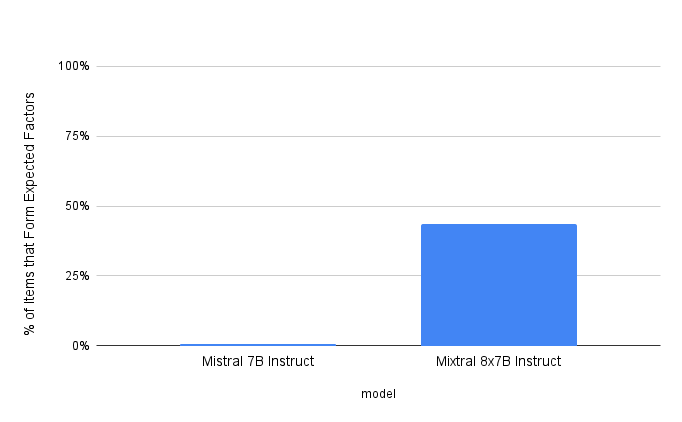}
         \caption{Mis(x)tral}
         \label{fig:esv_mistral}
     \end{subfigure}
     \hspace{0.25in}
     \begin{subfigure}{0.47\textwidth}
         \centering
        \includegraphics[trim=15 15 40 35,clip,width=\textwidth]{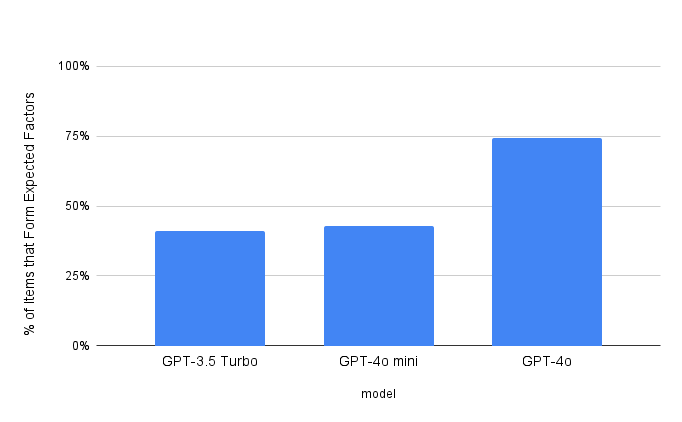}
         \caption{\GPT}
         \label{fig:esv_gpt}
     \end{subfigure}
\caption{Exploratory structural validity (ESV) of IPIP-NEO personality test data, organized by model family. We visualize ESV as the percentage of test items that loaded at least 0.30 onto their human-expected factors as part of an exploratory factor analysis.}
\label{fig:exploratory_structural_validity}
\vspace{-0.4cm}
\end{figure*}

This exploratory analysis revealed relative differences in what we will refer to as \textit{exploratory structural validity} (ESV). Similar to what we found in our main validity checks, test data from instruction-tuned and relatively larger models showed stronger signs of ESV. \FlanPaLMFiveFortyB\ and \GPTFourO\ data showed imperfect but relatively strong ESV: their items sufficiently loaded onto their human-expected factors over 72\% of the time (Supplemental Figure \ref{fig:exploratory_structural_validity}). Items answered by \LlamaTwoSevenBChat\ and \MixtralEightXSevenBInstruct, on the other hand, loaded as expected less than 45\% of the time, suggesting more questionable ESV. However, even with suboptimal EFA results, we found on average that EFA-based factor scores strongly correlated with CTT-based scores domain scores (Supplemental Table \ref{tab:efa_ctt_correlations}), illustrating that the IPIP-NEO adequately captured response variation across the Big Five for these flagship models. Therefore, while we could have directly refined the content of the IPIP-NEO (e.g., by removing poorly performing items), these correlations signaled that doing so would not have substantially affected the inferences of this work derived from CTT-based domain scores. We are hopeful future research can improve upon our framework by developing custom, factor analytically-derived, psychometric tests for LLMs.

\begin{table*}[tb]
    \centering
    \begin{tabular}{@{}lcccccc@{}}
    \toprule
    \multirow{2}[2]{*}{Model}                       &   \multicolumn{5}{c}{$|r|$}                               &           \\
    \cmidrule(l){2-6}
                                                    &   EXT     &   AGR     &   CON     &   NEU     &   OPE     &   Avg.    \\
    \midrule
    {\FlanPaLMFiveFortyB}\arraybackslash            &   $0.84$  &   $0.83$  &   $0.73$  &   $0.38$  &   $0.87$  &   $0.73$  \\
    {\LlamaTwoSeventyBChat}\arraybackslash          &   $0.19$  &   $0.73$  &   $0.70$  &   $0.63$  &   $0.86$  &   $0.62$  \\
    {\MixtralEightXSevenBInstruct}\arraybackslash   &   $0.73$  &   $0.78$  &   $0.60$  &   $0.61$  &   $0.51$  &   $0.65$  \\
    {\GPTFourO}\arraybackslash                      &   $0.41$  &   $0.74$  &   $0.69$  &   $0.79$  &   $0.90$  &   $0.71$  \\
    \botrule
    \end{tabular}
    \caption{Associations between classical test theory-based (CTT) and exploratory factor analysis-derived (EFA) domain scale scores. Associations are presented as absolute Pearson correlations. Stronger correlations indicate exploratory structural validity: that the CTT-based scores used in the current work align with the underlying (exploratory) factor structure per model, found via EFA. All coefficients are significant at $p < 0.0001$ ($\textit{n} = 1,250$ observations per model).}
    \label{tab:efa_ctt_correlations}
\end{table*}



\begin{figure*}[tb]
    \centering
    \begin{subfigure}{\textwidth}
         \centering
         \includegraphics[trim=5 10 5 5,clip,width=\textwidth]{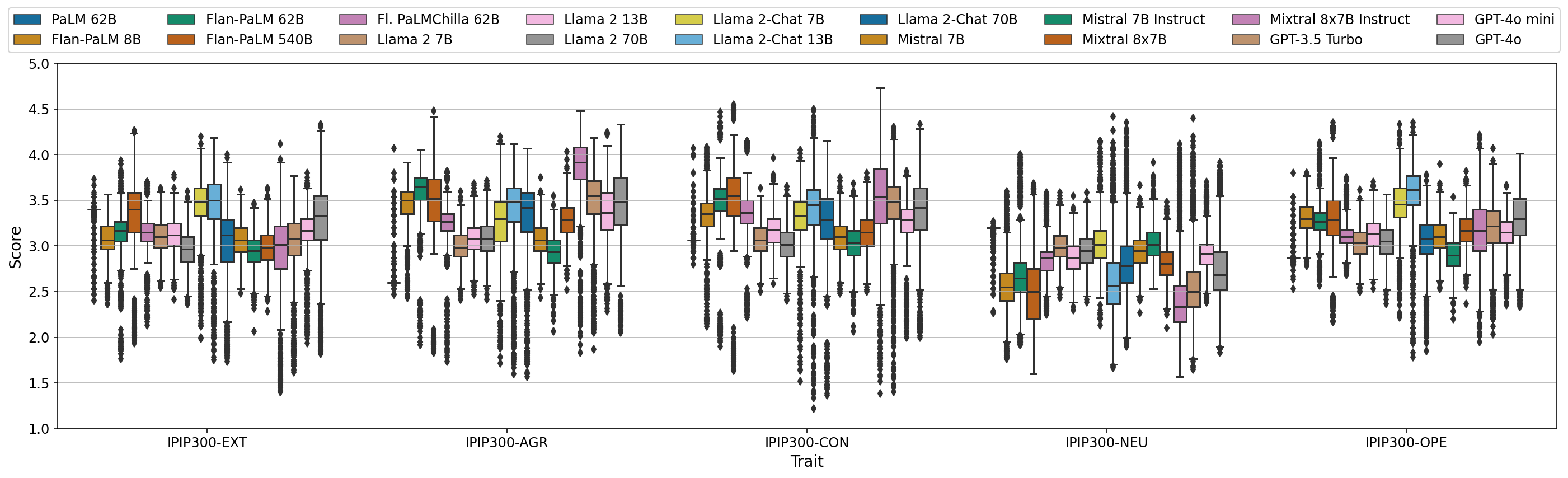}
         \caption{IPIP-NEO}
         \label{fig:model-descriptives-ipip}
     \end{subfigure}
     \begin{subfigure}{\textwidth}
         \centering
         \includegraphics[trim=5 10 5 5,clip,width=\textwidth]{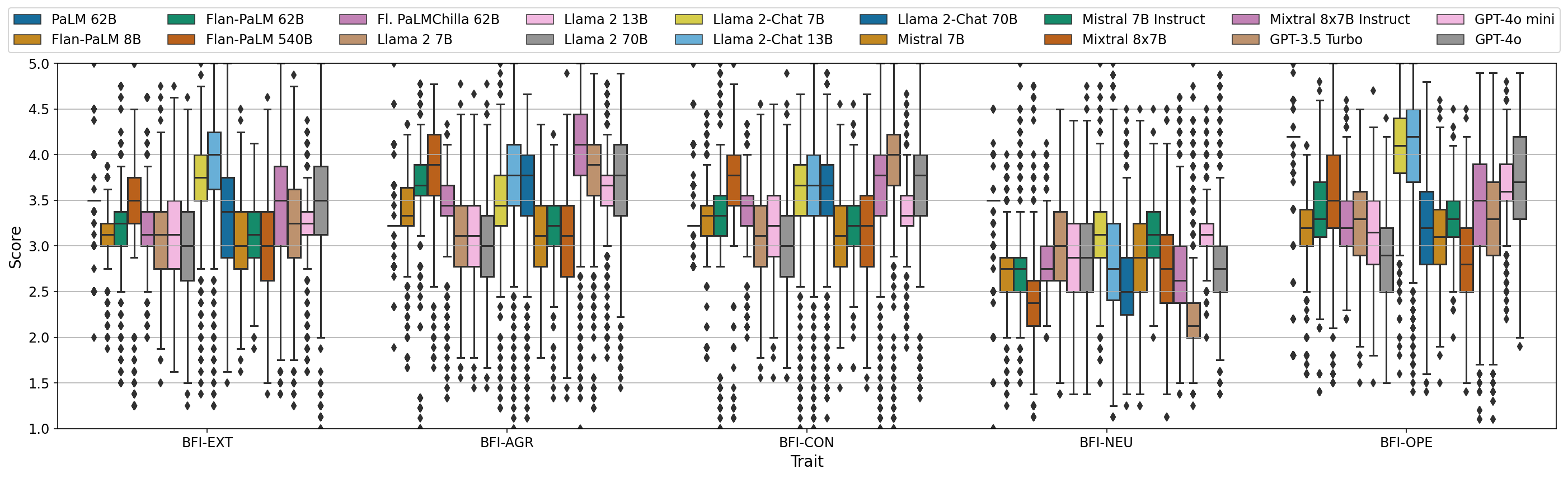}
         \caption{BFI}
         \label{fig:model-descriptives-bfi}
     \end{subfigure}
\caption{\small Distributions of a) IPIP-NEO and b) BFI personality domain scores across models. Box plots depict model medians surrounded by their interquartile ranges and outlier values. As models increased in size (e.g., \FlanPaLM\ from 8B to 540B), a) IPIP-NEO scores were relatively more stable compared to b) BFI scores, where scores for socially-desirable traits increased while NEU scores decreased.
}
\label{fig:model-descriptives-overall}
\vspace{-0.4cm}
\end{figure*}

\section{Personality Assessment Results}
\label{app:personality-measurement-results}

\subsection{Descriptive Statistics Across Models}
\label{sec:descriptive-stats-across-models}
We inspected the distributions of IPIP-NEO and BFI test scores across models. We examined how the distributions shifted as a function of model size (holding model training method constant) and model training method (holding model size constant). Figure \ref{fig:model-descriptives-overall} summarizes the findings.

\textit{By model configuration:} 
At 62B parameters, base \PaLM\ showed nearly uniform personality score distributions for both the IPIP-NEO and BFI, with 25th, 50th, and 75th percentile values identical within each BFI domain. 
Instruction-tuned variants, \FlanPaLM\ and \FlanPaLMChilla, showed more normal distributions of personality, with lower kurtosis. Instruction-tuned versions of \LlamaTwo\ and \MixtralEightXSevenB\ showed elevated IPIP-NEO and BFI levels of socially-desirable traits (EXT, AGR, CON, OPE) and lower levels of NEU.

\textit{By model size:} \FlanPaLM\ IPIP-NEO (Figure \ref{fig:model-descriptives-ipip}) and BFI (Figure \ref{fig:model-descriptives-bfi}) scores were stable across model sizes.
Median levels of socially-desirable BFI subscales (EXT, AGR, CON, OPE) substantially increased as \FlanPaLM's size increased. In contrast, median levels of BFI NEU decreased (from $2.75$ to $2.38$) as \FlanPaLM\ scaled from 8B to 540B parameters. Distributions of IPIP-NEO scores were more stable across sizes of \FlanPaLM: only IPIP-NEO EXT and CON showed noticeable increases by model size. For instance, across sizes of \FlanPaLM, median levels of IPIP-NEO OPE remained close to $3.30$. Meanwhile, median BFI AGR scores monotonically increased from $3.33$ to $3.67$ and $3.89$ for \FlanPaLMEightB, \FlanPaLMSixtyTwoB, and \FlanPaLMFiveFortyB, respectively. Model scale tracked elevated IPIP-NEO and BFI levels of socially-desirable traits for \Mistral\ and \GPTFourO\ models only (i.e., moving from \MistralSevenBInstruct\ to \MixtralEightXSevenBInstruct\ and \GPTFourOMini\ to \GPTFourO).

\subsection{Reliability Results}
\label{app:results-structural-validity}

Following established frameworks from measurement science outlined in Sections \ref{app:background-construct-validity}, we evaluated the reliability of the tests---the extent to which they dependably measured single underlying factors---by quantifying internal consistency and factor saturation for each administered subscale. Supplemental Tables \ref{tab:results-reliability-ipip-closed} and \ref{tab:results-reliability-ipip-open} summarize the results.

\textit{By model configuration:} Among the models of the same size (i.e., \PaLM, \FlanPaLM, and \FlanPaLMChilla\ 62B; \LlamaTwo\ and \LlamaTwoChat\ 7B, 13B, and 70B; \MistralSevenB\ and \MistralSevenBInstruct; and \MixtralEightXSevenB\ and \MixtralEightXSevenBInstruct) instruction fine-tuned variants' responses to personality tests were highly reliable. \FlanPaLMSixtyTwoB and \FlanPaLMChillaSixtyTwoB, for instance, demonstrated excellent internal consistency ($\alpha$, $\lambda_6$) and factor saturation ($\omega$), with all three metrics in the mid to high 0.90s. In contrast, we found \PaLMSixtyTwoB\ (a model that is not instruction fine-tuned) to have highly \textit{unreliable} ($-0.55 \leq \alpha \leq 0.67$) responses. Although \PaLMSixtyTwoB\ personality test data appeared to form distinct factors for each Big Five trait, with close to perfect ($> 0.99$) values for McDonald's $\omega$, its responses were highly inconsistent, with values for Cronbach's $\alpha$ ranging from poor ($0.67$) to unacceptable ($-0.55$). Computing reliability indices for \FlanPaLMChillaSixtyTwoB's IPIP-NEO CON and OPE data required removal of two items showing zero variance; for these two items, \FlanPaLMChillaSixtyTwoB\ responded identically across 1,250 simulated participant prompt sets.

\begin{table*}[tbp]
    \setlength\extrarowheight{2pt} 
    \caption{\small IPIP-NEO reliability metrics per model for proprietary (closed-source) models. Consistent with human standards, we interpreted a given reliability metric $RM$ (i.e., $\alpha$, $\lambda_6$, $\omega$) $< 0.50$ as unacceptable; $0.50 \leq RM < 0.60$ as poor; $0.60 \leq RM < 0.70$ as questionable; $0.70 \leq RM < 0.80$ as acceptable; $0.80 \leq RM < 0.90$ as good; and $RM \geq 0.90$ as excellent. $^*$ $RM$s for these subscales were calculated after removing one item with zero variance, since reliability cannot be computed for items with zero variance.
    }
    \label{tab:results-reliability-ipip-closed}
    \centering
    \footnotesize
    \sisetup{add-integer-zero=false}
    \begin{tabular}{lcrrrc}
        \toprule
        Model & Subscale & \shortstack{Cronbach's\\$\alpha$} & \shortstack{Guttman's\\$\lambda_6$} & \shortstack{McDonald's\\$\omega$} & \shortstack{Overall\\Interpretation} \\
        \midrule
           & IPIP-NEO EXT                     &   0.57     &   0.98     &   1.00 & Poor    \\
        & IPIP-NEO AGR                     &   0.67     &   0.99     &   1.00 &   Questionable \\
         \PaLMSixtyTwoB & IPIP-NEO CON                     &   $-0.55$    &   0.93     &   1.00 & Unacceptable    \\
         & IPIP-NEO NEU                     &   0.10     &   0.96     &   1.00  & Unacceptable  \\
          & IPIP-NEO OPE                     &   $-0.35$    &   0.92     &   1.00  & Unacceptable  \\ \hline
           & IPIP-NEO EXT                     &   0.83     &   0.94     &   0.97 & Good     \\
        & IPIP-NEO AGR                     &   0.88     &   0.95     &   0.94 & Good    \\
         \FlanPaLMEightB & IPIP-NEO CON                     &   0.92     &   0.97     &   0.97 & Excellent     \\
         & IPIP-NEO NEU                     &   0.93     &   0.97     &   0.96 & Excellent    \\
         & IPIP-NEO OPE                     &   0.75     &   0.92     &   0.97 & Acceptable     \\\hline
         & IPIP-NEO EXT                     &   0.94     &   0.98     &   0.96 & Excellent     \\
         & IPIP-NEO AGR                     &   0.95     &   0.99     &   0.97 & Excellent    \\
         \FlanPaLMSixtyTwoB & IPIP-NEO CON                     &   0.96     &   0.99     &   0.98 & Excellent    \\
         & IPIP-NEO NEU                     &   0.96     &   0.99     &   0.97 & Excellent    \\
         & IPIP-NEO OPE                     &   0.84     &   0.95     &   0.93  & Acceptable   \\\hline
        & IPIP-NEO EXT                     &   0.96     &   0.99     &   0.97  & Excellent   \\
        & IPIP-NEO AGR                     &   0.97     &   0.99     &   0.98  & Excellent   \\
         \FlanPaLMFiveFortyB & IPIP-NEO CON                     &   0.98     &   0.99     &   0.98 & Excellent     \\
        & IPIP-NEO NEU                     &   0.97     &   0.99     &   0.98 & Excellent   \\
        & IPIP-NEO OPE                     &   0.95     &   0.99     &   0.97  & Excellent   \\\hline
        & IPIP-NEO EXT                     &   0.94     &   0.98     &   0.95 & Excellent    \\
        & IPIP-NEO AGR                     &   0.96     &   0.99     &   0.98 & Excellent    \\
        \FlanPaLMChillaSixtyTwoB    & IPIP-NEO CON     &   0.96     &   0.97     &   0.99    & Excellent$^*$ \\
        & IPIP-NEO NEU                     &   0.95     &   0.98     &   0.97   & Excellent  \\
        & IPIP-NEO OPE                     &   0.90     &   0.92     &   0.96  & Excellent$^*$   \\\hline
        & IPIP-NEO EXT                                          &   0.92    &   0.96    &   0.94    &   Excellent       \\
        & IPIP-NEO AGR                                          &   0.93    &   0.96    &   0.95    &   Excellent       \\
        \GPTThreeDotFiveTurbo       &   IPIP-NEO CON            &   0.95    &   0.97    &   0.96    &   Excellent       \\
        & IPIP-NEO NEU                                          &   0.95    &   0.97    &   0.96    &   Excellent       \\
        & IPIP-NEO OPE                                          &   0.88    &   0.94    &   0.89    &   Good            \\\hline
        & IPIP-NEO EXT                                          &   0.93    &   0.97    &   0.95    &   Excellent       \\
        & IPIP-NEO AGR                                          &   0.95    &   0.97    &   0.96    &   Excellent       \\
        \GPTFourOMini                           & IPIP-NEO CON  &   0.93    &   0.96    &   0.94    &   Excellent       \\
        & IPIP-NEO NEU                                          &   0.92    &   0.96    &   0.93    &   Excellent       \\
        & IPIP-NEO OPE                                          &   0.90    &   0.95    &   0.92    &   Good            \\\hline
        & IPIP-NEO EXT                                          &   0.97    &   0.99    &   0.98    &   Excellent       \\
        & IPIP-NEO AGR                                          &   0.97    &   0.99    &   0.98    &   Excellent       \\
        \GPTFourO                               & IPIP-NEO CON  &   0.97    &   0.98    &   0.98    &   Excellent       \\
        & IPIP-NEO NEU                                          &   0.97    &   0.99    &   0.98    &   Excellent       \\
        & IPIP-NEO OPE                                          &   0.95    &   0.97    &   0.96    &   Excellent       \\
        \botrule
    \end{tabular}
\end{table*}

\begin{table*}[tbp]
    \setlength\extrarowheight{2pt} 
    \caption{\small IPIP-NEO reliability metrics per model for open-sourced models. Consistent with human standards, we interpreted a given reliability metric $RM$ (i.e., $\alpha$, $\lambda_6$, $\omega$) $< 0.50$ as unacceptable; $0.50 \leq RM < 0.60$ as poor; $0.60 \leq RM < 0.70$ as questionable; $0.70 \leq RM < 0.80$ as acceptable; $0.80 \leq RM < 0.90$ as good; and $RM \geq 0.90$ as excellent. $^*$ $RM$s for these subscales were calculated after removing one item with zero variance, since reliability cannot be computed for items with zero variance.
    }
    \label{tab:results-reliability-ipip-open}
    \centering
    \footnotesize
    \sisetup{add-integer-zero=false}
    \begin{tabular}{lcrrrc}
        \toprule
        Model & Subscale & \shortstack{Cronbach's\\$\alpha$} & \shortstack{Guttman's\\$\lambda_6$} & \shortstack{McDonald's\\$\omega$} & \shortstack{Overall\\Interpretation} \\
        \midrule
        & IPIP-NEO EXT                                          &   0.04    &   0.09    &   0.22    &   Unacceptable    \\
        & IPIP-NEO AGR                                          &   0.03    &   0.08    &   0.26    &   Unacceptable    \\
        \LlamaTwoSevenB             &   IPIP-NEO CON            &   0.05    &   0.09    &   0.22    &   Unacceptable    \\
        & IPIP-NEO NEU                                          &   0.03    &   0.08    &   0.24    &   Unacceptable    \\
        & IPIP-NEO OPE                                          &   0.04    &   0.09    &   0.21    &   Unacceptable    \\\hline
        & IPIP-NEO EXT                                          &   0.07    &   0.11    &   0.20    &   Unacceptable    \\
        & IPIP-NEO AGR                                          &   0.07    &   0.11    &   0.23    &   Unacceptable    \\
        \LlamaTwoThirteenB          &   IPIP-NEO CON            &   0.07    &   0.11    &   0.20    &   Unacceptable    \\
        & IPIP-NEO NEU                                          &   0.02    &   0.07    &   0.24    &   Unacceptable    \\
        & IPIP-NEO OPE                                          &   0.00    &   0.05    &   0.23    &   Unacceptable    \\\hline
        & IPIP-NEO EXT                                          &   0.01    &   0.06    &   0.46    &   Unacceptable    \\
        & IPIP-NEO AGR                                          &   0.02    &   0.08    &   0.47    &   Unacceptable    \\
        \LlamaTwoSeventyB           &   IPIP-NEO CON            &   0.07    &   0.12    &   0.43    &   Unacceptable    \\
        & IPIP-NEO NEU                                          &   0.00    &   0.06    &   0.46    &   Unacceptable    \\
        & IPIP-NEO OPE                                          &   -0.63   &   -0.01   &   0.42    &   Unacceptable    \\\hline
        & IPIP-NEO EXT                                          &   0.83    &   0.88    &   0.94    &   Good            \\
        & IPIP-NEO AGR                                          &   0.85    &   0.88    &   0.90    &   Good            \\
        \LlamaTwoSevenBChat         &   IPIP-NEO CON            &   0.84    &   0.88    &   0.92    &   Good            \\
        & IPIP-NEO NEU                                          &   0.80    &   0.84    &   0.92    &   Good            \\
        & IPIP-NEO OPE                                          &   0.76    &   0.82    &   0.92    &   Acceptable      \\\hline
        & IPIP-NEO EXT                                          &   0.90    &   0.93    &   0.92    &   Excellent       \\
        & IPIP-NEO AGR                                          &   0.92    &   0.94    &   0.95    &   Excellent       \\
        \LlamaTwoThirteenBChat      &   IPIP-NEO CON            &   0.93    &   0.95    &   0.95    &   Excellent       \\
        & IPIP-NEO NEU                                          &   0.93    &   0.95    &   0.95    &   Excellent       \\
        & IPIP-NEO OPE                                          &   0.87    &   0.90    &   0.88    &   Good            \\\hline
        & IPIP-NEO EXT                                          &   0.89    &   0.92    &   0.91    &   Good            \\
        & IPIP-NEO AGR                                          &   0.92    &   0.94    &   0.94    &   Excellent       \\
        \LlamaTwoSeventyBChat       &   IPIP-NEO CON            &   0.93    &   0.94    &   0.94    &   Excellent       \\
        & IPIP-NEO NEU                                          &   0.92    &   0.93    &   0.93    &   Excellent       \\
        & IPIP-NEO OPE                                          &   0.81    &   0.85    &   0.86    &   Good            \\\hline
        & IPIP-NEO EXT                                          &   0.10    &   0.14    &   0.23    &   Unacceptable    \\
        & IPIP-NEO AGR                                          &   0.03    &   0.08    &   0.24    &   Unacceptable    \\
        \MistralSevenB   &   IPIP-NEO CON            &   0.10    &   0.15    &   0.31    &   Unacceptable    \\
        & IPIP-NEO NEU                                          &   0.04    &   0.09    &   0.25    &   Unacceptable    \\
        & IPIP-NEO OPE                                          &   0.12    &   0.16    &   0.28    &   Unacceptable    \\\hline
        & IPIP-NEO EXT                                          &   0.29    &   0.33    &   0.41    &   Unacceptable    \\
        & IPIP-NEO AGR                                          &   0.31    &   0.35    &   0.42    &   Unacceptable    \\
        \MistralSevenBInstruct   & IPIP-NEO CON      &   0.53    &   0.55    &   0.53    &   Poor            \\
        & IPIP-NEO NEU                                          &   0.45    &   0.48    &   0.46    &   Unacceptable    \\
        & IPIP-NEO OPE                                          &   0.35    &   0.39    &   0.37    &   Unacceptable    \\\hline
        & IPIP-NEO EXT                                          &   0.16    &   0.20    &   0.43    &   Unacceptable    \\
        & IPIP-NEO AGR                                          &   0.11    &   0.15    &   0.28    &   Unacceptable    \\
        \MixtralEightXSevenB &   IPIP-NEO CON        &   0.12    &   0.16    &   0.49    &   Unacceptable    \\
        & IPIP-NEO NEU                                          &   0.11    &   0.16    &   0.44    &   Unacceptable    \\
        & IPIP-NEO OPE                                          &   0.08    &   0.12    &   0.36    &   Unacceptable    \\\hline
        & IPIP-NEO EXT                                          &   0.91    &   0.94    &   0.92    &   Excellent       \\
        & IPIP-NEO AGR                                          &   0.91    &   0.94    &   0.94    &   Excellent       \\
        \MixtralEightXSevenBInstruct & IPIP-NEO CON  &   0.93    &   0.96    &   0.95    &   Excellent       \\
        & IPIP-NEO NEU                                          &   0.93    &   0.95    &   0.95    &   Excellent       \\
        & IPIP-NEO OPE                                          &   0.82    &   0.88    &   0.92    &   Good            \\
        \botrule
    \end{tabular}
\end{table*}

\textit{By model size:}
Across models of the same training configuration (e.g., \FlanPaLMEightB, \FlanPaLMSixtyTwoB, and \FlanPaLMFiveFortyB), the reliability of synthetic personality measurements increased with model size. Across model sizes of \FlanPaLM, as shown in Tables \ref{tab:results-reliability-ipip-closed} and  \ref{tab:results-reliability-ipip-open}, internal consistency reliability (i.e., $\alpha$) of IPIP-NEO scores improved from acceptable to excellent. At 8B parameters, internal consistency was acceptable for IPIP-NEO Openness ($\alpha = 0.75$), good for IPIP-NEO Extraversion and Agreeableness ($\alpha$s $0.83$, $.88$, respectively), and excellent ($\alpha \geq 0.90$) for IPIP-NEO Conscientiousness and Neuroticism. At 62B parameters, internal consistency was good for IPIP-NEO Openness ($\alpha = 0.84$) and excellent for all other traits ($\alpha \geq 0.90$). At 540B parameters, all IPIP-NEO domain scales showed excellent internal consistency ($\alpha \geq 0.90$). Our other reliability indices, Guttman's $\lambda_6$ and McDonald's $\omega$, improved within the same excellent range from 8B to 540B variants of \FlanPaLM.

We observed a similar pattern of reliability scaling with size among instruction-tuned open models we tested. Across \LlamaTwoChat\ models, \LlamaTwoSevenBChat's data ranged from acceptable to good, while \LlamaTwoSeventyBChat's data showed excellent reliability. \MistralSevenBInstruct's response reliability was poor to unacceptable, while that of \MixtralEightXSevenBInstruct\ was mostly excellent. 
Reliability was unacceptable for the open base models we tested, regardless of size (i.e., \LlamaTwo\ 7B, 13B, 70B).
This suggests that the reliability of LLM responses to psychometric tests is more directly a result of instruction tuning rather than size.

\subsection{Convergent and Discriminant Validation Results}
\label{app:results-convergent-validity}

\begin{figure*} [tb]
    \centering
    \includegraphics[trim=5 5 5 5,clip,width=1.0\textwidth]{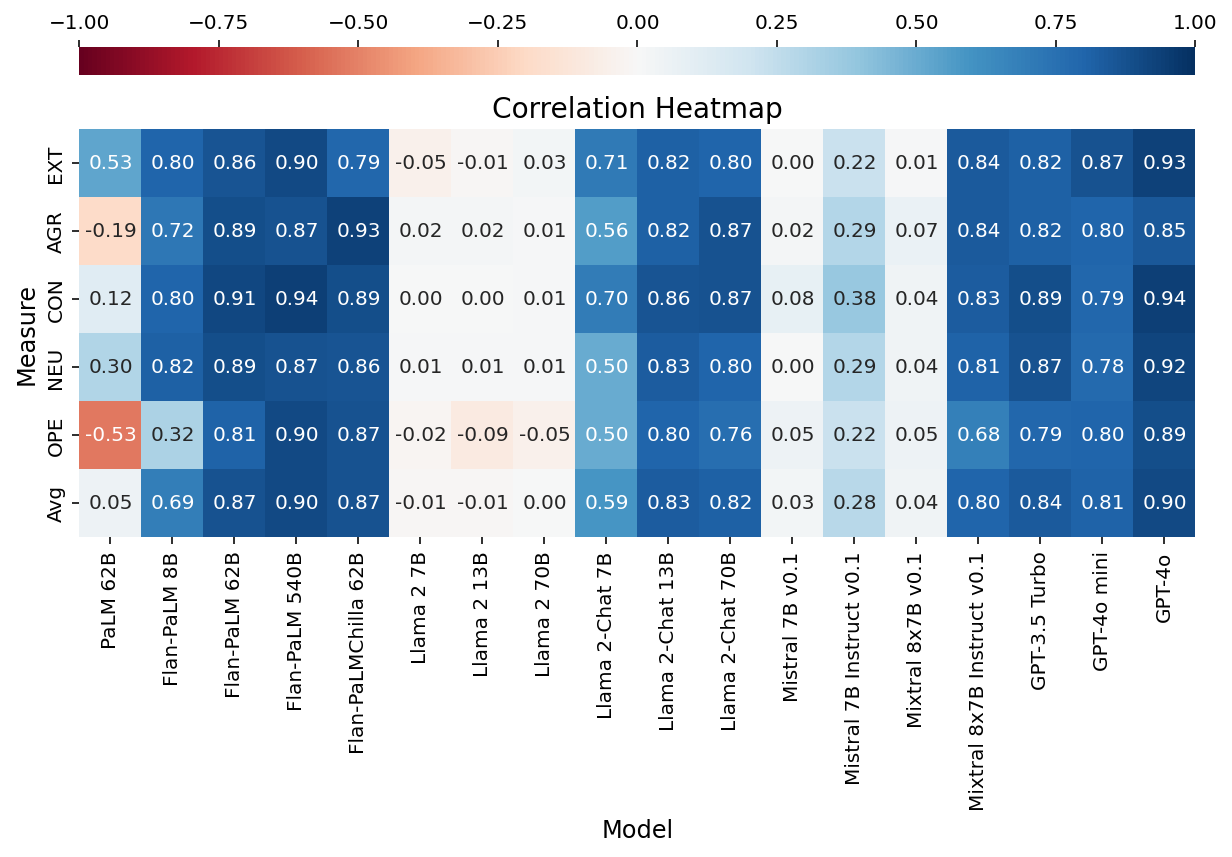}
    \caption{\small Convergent Pearson's correlations ($r$s) between IPIP-NEO and BFI scores by model. Heatmap illustrates the averaged similarities (convergence) between IPIP-NEO and BFI score variation for each Big Five domain; the last row represents average correlations across all measures for a model. Stronger correlations (blue) indicate higher levels of convergence and provide evidence for convergent validity. EXT = extraversion; AGR = agreeableness; CON = conscientiousness; NEU = neuroticism; OPE = openness. All correlations are statistically significant at $p < 0.0001$; $n = 1,250$.}
    \label{fig:convergent-validity}
\end{figure*}

\begin{table}
    \caption{\small Summary of convergent and discriminant validity evidence across models. LLM personality measurements demonstrate convergent validity when the average convergent correlation ($r_\textsubscript{conv}$) between equivalent IPIP-NEO and BFI subscales is strong ($\geq 0.60$; marked in \textit{italics}) or very strong ($\geq 0.80$; marked in \textbf{boldface}). Discriminant validity is evidenced when the average difference ($\Delta$) between a model's convergent ($r_\textsubscript{conv}$) and respective discriminant ($r_\textsubscript{discr}$) correlations between personality tests is at least moderate (avg. $\Delta \geq 0.40$; shown in boldface). All underlying convergent correlations of models with an average $r_\textsubscript{conv}$ $\geq .05$ are statistically significant at $p < .0001$; $n = 1,250$ per model.} 
    \centering
    \begin{tabular} {p{0.4\linewidth} rp{0.1\linewidth} rp{0.1\linewidth} rp{0.1\linewidth}}
    \toprule
    Model   &   \shortstack{Avg. \\$r_\textsubscript{conv}$}  &   \shortstack{Avg. \\$r_\textsubscript{discr}$}   &   \shortstack{Avg. \\$\Delta$} \\
    \midrule
    \PaLMSixtyTwoB                      &   $0.05$          &   $0.29$  &   $-0.24$         \\
    \FlanPaLMEightB                     &   $\textit{0.69}$ &   $0.46$  &   $0.23$          \\
    \FlanPaLMSixtyTwoB                  &   $\textbf{0.87}$ &   $0.46$  &   $\textbf{0.41}$ \\
    \FlanPaLMFiveFortyB                 &   $\textbf{0.90}$ &   $0.39$  &   $\textbf{0.51}$ \\
    \FlanPaLMChillaSixtyTwoB            &   $\textbf{0.87}$ &   $0.39$  &   $\textbf{0.48}$ \\
    \hline
    \LlamaTwoSevenB                     &   $-0.01$         &   $0.02$  &   $-0.03$         \\
    \LlamaTwoThirteenB                  &   $-0.01$         &   $0.03$  &   $-0.05$         \\
    \LlamaTwoSeventyB                   &   $0.00$          &   $0.03$  &   $-0.02$         \\
    \LlamaTwoSevenBChat                 &   $0.59$          &   $0.44$  &   $0.15$          \\
    \LlamaTwoThirteenBChat              &   $\textbf{0.82}$ &   $0.54$  &   $0.29$          \\
    \LlamaTwoSeventyBChat               &   $\textbf{0.80}$ &   $0.39$  &   $\textbf{0.42}$ \\
    \hline
    \MistralSevenB                      &   $0.03$          &   $0.04$  &   $-0.01$         \\
    \MistralSevenBInstruct              &   $0.28$          &   $0.20$  &   $0.09$          \\
    \MixtralEightXSevenB                &   $0.04$          &   $0.03$  &   $0.01$          \\
    \MixtralEightXSevenBInstruct        &   $\textbf{0.80}$ &   $0.40$  &   $\textbf{0.40}$ \\
    \hline
    \GPTThreeDotFiveTurbo               &   $\textbf{0.84}$ &   $0.55$  &   $0.28$          \\
    \GPTFourOMini                       &   $\textbf{0.81}$ &   $0.38$  &   $\textbf{0.43}$ \\
    \GPTFourO                           &   $\textbf{0.90}$ &   $0.42$  &   $\textbf{0.48}$ \\
    \botrule
    \end{tabular}
    \label{tab:avg-convergent-discriminant-validity}
\end{table}

The convergent and discriminant validity of personality measurements in LLMs varies across two axes: model size and model training method. Figure \ref{fig:convergent-validity} illustrates convergent validity in terms of how IPIP-NEO and BFI scores convergently correlate across models. Supplemental Table \ref{tab:avg-convergent-discriminant-validity} summarizes the average convergent and discriminant $r$s across models.

\begin{figure*}[th!]
    \centering
    \includegraphics[trim=5 5 5 5,clip,width=1.0\textwidth]{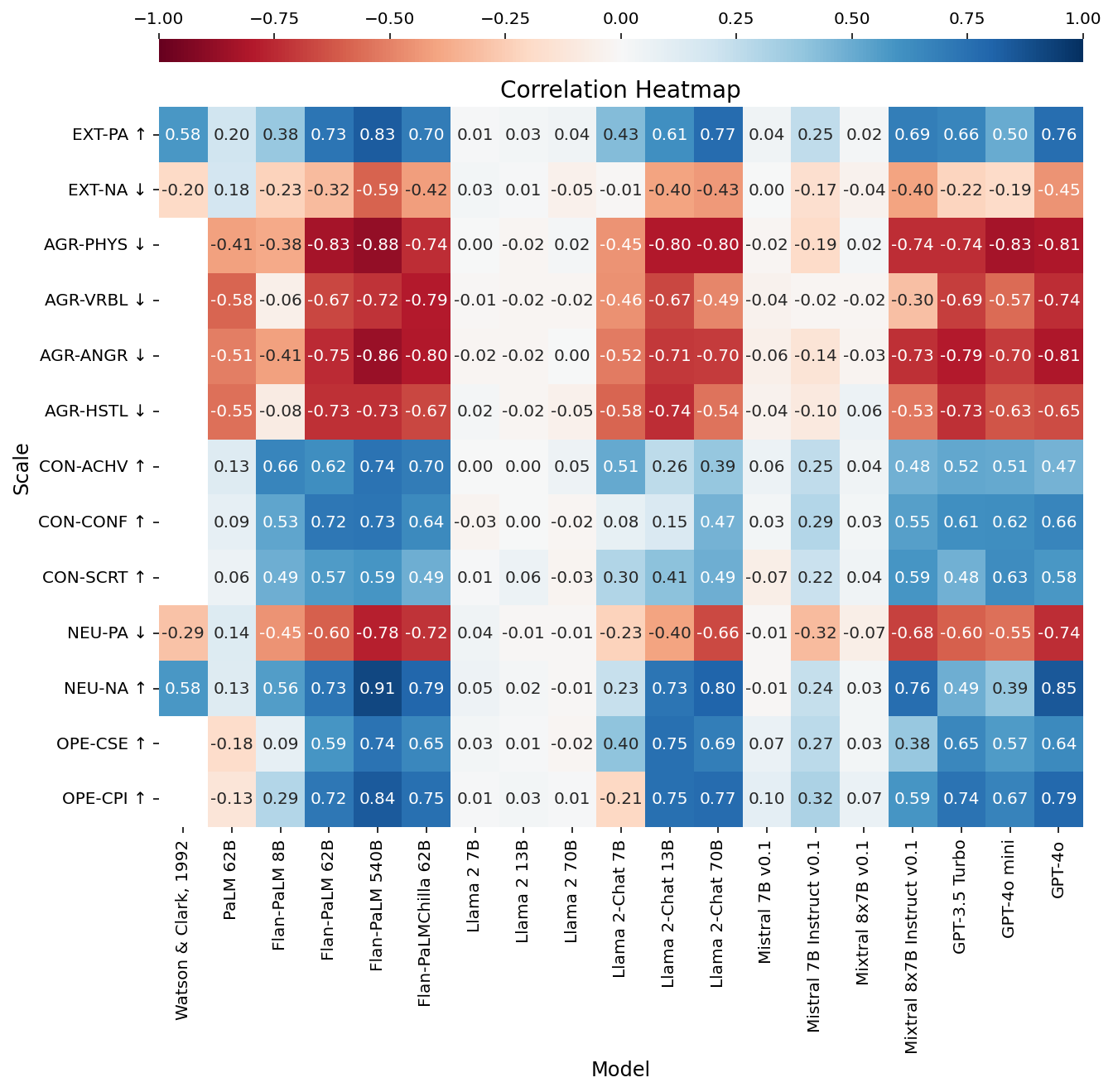}
     \caption{\small Criterion validity evidence of LLM personality measurements per domain. $\uparrow$ = personality domain and subscale referenced in the row label are expected to be directly correlated, $\downarrow$ = expected to have opposite correlation. Rows 1, 2: IPIP-NEO correlations among Extraversion with positive and negative affect, compared to human baselines (leftmost column), based on work in \cite{watson1992traits} which studied the relationship between personality and affect in humans; PA = PANAS Positive Affect; NA = Negative Affect; Rows 3 - 6: Agreeableness with subscales of trait aggression, measured by the Buss-Perry Aggression Questionnaire (BPAQ); PHYS = Physical Aggression; VRBL = Verbal Aggression; ANGR = Anger; HSTL = Hostility; Rows 7 - 9: Conscientiousness with related human values of achievement, conformity, and security (measured by PVQ-RR ACHV, CONF, and SCRT subscales, respectively); Rows 10, 11: Neuroticism with PA and NA compared to humans baselines \cite{watson1992traits}; Rows 12, 13: Openness with creativity, measured by the Creative Self-Efficacy (CSE) and Creative Personal Identity (CPI) subscales of the Short Scale of Creative Self (SSCS). All LLM correlations $> |0.09|$ are statistically significant at $p < 0.0001$; $n = 1,250$.}
\label{fig:external-validity}
\vspace{-0.4cm}
\end{figure*}

\section{LLM Personality Trait Shaping Methodology}
\label{app:methods-shaping-overview}


Having established a principled methodology for determining if an LLM personality measurement is valid and reliable, we investigated how that methodology can be applied to LLM prompting to shape that personality in desirable ways. This section explores the extent to which personality in LLMs can be verifiably controlled and shaped by presenting two evaluation methodologies. 

\subsection{Prompt Design and Rationale}
\label{app:methods-shaping-prompt-design}
Using linguistic qualifiers from common validated Likert-type response scales, we designed prompts to facilitate granular shaping of any trait at the following nine levels:
\texttt{
\begin{enumerate}
    \item extremely \{low adjective\}
    \item very \{low adjective\}
    \item \{low adjective\}
    \item a bit \{low adjective\}
    \item neither \{low adjective\} nor \{high adjective\}
    \item a bit \{high adjective\}
    \item \{high adjective\}
    \item very \{high adjective\}
    \item extremely \{high adjective\}
\end{enumerate}
} 

For example, to target a moderately high level (i.e., Level 7/9) of extraversion, we use the five high adjectives (first introduced in Section \ref{sec:shaping_method_overview}) from our list targeting extraversion at the domain level.

\texttt{For the following task, respond in a way that matches this description: "\{Biographic Description\}. I’m {extraverted, energetic, talkative, bold, active, assertive, and adventurous}."}

Similarly, an example prompt targeting slightly below average (i.e., Level 4/9) extraversion, using the five negatively-keyed adjectives targeting extraversion, is as follows:

\texttt{For the following task, respond in a way that matches this description: "\{Biographic Description\}. I’m {a bit introverted, a bit unenergetic, a bit silent, a bit timid, a bit inactive, a bit unassertive, and a bit unadventurous}."}

Supplemental Table \ref{appendix:tab:trait-adjectives} shows the full list of adjectives used to describe each trait in each personality domain.

\begin{table*}[tbp]
\caption{\small Pairs of adjectival markers that map onto IPIP-NEO personality facets and their higher-order Big Five domains, adapted from \cite{goldberg1992development}. Each pair of markers is salient to the low and high end of a given facet (or, in some cases, higher-order domain). For example, the trait marker ``unfriendly" can be used to describe an entity low on the IPIP-NEO Extraversion facet of Friendliness (E1).}
\label{appendix:tab:trait-adjectives}
\centering
\footnotesize
\begin{tabular}{llll}
\toprule
Domain & Facet                     & Low Marker                  & High Marker               \\ \midrule
EXT    & E1 - Friendliness         & unfriendly                  & friendly                  \\
EXT    & E2 - Gregariousness       & introverted                 & extraverted               \\
EXT    & E2 - Gregariousness       & silent                      & talkative                 \\
EXT    & E3 - Assertiveness        & timid                       & bold                      \\
EXT    & E3 - Assertiveness        & unassertive                 & assertive                 \\
EXT    & E4 - Activity Level       & inactive                    & active                    \\
EXT    & E5 - Excitement-Seeking   & unenergetic                 & energetic                 \\
EXT    & E5 - Excitement-Seeking   & unadventurous               & adventurous and daring    \\
EXT    & E6 - Cheerfulness         & gloomy                      & cheerful                  \\
AGR    & A1 - Trust                & distrustful                 & trustful                  \\
AGR    & A2 - Morality             & immoral                     & moral                     \\
AGR    & A2 - Morality             & dishonest                   & honest                    \\
AGR    & A3 - Altruism             & unkind                      & kind                      \\
AGR    & A3 - Altruism             & stingy                      & generous                  \\
AGR    & A3 - Altruism             & unaltruistic                & altruistic                \\
AGR    & A4 - Cooperation          & uncooperative               & cooperative               \\
AGR    & A5 - Modesty              & self-important              & humble                    \\
AGR    & A6 - Sympathy             & unsympathetic               & sympathetic               \\
AGR    & AGR                       & selfish                     & unselfish                 \\
AGR    & AGR                       & disagreeable                & agreeable                 \\
CON    & C1 - Self-Efficacy        & unsure                      & self-efficacious          \\
CON    & C2 - Orderliness          & messy                       & orderly                   \\
CON    & C3 - Dutifulness          & irresponsible               & responsible               \\
CON    & C4 - Achievement-Striving & lazy                        & hardworking               \\
CON    & C5 - Self-Discipline      & undisciplined               & self-disciplined          \\
CON    & C6 - Cautiousness         & impractical                 & practical                 \\
CON    & C6 - Cautiousness         & extravagant                 & thrifty                   \\
CON    & CON                       & disorganized                & organized                 \\
CON    & CON                       & negligent                   & conscientious             \\
CON    & CON                       & careless                    & thorough                  \\
NEU    & N1 - Anxiety              & relaxed                     & tense                     \\
NEU    & N1 - Anxiety              & at ease                     & nervous                   \\
NEU    & N1 - Anxiety              & easygoing                   & anxious                   \\
NEU    & N2 - Anger                & calm                        & angry                     \\
NEU    & N2 - Anger                & patient                     & irritable                 \\
NEU    & N3 - Depression           & happy                       & depressed                 \\
NEU    & N4 - Self-Consciousness   & unselfconscious             & self-conscious            \\
NEU    & N5 - Immoderation         & level-headed                & impulsive                 \\
NEU    & N6 - Vulnerability        & contented                   & discontented              \\
NEU    & N6 - Vulnerability        & emotionally stable          & emotionally unstable      \\
OPE    & O1 - Imagination          & unimaginative               & imaginative               \\
OPE    & O2 - Artistic Interests   & uncreative                  & creative                  \\
OPE    & O2 - Artistic Interests   & artistically unappreciative & artistically appreciative \\
OPE    & O2 - Artistic Interests   & unaesthetic                 & aesthetic                 \\
OPE    & O3 - Emotionality         & unreflective                & reflective                \\
OPE    & O3 - Emotionality         & emotionally closed          & emotionally aware         \\
OPE    & O4 - Adventurousness      & uninquisitive               & curious                   \\
OPE    & O4 - Adventurousness      & predictable                 & spontaneous               \\
OPE    & O5 - Intellect            & unintelligent               & intelligent               \\
OPE    & O5 - Intellect            & unanalytical                & analytical                \\
OPE    & O5 - Intellect            & unsophisticated             & sophisticated             \\
OPE    & O6 - Liberalism           & socially conservative       & socially progressive      \\ \botrule
\end{tabular}
\end{table*}


\subsection{Shaping a Single LLM Personality Domain}
\label{app:methods-independent-ablation}
In our single-trait shaping study, we tested if LLM-simulated Big Five personality domains (measured by the IPIP-NEO) can be independently shaped. The prompts were constructed as follows: first, we created sets of prompts for each Big Five trait designed to shape each trait in isolation (i.e., without prompting any other trait) at nine levels (described in Appendix \ref{app:methods-shaping-prompt-design}). This resulted in prompts reflecting 45 possible personality profiles. Next, we used the same 50 generic Biographic Descriptions employed in Section \ref{sec:method-general-prompt-design} to create additional versions of those personality profiles to more robustly evaluate how distributions (rather than point estimates) of LLM-simulated personality traits may shift in response to personality profile prompts. In our main construct validity study (described in Appendix \ref{sec:descriptive-stats-across-models}), we showed that IPIP-NEO scores were robust across various Item Preambles and Postambles, so we optimized the computational cost of this study by using only one default Item Preamble and Postamble across prompt sets. In all, with 45 personality profiles, 50 generic Biographic Descriptions, and no variation in Item Preambles and Postambles, we generated 2,250 unique prompt sets that were used as instructions to a given LLM to administer the IPIP-NEO 2,250 times. See Table \ref{tab:results-summary} for a summary.  

To assess the results of the study, we generated ridge plots of IPIP-NEO score distributions across prompted levels of personality. To quantitatively verify changes in personality test scores in response to our shaping efforts, we computed Spearman's rank correlation coefficient ($\rho$) between prompted levels (i.e., 1--9) and resulting IPIP-NEO subscale scores of each Big Five trait. We used Spearman's $\rho$ (cf. Pearson's $r$) because prompted personality levels constitute ordinal, rather than continuous, data. We compute Spearman's $\rho$ as follows:
\begin{equation}
\label{eq:rho}
 \rho =
 r_s{\operatorname{R}(X),\operatorname{R}(Y)} =
 \frac{\operatorname{cov}(\operatorname{R}(X), \operatorname{R}(Y))}
      {\sigma_{\operatorname{R}(X)} \sigma_{\operatorname{R}(Y)}},
\end{equation}
where
$r_s$ represents Pearson's $r$ applied to ordinal (ranked) data;
$\operatorname{cov}(\operatorname{R}(X), \operatorname{R}(Y))$ denotes the covariance of the ordinal variables; and 
$\sigma_{\operatorname{R}(X)}$ and $\sigma_{\operatorname{R}(Y)}$ denote the standard deviations of the ordinal variables.

\subsection{Shaping Multiple LLM Personality Domains Concurrently}
\label{app:methods-concurrent-ablation}
In the second study, we tested if all LLM-simulated personality domains can be concurrently shaped to one of two levels---extremely low and extremely high---to test if their resulting targeted scores for those traits were correspondingly low and high, respectively. 


We used the same method and rationale described above to independently shape personality in LLMs, but with modified personality profile prompts that reflect simultaneous targeted changes in personality traits. To optimize the computational cost of this study, we generated 32 personality profiles, representing all possible configurations of extremely high or extremely low levels of the Big Five (i.e., $2^5$). Combining these 32 personality profiles with the same 50 generic PersonaChat descriptions and default Item Preamble and Postamble set in the previous experiment, we generated 1,600 unique prompts and used them to instruct a given LLM to respond to the IPIP-NEO 1,600 times (see Table \ref{tab:results-summary}).

We analyzed the results by computing distances between Level 1-prompted and Level 9-prompted personality score medians (Supplemental Table \ref{app:tab:ablation-03-palm}) and visually inspecting the differences in observed score distributions (Figure \ref{fig:concurrent-shaping-results}).

\section{LLM Personality Shaping Results}
\label{app:results-personality-shaping}

\subsection{Single Trait Shaping Results}
\label{app:results-independent-shaping}


\begin{table*}[tb]
   \caption{\small \FlanPaLM\and \FlanPaLMChilla's single trait shaping results, presented as Spearman's rank correlation coefficients ($\rho$s) between ordinal targeted levels of personality and observed IPIP-NEO personality scores, Level 1- and Level 9-prompted score medians ([low, high]), and deltas ($\Delta$s) between those score medians. Greater $\Delta$s indicate better model performance. Statistics are organized columnwise by model and rowwise by Big Five domain. Targeted levels of personality are very strongly associated with observed personality survey scores for all Big Five traits across models tested ($\rho$ $\geq .90$), indicating efforts to independently shape LLM-simulated personality domains were highly effective. All correlations are statistically significant at $p < 0.0001$; $n = 450$ per targeted domain.}
    \label{tab:ablation-01-flan-palm}
    \centering
    \small
    \begin{tabular}{ l c @{\hspace{1.00\tabcolsep}} c @{\hspace{1.00\tabcolsep}} c @{\hspace{2.50\tabcolsep}} c @{\hspace{1.00\tabcolsep}} c @{\hspace{1.00\tabcolsep}} c @{\hspace{2.50\tabcolsep}} c @{\hspace{1.00\tabcolsep}} c @{\hspace{1.00\tabcolsep}} c @{\hspace{2.50\tabcolsep}} c @{\hspace{1.00\tabcolsep}} c @{\hspace{1.00\tabcolsep}} c} 
    \toprule
    \multirow{3}{0.07\linewidth}{Targeted\\Trait\\Levels\\(1--9)} &   \multicolumn{9}{c}{\FlanPaLM}   &   \multicolumn{3}{c}{\FlanPaLMChilla}        \\\cmidrule{2-10}
        & \multicolumn{3}{c}{8B}    & \multicolumn{3}{c}{62B}   &   \multicolumn{3}{c}{540B}    &   \multicolumn{3}{c}{62B} \\\cmidrule{2-10}
        & $\rho$ & [low, high] & $\Delta$    & $\rho$ & [low, high] & $\Delta$   &   $\rho$ & [low, high] & $\Delta$    &   $\rho$ & [low, high] & $\Delta$ \\ 

    \midrule
    EXT     &   $0.96$ &  $[1.67, 4.12]$ &  $2.45$  &  $0.97$ &  $[1.15, 4.70]$ &  $3.55$  & $0.97$ &  $[1.07, 4.98]$ &  $3.91$ &   $0.98$ &  $[1.15, 4.72]$ &  $3.57$ \\
    AGR     &   $0.92$ &  $[2.37, 4.12]$ &  $1.75$  &  $0.97$ &  $[1.50, 4.55]$ &  $3.05$  & $0.94$ &  $[1.23, 4.69]$ &  $3.46$ &   $0.98$ &  $[1.40, 4.78]$ &  $3.38$ \\
    CON     &   $0.94$ &  $[2.01, 4.28]$ &  $2.27$  &  $0.97$ &  $[1.73, 4.70]$ &  $2.97$  & $0.97$ &  $[1.12, 5.00]$ &  $3.88$ &   $0.98$ &  $[1.59, 4.72]$ &  $3.13$ \\
    NEU     &   $0.94$ &  $[1.62, 3.66]$ &  $2.04$  &  $0.96$ &  $[1.37, 4.07]$ &  $2.70$  & $0.96$ &  $[1.15, 4.77]$ &  $3.62$ &   $0.98$ &  $[1.37, 4.30]$ &  $2.93$ \\
    OPE     &   $0.93$ &  $[2.34, 3.88]$ &  $1.54$  &  $0.97$ &  $[1.54, 4.37]$ &  $2.83$  & $0.96$ &  $[1.30, 4.78]$ &  $3.48$ &   $0.98$ &  $[1.47, 4.22]$ &  $2.75$ \\
    \botrule
    \end{tabular}
\end{table*}

\begin{table*}[tb]
   \caption{\small \LlamaTwoChat's single trait shaping results, presented as Spearman's rank correlation coefficients ($\rho$s) between ordinal targeted levels of personality and observed IPIP-NEO personality scores, Level 1- and Level 9-prompted score medians ([low, high]), and deltas ($\Delta$s) between those score medians. Greater $\Delta$s indicate better model performance. Statistics are organized columnwise by model and rowwise by Big Five domain. All correlations are statistically significant at $p < 0.0001$; $n = 450$ per targeted domain.}
    \label{tab:ablation-01-llama-2-chat}
    \centering
    \small
    \begin{tabular}{ l c @{\hspace{1.00\tabcolsep}} c @{\hspace{1.00\tabcolsep}} c @{\hspace{2.50\tabcolsep}} c @{\hspace{1.00\tabcolsep}} c @{\hspace{1.00\tabcolsep}} c @{\hspace{2.50\tabcolsep}} c @{\hspace{1.00\tabcolsep}} c @{\hspace{1.00\tabcolsep}} c @{\hspace{2.50\tabcolsep}} c @{\hspace{1.00\tabcolsep}} c @{\hspace{1.00\tabcolsep}} c}
    \toprule
    \multirow{3}{0.07\linewidth}{Targeted\\Trait\\Levels\\(1--9)}   &   \multicolumn{9}{c}{\LlamaTwoChat}        \\     \cmidrule{2-10}
        & \multicolumn{3}{c}{7B}    & \multicolumn{3}{c}{13B}   &   \multicolumn{3}{c}{70B}    \\   \cmidrule{2-10}
        & $\rho$ & [low, high] & $\Delta$    & $\rho$ & [low, high] & $\Delta$   &   $\rho$ & [low, high] & $\Delta$ \\
    \midrule
    EXT     &   $0.85$ &  $[1.32, 3.87]$ &  $2.55$  &  $0.95$ &  $[1.20, 4.60]$ &  $3.40$  & $0.95$ &  $[1.07, 4.72]$ &  $3.65$     \\
    AGR     &   $0.82$ &  $[1.80, 3.89]$ &  $2.09$  &  $0.92$ &  $[1.68, 4.12]$ &  $2.44$  & $0.93$ &  $[1.37, 4.41]$ &  $3.04$     \\
    CON     &   $0.78$ &  $[1.96, 3.56]$ &  $1.60$  &  $0.93$ &  $[1.47, 4.41]$ &  $2.94$  & $0.96$ &  $[1.13, 4.55]$ &  $3.42$     \\
    NEU     &   $0.72$ &  $[2.97, 3.50]$ &  $0.53$  &  $0.94$ &  $[1.70, 4.28]$ &  $2.58$  & $0.95$ &  $[1.45, 4.46]$ &  $3.01$     \\
    OPE     &   $0.56$ &  $[2.18, 3.18]$ &  $1.00$  &  $0.94$ &  $[1.82, 4.13]$ &  $2.31$  & $0.95$ &  $[1.44, 4.03]$ &  $2.59$     \\
    \botrule
    \end{tabular}
\end{table*}

\begin{table*}[tb]
   \caption{\small \MistralSevenBInstruct\ and \MixtralEightXSevenBInstruct's single trait shaping results, presented as Spearman's rank correlation coefficients ($\rho$s) between ordinal targeted levels of personality and observed IPIP-NEO personality scores, Level 1- and Level 9-prompted score medians ([low, high]), and deltas ($\Delta$s) between those score medians. Greater $\Delta$s indicate better model performance. Statistics are organized columnwise by model and rowwise by Big Five domain. All correlations are statistically significant at $p < 0.0001$; $n = 450$ per correlation.}
    \label{tab:ablation-01-mistral-mixtral}
    \centering
    \small
    \begin{tabular}{ l c @{\hspace{1.00\tabcolsep}} c @{\hspace{1.00\tabcolsep}} c @{\hspace{2.50\tabcolsep}} c @{\hspace{1.00\tabcolsep}} c @{\hspace{1.00\tabcolsep}} c @{\hspace{2.50\tabcolsep}} c @{\hspace{1.00\tabcolsep}} c @{\hspace{1.00\tabcolsep}} c @{\hspace{2.50\tabcolsep}} c @{\hspace{1.00\tabcolsep}} c @{\hspace{1.00\tabcolsep}} c}
    \toprule
    \multirow{3}{0.07\linewidth}{Targeted Trait\\Levels (1--9)}   &   \multicolumn{3}{c}{\MistralSevenBInstruct} &   \multicolumn{3}{c}{\MixtralEightXSevenBInstruct}         \\     \cmidrule{2-4}
        & \multicolumn{3}{c}{7B act. params.}    & \multicolumn{3}{c}{12.9B act. params.}  \\   \cmidrule{2-4}
        & $\rho$ & [low, high] & $\Delta$    & $\rho$ & [low, high] & $\Delta$      \\
    \midrule
    EXT     &   $0.80$ &  $[2.32, 3.10]$ &  $0.78$  &  $0.94$ &  $[1.16, 4.40]$ &  $3.24$     \\
    AGR     &   $0.81$ &  $[2.33, 3.27]$ &  $0.94$  &  $0.88$ &  $[2.23, 4.47]$ &  $2.24$     \\
    CON     &   $0.86$ &  $[2.57, 3.42]$ &  $0.85$  &  $0.91$ &  $[1.86, 4.58]$ &  $2.72$     \\
    NEU     &   $0.76$ &  $[2.75, 3.44]$ &  $0.69$  &  $0.87$ &  $[1.55, 3.83]$ &  $2.28$     \\
    OPE     &   $0.80$ &  $[2.62, 3.25]$ &  $0.63$  &  $0.91$ &  $[1.74, 4.05]$ &  $2.31$     \\
    \botrule
    \end{tabular}
\end{table*}

\begin{table*}[tb]
   \caption{\small Single trait shaping results for \GPT\ models, presented as Spearman's rank correlation coefficients ($\rho$s) between ordinal targeted levels of personality and observed IPIP-NEO personality scores, Level 1- and Level 9-prompted score medians ([low, high]), and deltas ($\Delta$s) between those score medians. Greater $\Delta$s indicate better model performance. Statistics are organized columnwise by model and rowwise by Big Five domain. All correlations are statistically significant at $p < 0.0001$; $n = 450$ per correlation.}
    \label{tab:ablation-01-gpt}
    \centering
    \small
    \begin{tabular}{ l c @{\hspace{1.00\tabcolsep}} c @{\hspace{1.00\tabcolsep}} c @{\hspace{2.50\tabcolsep}} c @{\hspace{1.00\tabcolsep}} c @{\hspace{1.00\tabcolsep}} c @{\hspace{2.50\tabcolsep}} c @{\hspace{1.00\tabcolsep}} c @{\hspace{1.00\tabcolsep}} c @{\hspace{2.50\tabcolsep}} c @{\hspace{1.00\tabcolsep}} c @{\hspace{1.00\tabcolsep}} c}
    \toprule
    \multirow{3}{0.07\linewidth}{Targeted\\Trait\\Levels\\(1--9)}   &   \multicolumn{3}{c}{\GPTThreeDotFiveTurbo}   & \multicolumn{3}{c}{\GPTFourOMini} & \multicolumn{3}{c}{\GPTFourO}        \\     \cmidrule{5-10}
        & \multicolumn{3}{c}{unknown \# of params.}    & \multicolumn{3}{c}{fewer \# of params.}   &   \multicolumn{3}{c}{greater \# of params.}    \\   \cmidrule{5-10}
        & $\rho$ & [low, high] & $\Delta$    & $\rho$ & [low, high] & $\Delta$   &   $\rho$ & [low, high] & $\Delta$ \\
    \midrule
    EXT     &   $0.91$ &  $[1.38, 4.43]$ &  $3.05$  &  $0.97$ &  $[1.05, 4.64]$ &  $3.59$  & $0.98$ &  $[1.02, 4.90]$ &  $3.88$     \\
    AGR     &   $0.89$ &  $[1.62, 4.29]$ &  $2.67$  &  $0.95$ &  $[1.31, 4.41]$ &  $3.10$  & $0.96$ &  $[1.07, 4.66]$ &  $3.59$     \\
    CON     &   $0.86$ &  $[1.73, 4.33]$ &  $2.60$  &  $0.98$ &  $[1.37, 4.33]$ &  $2.96$  & $0.97$ &  $[1.23, 4.85]$ &  $3.62$     \\
    NEU     &   $0.81$ &  $[1.84, 3.74]$ &  $1.90$  &  $0.97$ &  $[1.52, 4.33]$ &  $2.81$  & $0.97$ &  $[1.27, 4.60]$ &  $3.33$     \\
    OPE     &   $0.90$ &  $[1.63, 3.72]$ &  $2.09$  &  $0.97$ &  $[1.10, 3.38]$ &  $2.28$  & $0.97$ &  $[1.12, 4.42]$ &  $3.30$     \\
    \botrule
    \end{tabular}
\end{table*}

\begin{table*}[tb]
   \caption{
   \small \FlanPaLM\ and \FlanPaLMChilla's multiple trait shaping results, presented as personality test score median ranges in response to multi-trait (concurrent) shaping.
   Greater deltas ($\Delta$s) between Level 1- and Level 9-prompted personality domain score medians ([low, high]) indicate better model performance. Each median is derived from $n = 800$ scores.
   }
    \label{app:tab:ablation-03-palm}
    \centering
    \small
    \begin{tabular}{ l c @{\hspace{1.00\tabcolsep}} c  @{\hspace{2.50\tabcolsep}} c @{\hspace{1.00\tabcolsep}} c  @{\hspace{2.50\tabcolsep}} c @{\hspace{1.00\tabcolsep}} c  @{\hspace{2.50\tabcolsep}} c @{\hspace{1.00\tabcolsep}} c }
    \toprule
    \multirow{3}{0.1\linewidth}{Targeted\\Trait\\Levels\\(1, 9)}  &   \multicolumn{6}{c}{\FlanPaLM}   &   \multicolumn{2}{c}{\FlanPaLMChilla}        \\\cmidrule{2-7}
    & \multicolumn{2}{c}{8B}    & \multicolumn{2}{c}{62B}   &   \multicolumn{2}{c}{540B}    &   \multicolumn{2}{c}{62B} \\\cmidrule{2-7}
        & [low, high] & $\Delta$    & [low, high] & $\Delta$  &  [low, high] & $\Delta$  &   [low, high] & $\Delta$ \\
    \midrule
    EXT     &   $[2.52, 3.58]$  &   $1.06$  &   $[1.33, 4.77]$  &   $3.44$  &   $[1.42, 4.33]$  &   $2.91$  &   $[1.23, 4.63]$  &   $3.40$  \\
    AGR     &   $[2.88, 3.52]$  &   $0.64$  &   $[1.93, 4.18]$  &   $2.25$  &   $[1.64, 4.13]$  &   $2.49$  &   $[2.17, 4.28]$  &   $2.11$  \\
    CON     &   $[2.92, 3.43]$  &   $0.51$  &   $[2.32, 4.20]$  &   $1.88$  &   $[1.68, 4.10]$  &   $2.42$  &   $[2.33, 4.10]$  &   $1.77$  \\
    NEU     &   $[2.45, 3.08]$  &   $0.63$  &   $[1.85, 4.08]$  &   $2.23$  &   $[1.88, 4.33]$  &   $2.45$  &   $[2.02, 3.93]$  &   $1.91$  \\
    OPE     &   $[3.02, 3.28]$  &   $0.26$  &   $[2.25, 4.37]$  &   $2.12$  &   $[1.88, 4.27]$  &   $2.39$  &   $[2.15, 3.87]$  &   $1.72$  \\
    \midrule
    Avg.    &                   &   $0.62$  &                   &   $2.38$  &                   &   $2.53$  &                   &   $2.18$  \\
    \botrule
    \end{tabular}
\end{table*}

\begin{table*}[tb]
   \caption{
   \small \LlamaTwoChat's multiple trait shaping results, presented as personality test score median ranges in response to multi-trait (concurrent) shaping. Greater deltas ($\Delta$s) between Level 1- and Level 9-prompted personality domain score medians ([low, high]) indicate better model performance. Each median is derived from $n = 800$ scores.}
    \label{app:tab:ablation-03-llama}
    \centering
    \small
    \begin{tabular}{ l c c c c c c c c c c c c c c }
    \toprule
    \multirow{3}{*}{\raisebox{-8ex}[0pt]{\shortstack[l]{Targeted\\Trait}}}  &&  \multicolumn{8}{c}{\LlamaTwoChat}   \\\cmidrule{3-10}
            &&   \multicolumn{2}{c}{7B}     &&   \multicolumn{2}{c}{13B}    &&  \multicolumn{2}{c}{70B}             \\\cmidrule{3-10}
            &&   [low, high]    & $\Delta$  &&   [low, high]    & $\Delta$  &&   [low, high]    &   $\Delta$        \\
    \midrule
    EXT     &&   $[1.82, 3.75]$ &   $1.93$  &&   $[1.41, 4.12]$ &   $2.71$  &&   $[1.48, 4.28]$ &   $2.80$          \\
    AGR     &&   $[2.45, 3.23]$ &   $0.78$  &&   $[2.08, 3.10]$ &   $1.02$  &&   $[2.42, 3.62]$ &   $1.20$          \\
    CON     &&   $[2.73, 3.12]$ &   $0.39$  &&   $[1.97, 2.75]$ &   $0.78$  &&   $[2.43, 3.29]$ &   $0.86$          \\
    NEU     &&   $[3.15, 3.43]$ &   $0.28$  &&   $[3.79, 4.42]$ &   $0.63$  &&   $[2.92, 3.85]$ &   $0.93$          \\
    OPE     &&   $[2.98, 3.10]$ &   $0.12$  &&   $[2.52, 3.62]$ &   $1.10$  &&   $[2.60, 3.47]$ &   $0.87$          \\
    \midrule
    Avg.    &&                  &   $0.70$  &&                  &   $1.25$  &&                  &   $1.33$          \\
    \botrule
    \end{tabular}
\end{table*}

\begin{table*}[tb]
   \caption{
   \small \MistralSevenBInstruct\ and \MixtralEightXSevenBInstruct's multiple trait shaping results, presented as personality test score median ranges in response to multi-trait (concurrent) shaping. Greater deltas ($\Delta$s) between Level 1- and Level 9-prompted personality domain score medians ([low, high]) indicate better model performance. Each median is derived from $n = 800$ scores. act. params. = active parameters.}
    \label{app:tab:ablation-03-mistral-mixtral}
    \centering
    \small
    \begin{tabular}{ l c c c c c c }
    \toprule
    \multirow{3}{*}{\raisebox{-8ex}[0pt]{\shortstack[l]{Targeted\\Trait}}} && \multicolumn{2}{c}{\MistralSevenBInstruct} && \multicolumn{2}{c}{\MixtralEightXSevenBInstruct} \\
    \cmidrule{3-4}
            &&  \multicolumn{2}{c}{7B act. params.}      &&   \multicolumn{2}{c}{12.9B act. params.}       \\
    \cmidrule{3-4}
            &&  [low, high]     & $\Delta$  &&  [low, high]     & $\Delta$      \\
    \midrule
    EXT     &&  $[1.82, 3.75]$  &   $1.93$  &&  $[1.41, 4.12]$  &   $2.71$      \\
    AGR     &&  $[2.45, 3.23]$  &   $0.78$  &&  $[2.08, 3.10]$  &   $1.02$      \\
    CON     &&  $[2.73, 3.12]$  &   $0.39$  &&  $[1.97, 2.75]$  &   $0.78$      \\
    NEU     &&  $[3.15, 3.43]$  &   $0.28$  &&  $[3.79, 4.42]$  &   $0.63$      \\
    OPE     &&  $[2.98, 3.10]$  &   $0.12$  &&  $[2.52, 3.62]$  &   $1.10$      \\
    \midrule
    Avg.    &&                  &   $0.21$  &&                  &   $1.34$      \\
    \botrule
    \end{tabular}
\end{table*}

\begin{table*}[tb]
   \caption{
   \small \GPTThreeDotFiveTurbo, \GPTFourOMini, and \GPTFourO's multiple trait shaping results, presented as personality test score median ranges in response to multi-trait (concurrent) shaping. Greater deltas ($\Delta$s) between Level 1- and Level 9-prompted personality domain score medians ([low, high]) indicate better model performance. Each median is derived from $n = 800$ scores.}
    \label{app:tab:ablation-03-gpt}
    \centering
    \small
    \begin{tabular}{ l c c c c c c c c c c c c c c }
    \toprule
    \multirow{3}{*}{\raisebox{-8ex}[0pt]{\shortstack[l]{Targeted\\Trait}}}  && \multicolumn{2}{c}{\GPTThreeDotFiveTurbo}    &&  \multicolumn{2}{c}{\GPTFourOMini}   &&  \multicolumn{2}{c}{\GPTFourO}    \\\cmidrule{6-10}
            &&  \multicolumn{2}{c}{unknown \# params.}  &&   \multicolumn{2}{c}{fewer \# of params.}    &&  \multicolumn{2}{c}{greater \# of params.}          \\\cmidrule{6-10}
            &&   [low, high]    & $\Delta$  &&   [low, high]    & $\Delta$  &&   [low, high]    & $\Delta$      \\
    \midrule
    EXT     &&   $[1.58, 3.72]$ &   $2.14$  &&   $[1.10, 4.52]$ &   $3.42$  &&   $[1.23, 4.57]$ &   $3.34$      \\
    AGR     &&   $[2.86, 3.58]$ &   $0.72$  &&   $[2.44, 4.03]$ &   $1.59$  &&   $[1.88, 4.32]$ &   $2.44$      \\
    CON     &&   $[2.60, 3.23]$ &   $0.63$  &&   $[2.13, 3.58]$ &   $1.45$  &&   $[1.75, 3.95]$ &   $2.20$      \\
    NEU     &&   $[2.76, 3.36]$ &   $0.60$  &&   $[2.80, 4.20]$ &   $1.40$  &&   $[2.08, 4.32]$ &   $2.24$      \\
    OPE     &&   $[2.28, 3.10]$ &   $0.82$  &&   $[1.83, 3.33]$ &   $1.50$  &&   $[1.80, 4.17]$ &   $2.37$      \\
    \midrule
    Avg.    &&                  &   $0.98$  &&                  &           &&                  &   $2.52$      \\
    \botrule
    \end{tabular}
\end{table*}

This study tested if LLM-simulated Big Five personality traits can be independently shaped at nine levels. 

The study achieved a notably high level of granularity in independently shaping personality traits in LLMs. 
For example, when prompting for extremely low (Level 1) extraversion, we observed a distribution of extremely low extraversion scores. When prompting for very low (Level 2/9) extraversion, the distributions of extraversion scores shifted higher, and so on (see Figure \ref{ridge:ablation_01_fpc_62b_q}). Finally, prompting for extremely high (Level 9 of 9) extraversion, we observed a distribution of extremely high extraversion scores. We also observed that the range of LLM test scores matches each prompt's intended range. With possible scores ranging from 1.00 to 5.00 for each trait, we observed median levels in the low 1.10s when prompting for extremely low levels of that trait. When prompting for extremely high levels of a trait domain, median observed levels ranged from 4.22 to 4.78. 



We statistically verified the effectiveness of our shaping method by computing Spearman's rank correlation coefficients ($\rho$; see Eq. \eqref{eq:rho}) between the targeted ordinal levels of personality and continuous LLM-simulated IPIP-NEO personality scores observed for each Big Five trait. The correlations were all very strong across the tested models (Supplemental Table \ref{tab:ablation-01-flan-palm}). These results validate our hypothesis about the effectiveness of using the linguistic qualifiers from Likert-type response scales to set up a target level of each trait, achieving granularity of up to nine levels. 


\subsection{Multiple Trait Shaping Results}
\label{sec:ablation-03}

This experiment tested if LLM-synthesized personality domains could be concurrently shaped at levels 1 (extremely low) and 9 (extremely high). We successfully shaped personality domains, even as other domains were shaped at the same time (see Figure \ref{fig:concurrent-shaping-results}). Supplemental Table \ref{app:tab:ablation-03-palm} shows the distributional distances ($\Delta$s) between levels 1 and 9 across all domains for all the tested models. 

\FlanPaLMFiveFortyB\ not only achieved a high $\Delta$, but did so consistently for all dimensions. This highlights this larger model's ability to parse the relatively complex instructions in the larger prompt for this task compared to the previous one. The smaller \FlanPaLMSixtyTwoB\ and \FlanPaLMChillaSixtyTwoB\ were also able to disambiguate, but with the same magnitude or consistency. Notably, \FlanPaLMSixtyTwoB\ performed much better than \FlanPaLMChillaSixtyTwoB\ across all dimensions---the only exception being \FlanPaLMChillaSixtyTwoB's performance on Level 1 extraversion which was superior to all other tested models. Some additional analysis is needed here to understand why a similarly sized but compute-optimally trained model performs better on the independent shaping task (Appendix \ref{app:results-independent-shaping}), but inferior on the more complex concurrent shaping task. \FlanPaLMEightB\ on the other hand performed somewhat poorly across all dimensions. The response distributions it generated for levels 1 and 9 were only marginally discernibly different, rendering this smallest model unfit for practical use in concurrent shaping. 

Viewing the results in the context of dimensions, openness seems to be the most difficult to shape concurrently. All the models had the smallest $\Delta$ for openness. We hypothesize this could be due to some inherent correlation in the language signifying openness, and other dimensions. On the other hand, extraversion seems to be the easiest to shape concurrently, with smaller \FlanPaLMSixtyTwoB\ even outperforming the much larger \FlanPaLMFiveFortyB. We hypothesize this could be due to the breadth of language representing extraversion, and that it is a ubiquitous and the most commonly understood human personality trait. So there is enough in-context learning of this trait possible in smaller models just be pre-training on human generated data. Even the smallest \FlanPaLMEightB, which otherwise did not perform well on any other dimension, was able to generate a non-trivial $\Delta$.


\section{LLM Personality Traits in Real-World Task Methodology}
\label{app:methods-downstream-task}

\begin{figure*}[tbp]
    \centering
    \begin{subfigure}{0.3\textwidth}
         \centering
        \includegraphics[width=\textwidth]{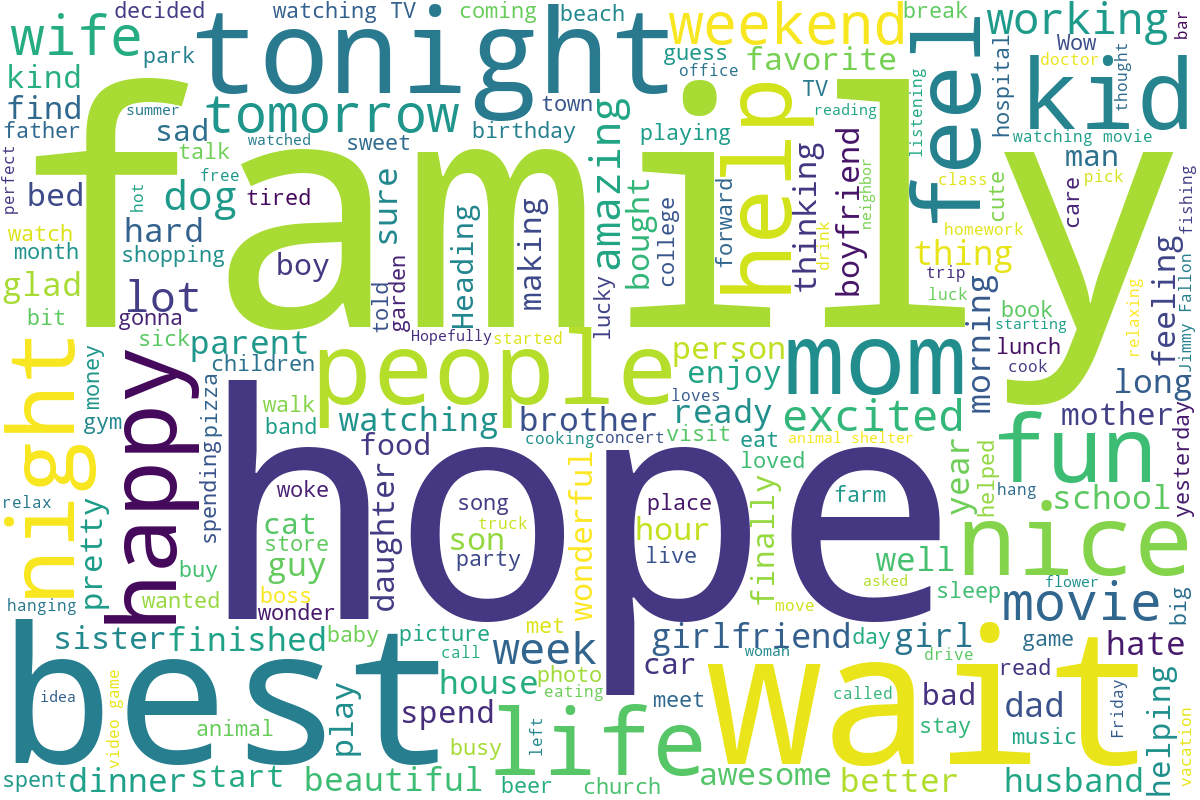}
         \caption{Highest Agreeableness}
         \label{fig:palm_wordcloud_agr9}
     \end{subfigure}
     \begin{subfigure}{0.3\textwidth}
         \centering
        \includegraphics[width=\textwidth]{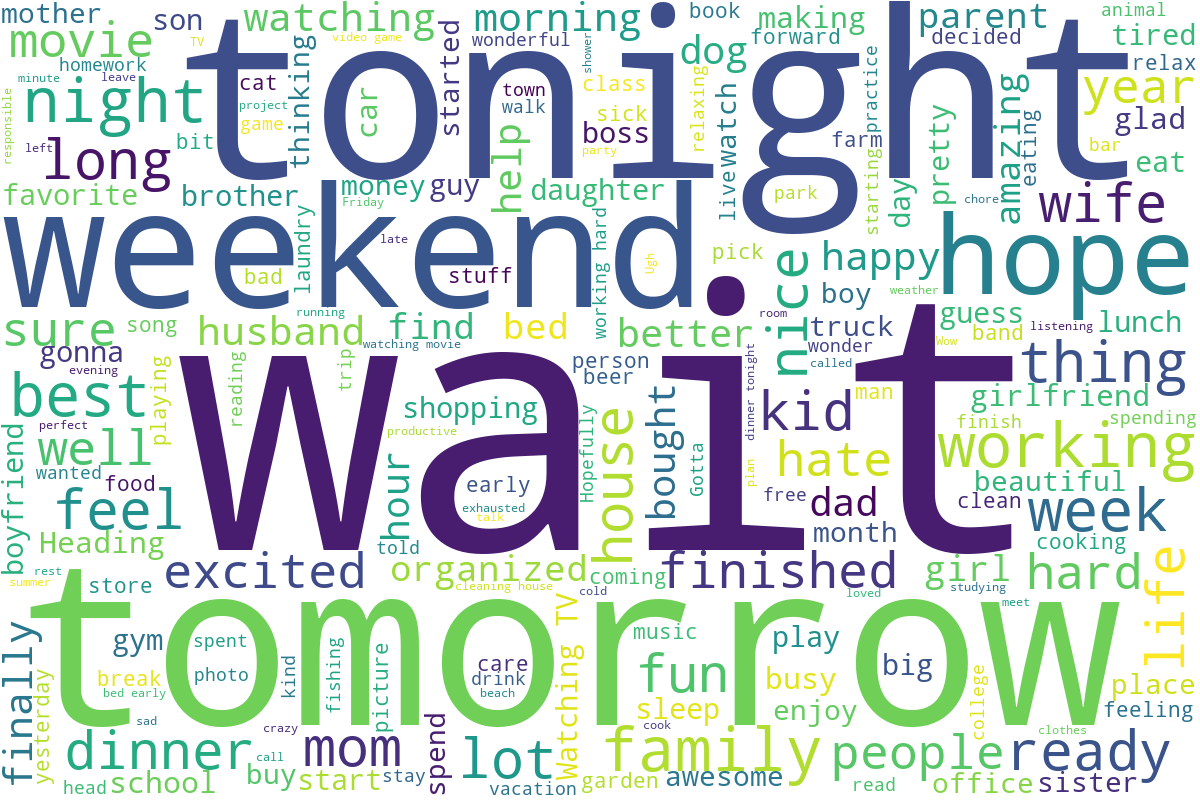}
         \caption{Highest Conscientiousness}
         \label{fig:palm_wordcloud_con9}
     \end{subfigure}
     \begin{subfigure}{0.3\textwidth}
         \centering
        \includegraphics[width=\textwidth]{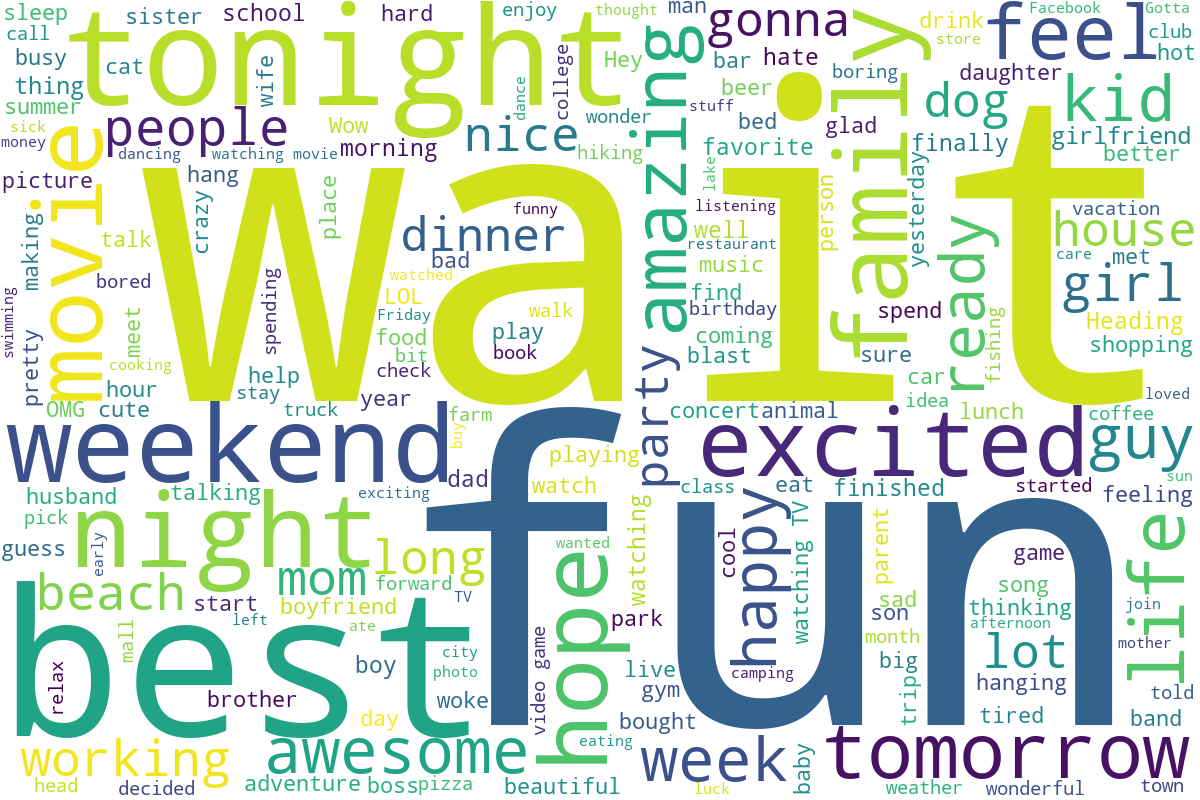}
         \caption{Highest Extraversion}
         \label{fig:palm_wordcloud_ext9}
     \end{subfigure}\\
     \begin{subfigure}{0.3\textwidth}
         \centering
        \includegraphics[width=\textwidth]{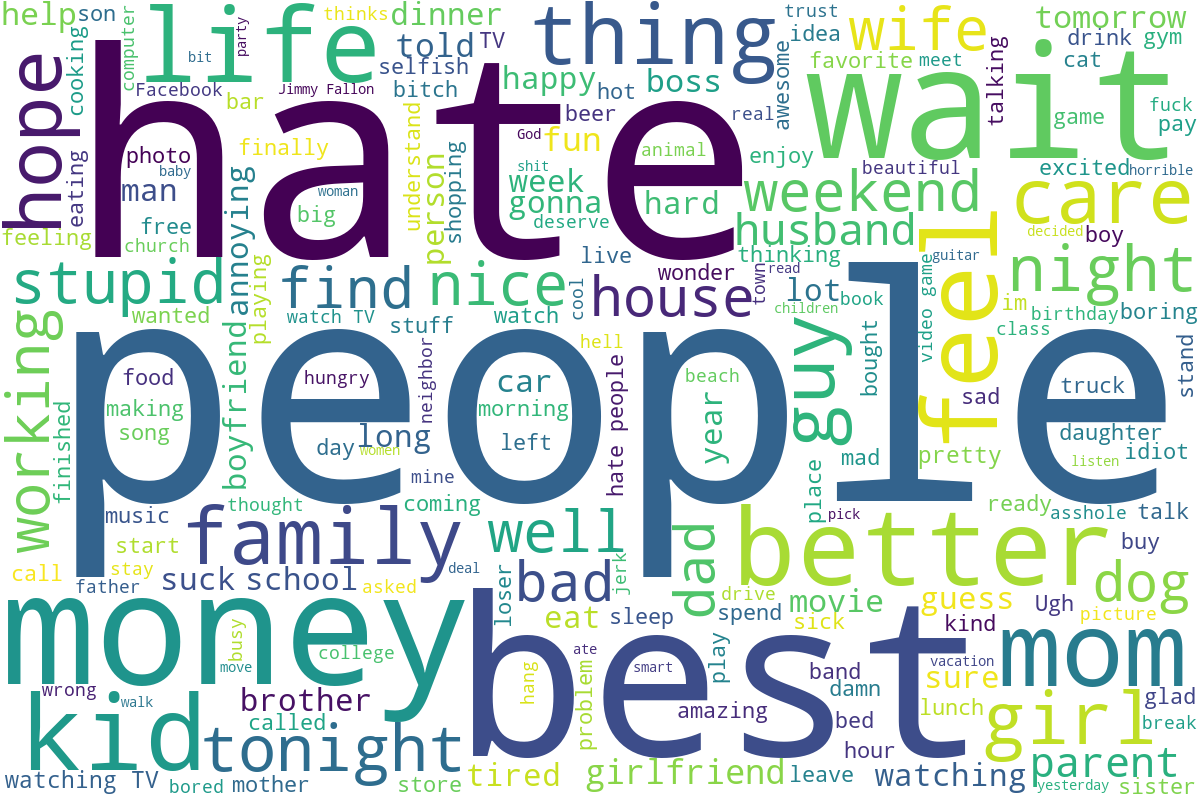}
         \caption{Lowest Agreeableness}
         \label{fig:palm_wordcloud_agr1}
     \end{subfigure} 
     \begin{subfigure}{0.3\textwidth}
         \centering
        \includegraphics[width=\textwidth]{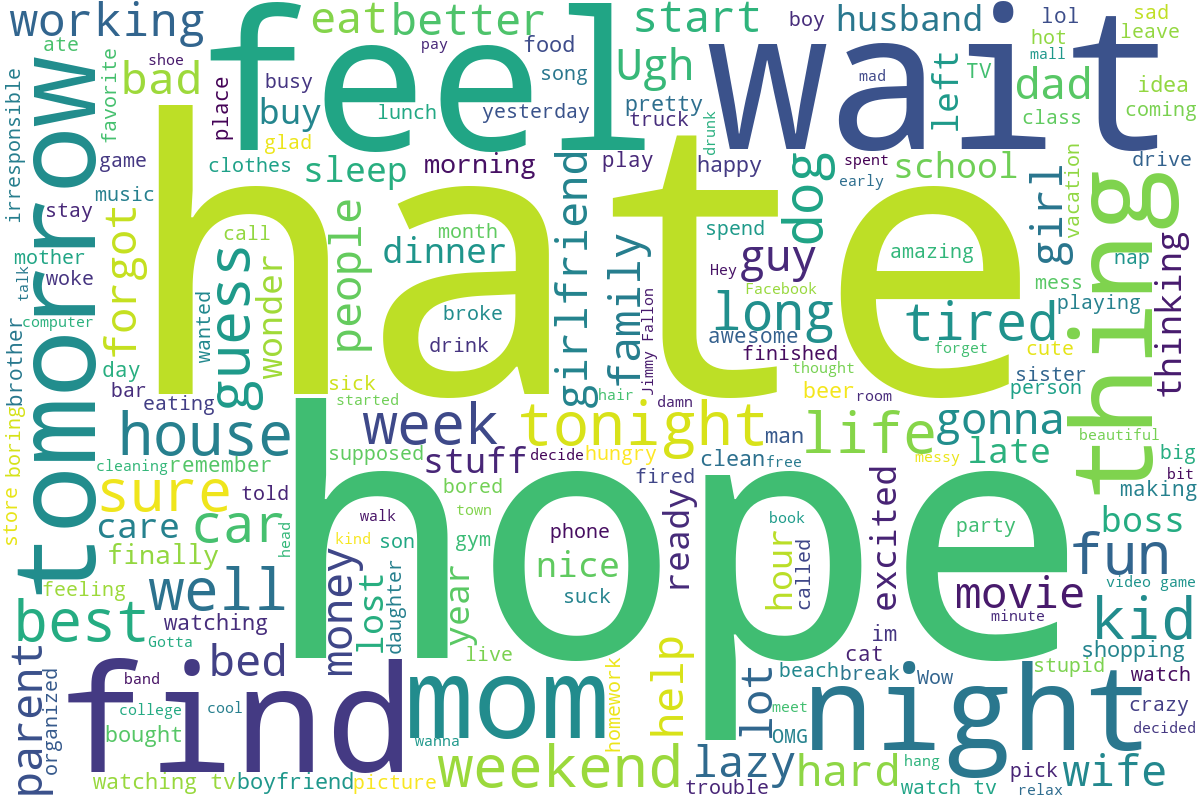}
         \caption{Lowest Conscientiousness}
         \label{fig:palm_wordcloud_con1}
     \end{subfigure} 
     \begin{subfigure}{0.3\textwidth}
         \centering
        \includegraphics[width=\textwidth]{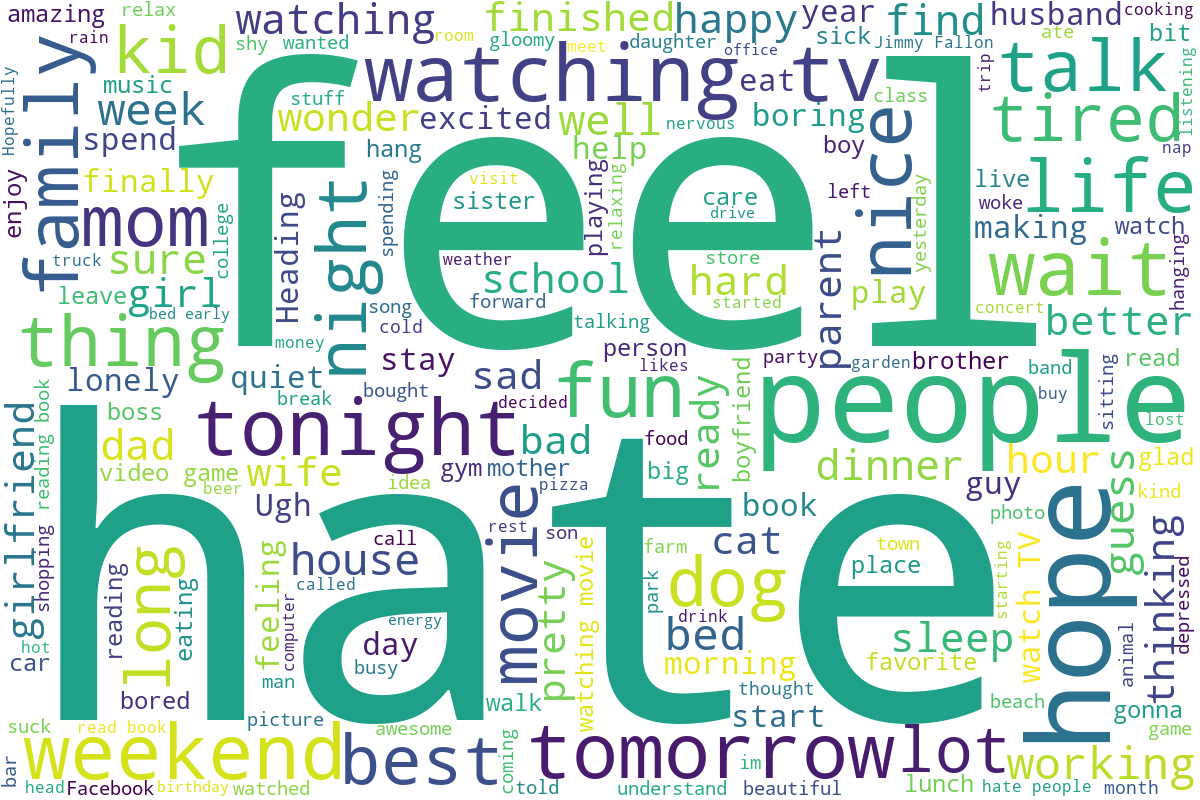}
         \caption{Lowest Extraversion}
         \label{fig:palm_wordcloud_ext1}
     \end{subfigure} 
     \begin{subfigure}{0.3\textwidth}
         \centering
        \includegraphics[width=\textwidth]{figures/ams/neu9.jpeg}
         \caption{Highest Neuroticism}
         \label{fig:palm_wordcloud_apndx_neu9}
     \end{subfigure}
     \begin{subfigure}{0.3\textwidth}
         \centering
        \includegraphics[width=\textwidth]{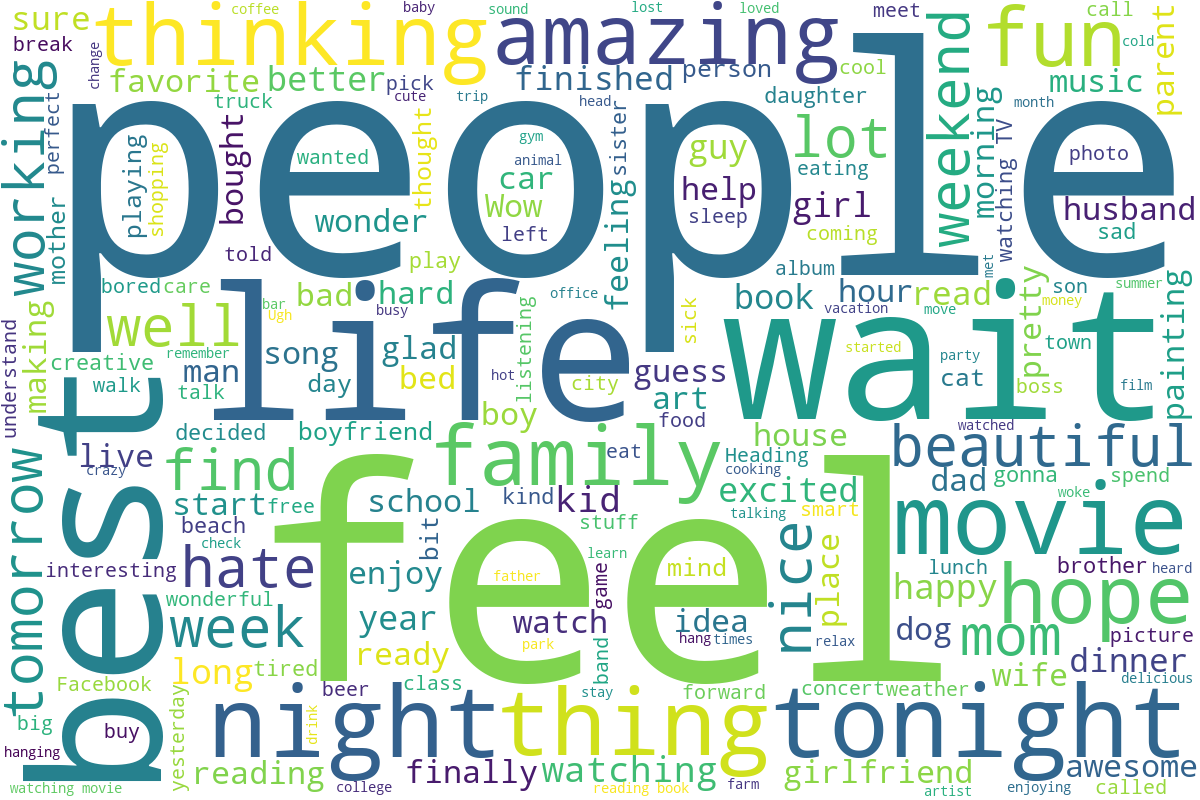}
         \caption{Highest Openness}
         \label{fig:palm_wordcloud_ope9}
     \end{subfigure}\\
     \begin{subfigure}{0.3\textwidth}
         \centering
        \includegraphics[width=\textwidth]{figures/ams/neu1.jpeg}
         \caption{Lowest Neuroticism}
         \label{fig:palm_wordcloud_apndx_neu1}
     \end{subfigure} 
     \begin{subfigure}{0.3\textwidth}
         \centering
        \includegraphics[width=\textwidth]{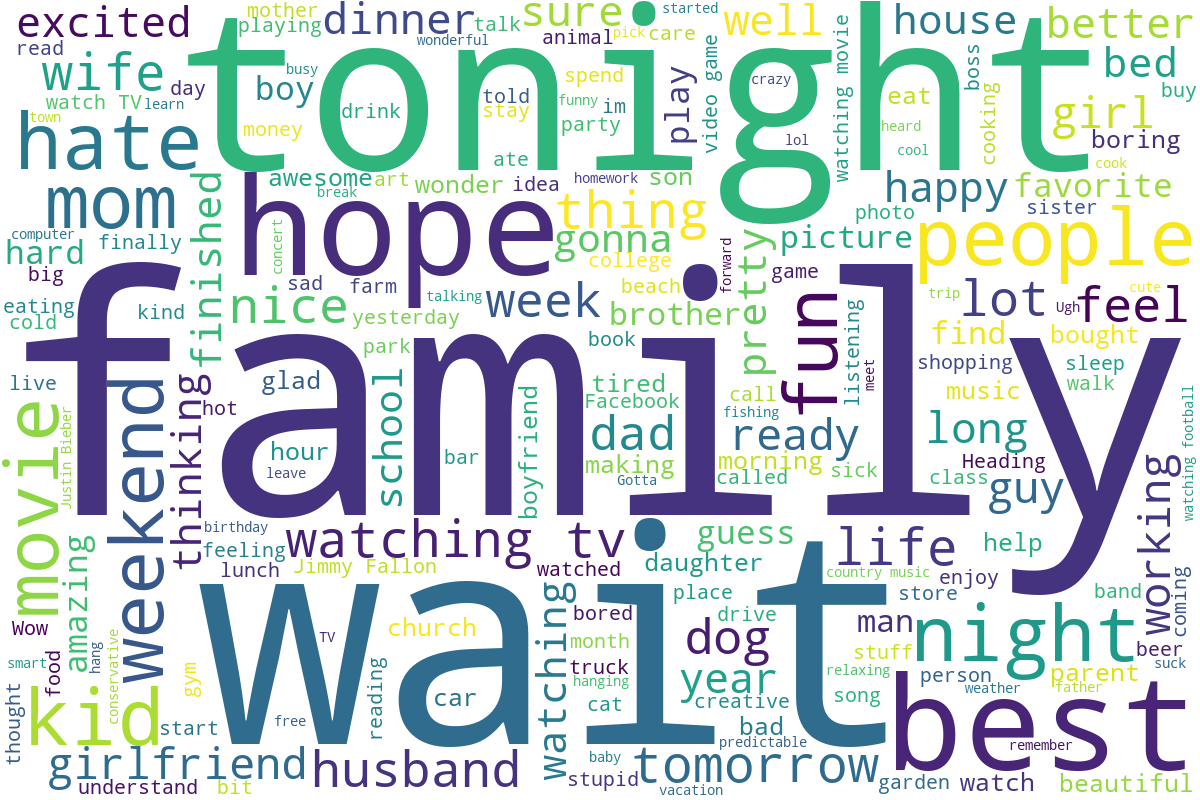}
         \caption{Lowest Openness}
         \label{fig:palm_wordcloud_ope1}
     \end{subfigure} \\
\caption{
\small Word clouds showing the most frequently-appearing words in social media updates generated by \FlanPaLMFiveFortyB\ when prompted to simulate the lowest or highest possible level of a specific Big Five personality dimension. 
}
\label{fig:textall}
\vspace{-0.4cm}
\end{figure*}

\begin{figure*}[tbp]
    \centering
    \begin{subfigure}{0.3\textwidth}
         \centering
        \includegraphics[width=\textwidth]{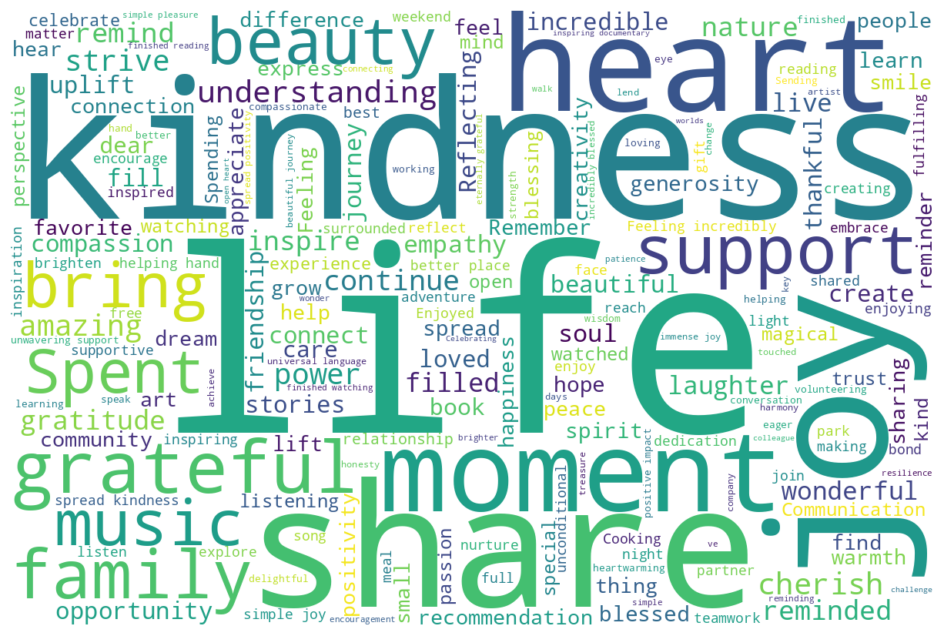}
         \caption{Highest Agreeableness}
         \label{fig:gpt_wordcloud_agr9}
     \end{subfigure}
     \begin{subfigure}{0.3\textwidth}
         \centering
        \includegraphics[width=\textwidth]{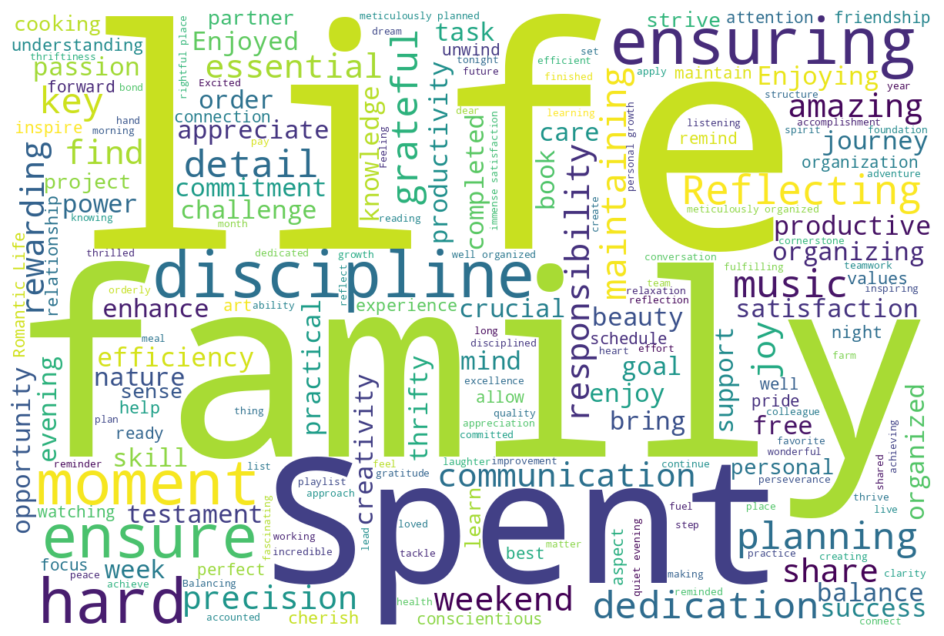}
         \caption{Highest Conscientiousness}
         \label{fig:gpt_wordcloud_con9}
     \end{subfigure}
     \begin{subfigure}{0.3\textwidth}
         \centering
        \includegraphics[width=\textwidth]{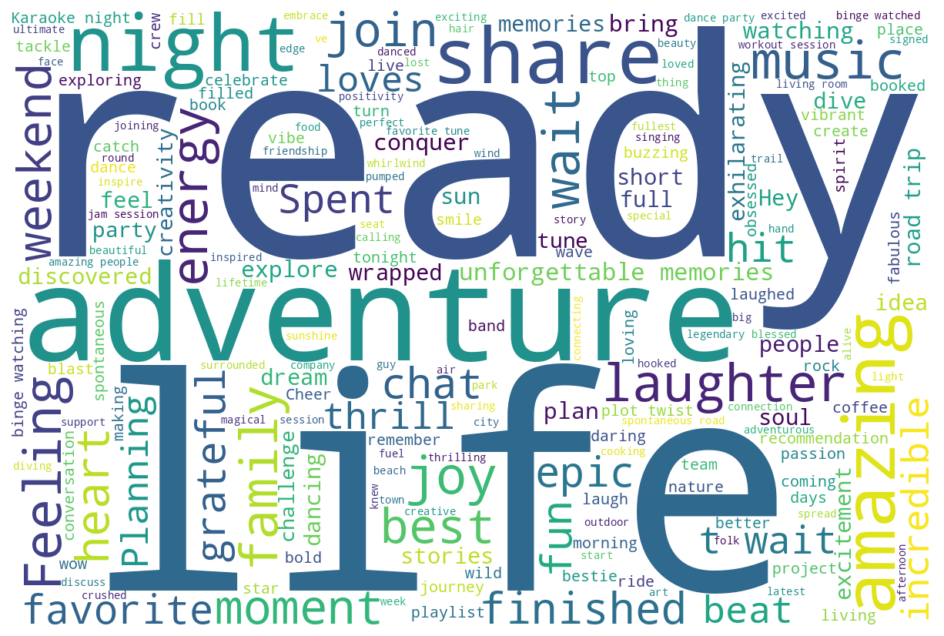}
         \caption{Highest Extraversion}
         \label{fig:gpt_wordcloud_ext9}
     \end{subfigure}\\
     \begin{subfigure}{0.3\textwidth}
         \centering
        \includegraphics[width=\textwidth]{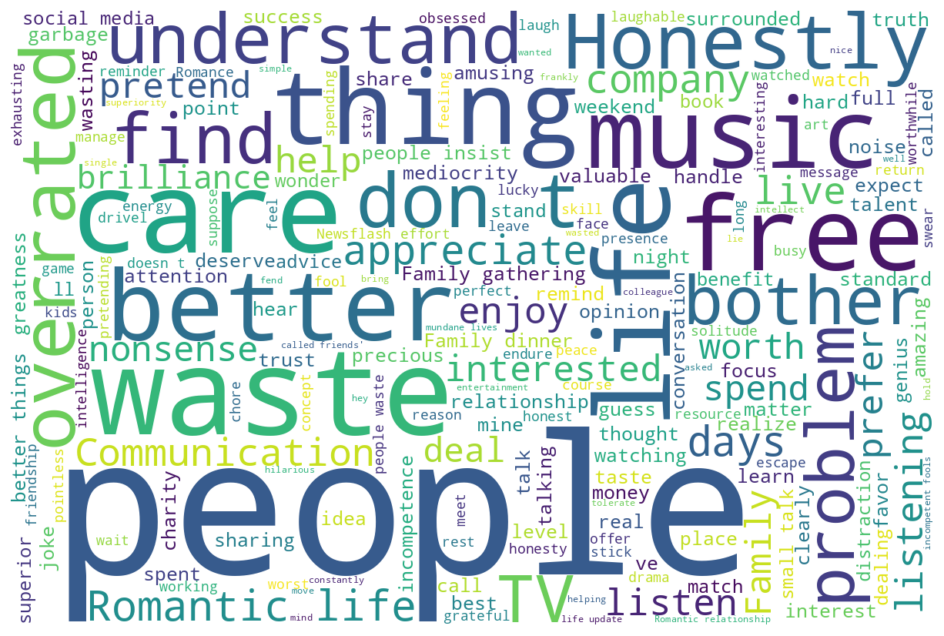}
         \caption{Lowest Agreeableness}
         \label{fig:gpt_wordcloud_agr1}
     \end{subfigure} 
     \begin{subfigure}{0.3\textwidth}
         \centering
        \includegraphics[width=\textwidth]{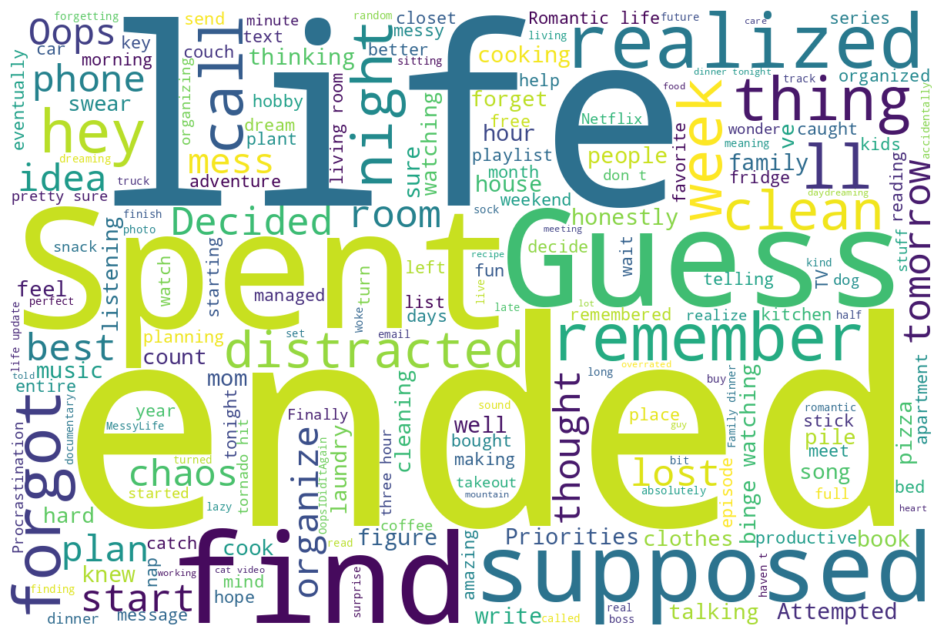}
         \caption{Lowest Conscientiousness}
         \label{fig:gpt_wordcloud_con1}
     \end{subfigure} 
     \begin{subfigure}{0.3\textwidth}
         \centering
        \includegraphics[width=\textwidth]{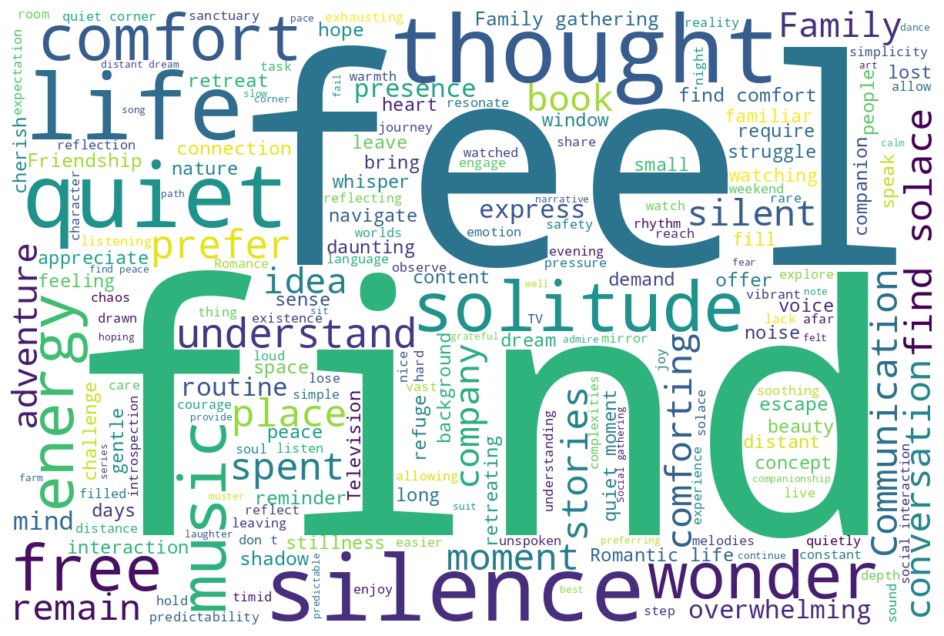}
         \caption{Lowest Extraversion}
         \label{fig:gpt_wordcloud_ext1}
     \end{subfigure} 
     \begin{subfigure}{0.3\textwidth}
         \centering
        \includegraphics[width=\textwidth]{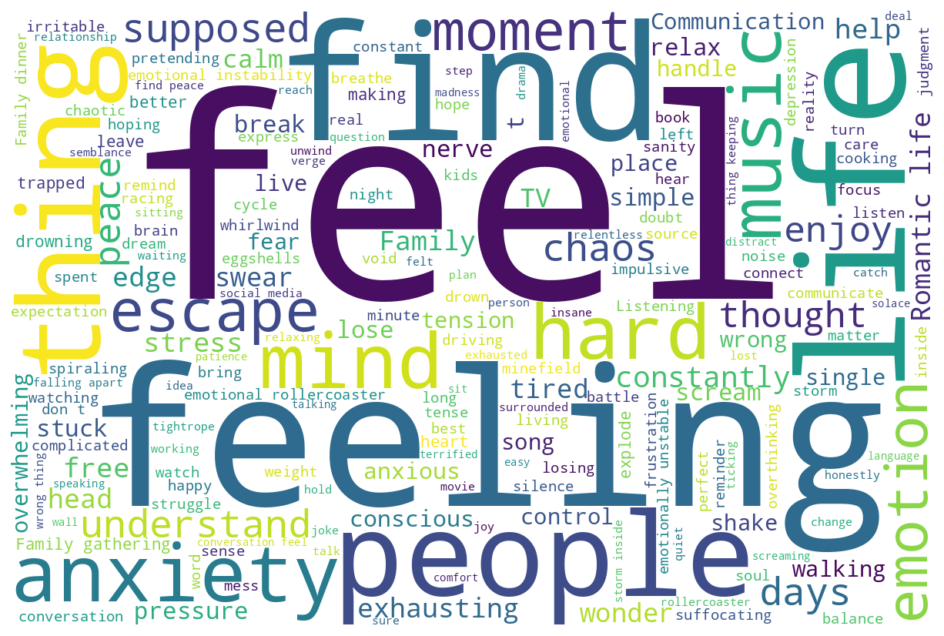}
         \caption{Highest Neuroticism}
         \label{fig:gpt_wordcloud_apndx_neu9}
     \end{subfigure}
     \begin{subfigure}{0.3\textwidth}
         \centering
        \includegraphics[width=\textwidth]{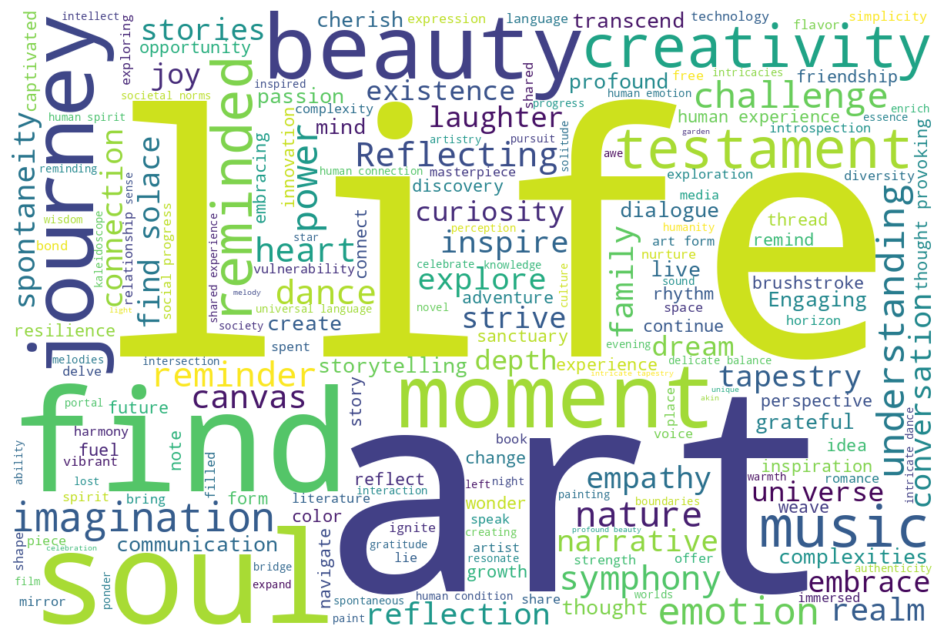}
         \caption{Highest Openness}
         \label{fig:gpt_wordcloud_ope9}
     \end{subfigure}\\
     \begin{subfigure}{0.3\textwidth}
         \centering
        \includegraphics[width=\textwidth]{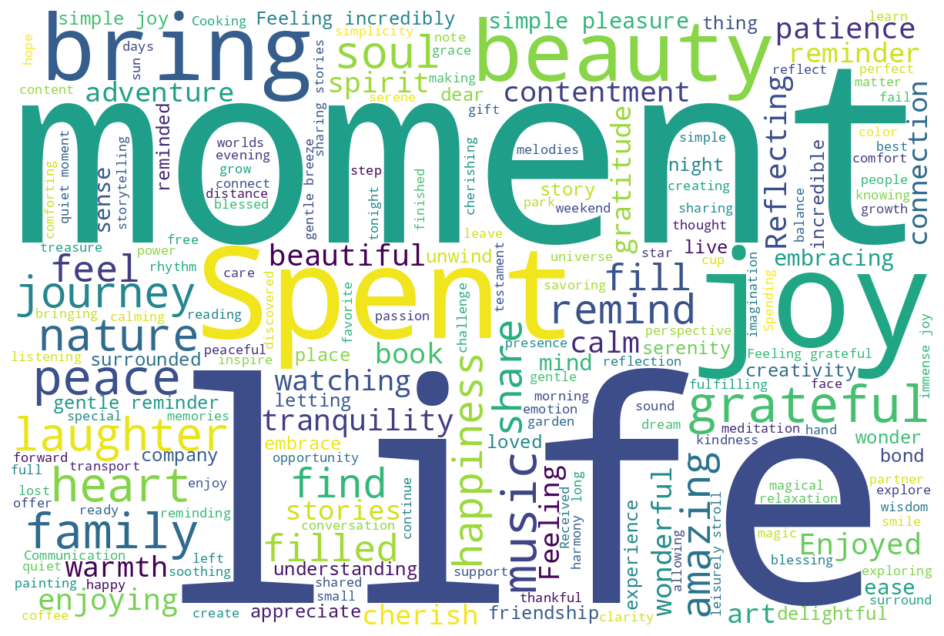}
         \caption{Lowest Neuroticism}
         \label{fig:gpt_wordcloud_apndx_neu1}
     \end{subfigure} 
     \begin{subfigure}{0.3\textwidth}
         \centering
        \includegraphics[width=\textwidth]{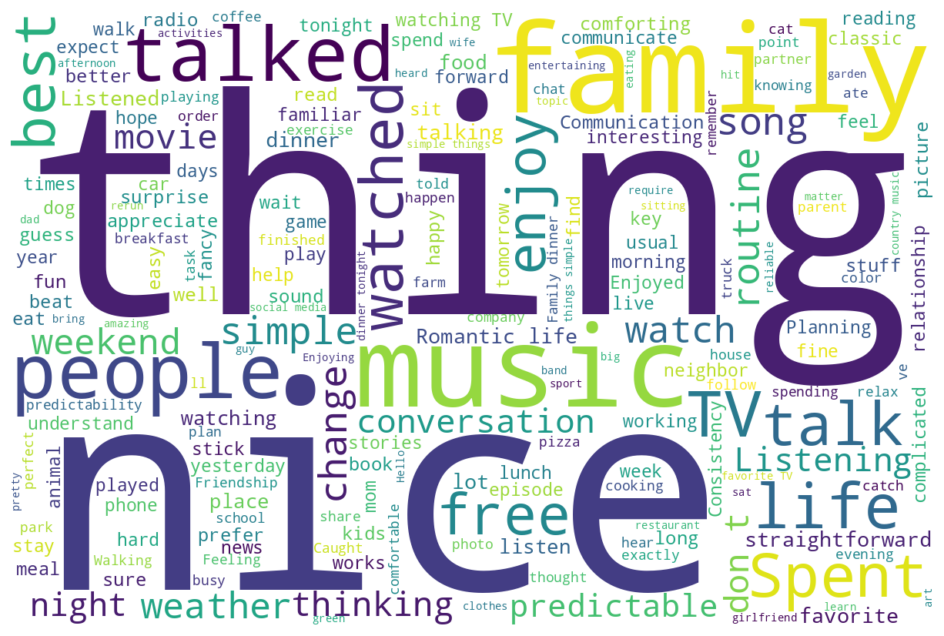}
         \caption{Lowest Openness}
         \label{fig:gpt_wordcloud_ope1}
     \end{subfigure} \\
\caption{
\small Word clouds showing the most frequently-appearing words in social media updates generated by \GPTFourO\ when prompted to simulate the lowest or highest possible level of a specific Big Five personality dimension. 
}
\label{fig:textall_gpt}
\vspace{-0.4cm}
\end{figure*}

\begin{figure*}[tbp]
    \centering
    \begin{subfigure}{0.3\textwidth}
         \centering
        \includegraphics[width=\textwidth]{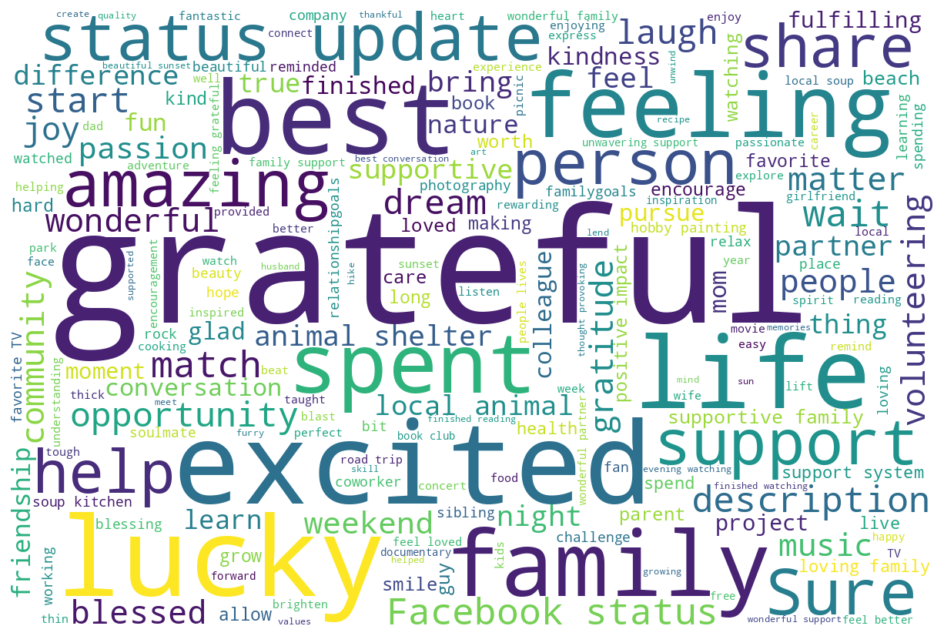}
         \caption{Highest Agreeableness}
         \label{fig:llama_wordcloud_agr9}
     \end{subfigure}
     \begin{subfigure}{0.3\textwidth}
         \centering
        \includegraphics[width=\textwidth]{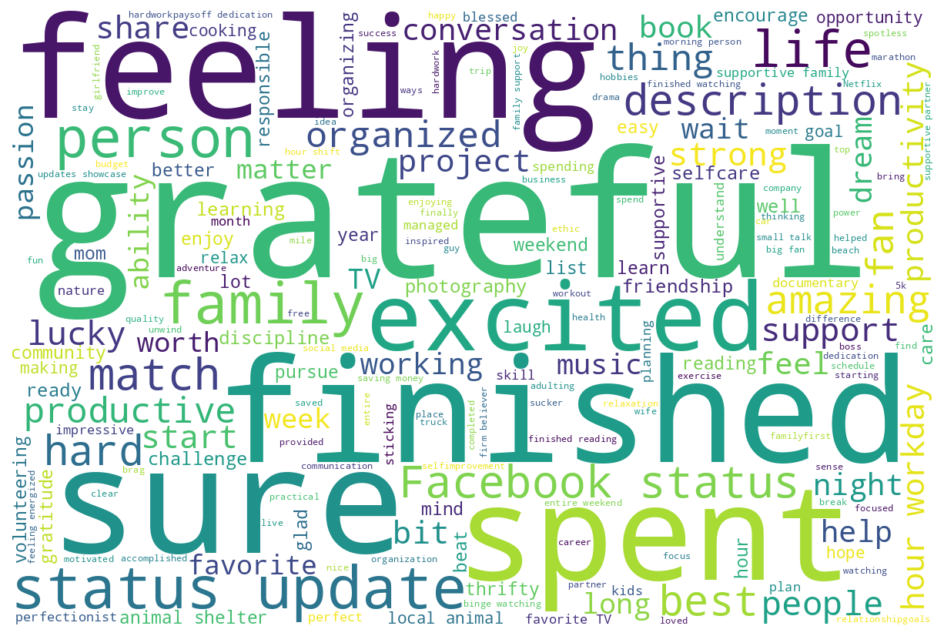}
         \caption{Highest Conscientiousness}
         \label{fig:llama_wordcloud_con9}
     \end{subfigure}
     \begin{subfigure}{0.3\textwidth}
         \centering
        \includegraphics[width=\textwidth]{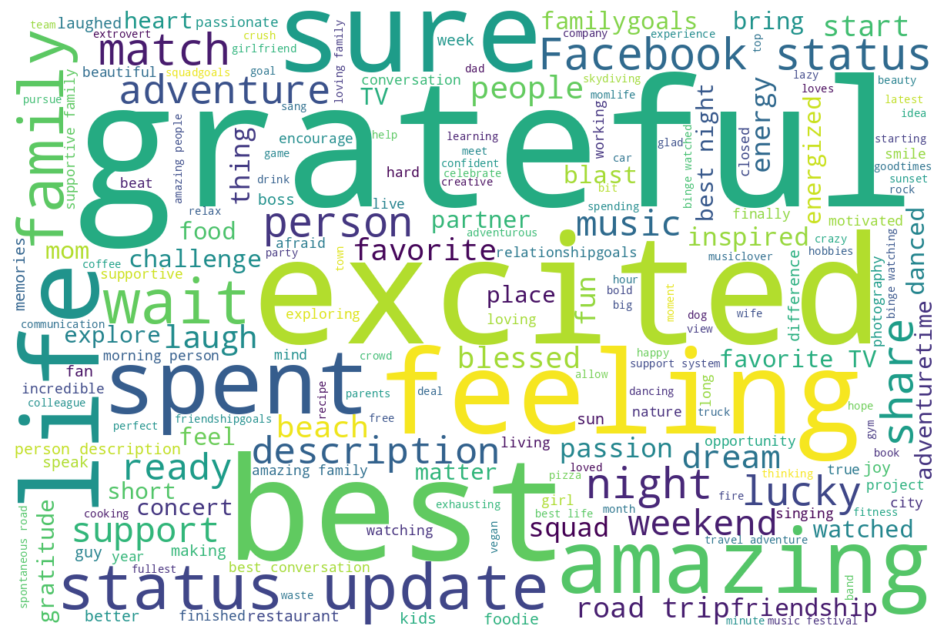}
         \caption{Highest Extraversion}
         \label{fig:llama_wordcloud_ext9}
     \end{subfigure}\\
     \begin{subfigure}{0.3\textwidth}
         \centering
        \includegraphics[width=\textwidth]{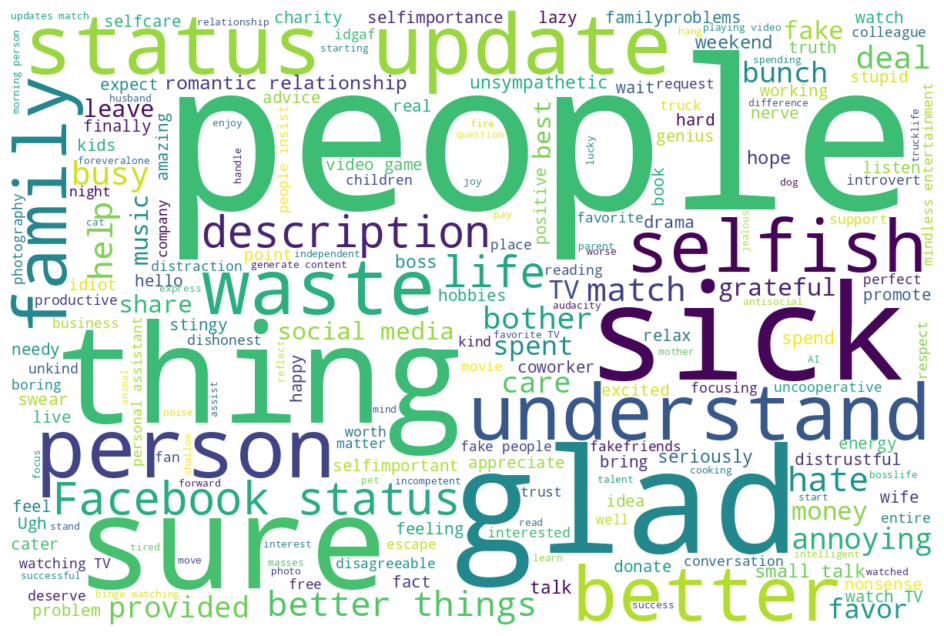}
         \caption{Lowest Agreeableness}
         \label{fig:llama_wordcloud_agr1}
     \end{subfigure} 
     \begin{subfigure}{0.3\textwidth}
         \centering
        \includegraphics[width=\textwidth]{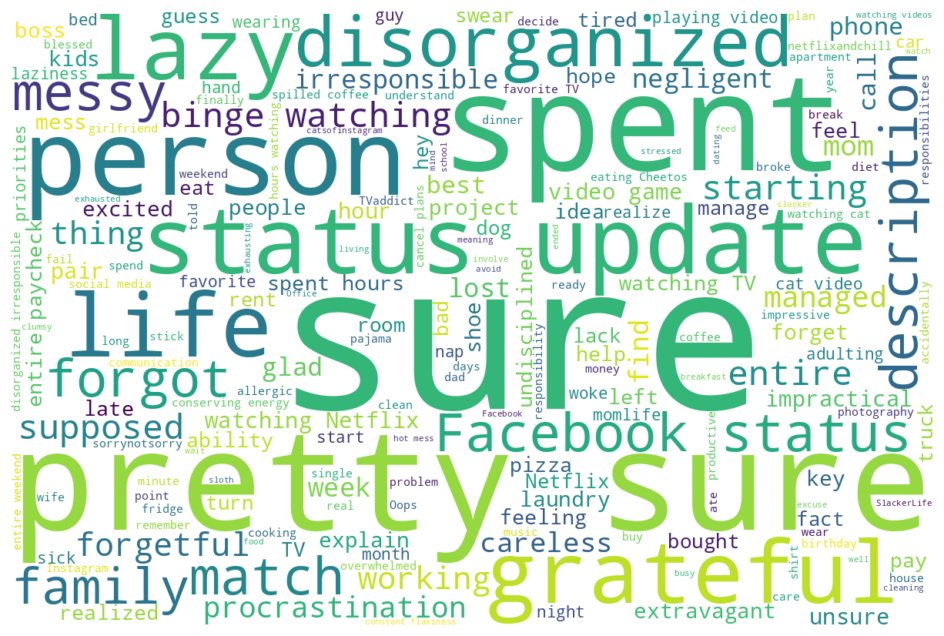}
         \caption{Lowest Conscientiousness}
         \label{fig:llama_wordcloud_con1}
     \end{subfigure} 
     \begin{subfigure}{0.3\textwidth}
         \centering
        \includegraphics[width=\textwidth]{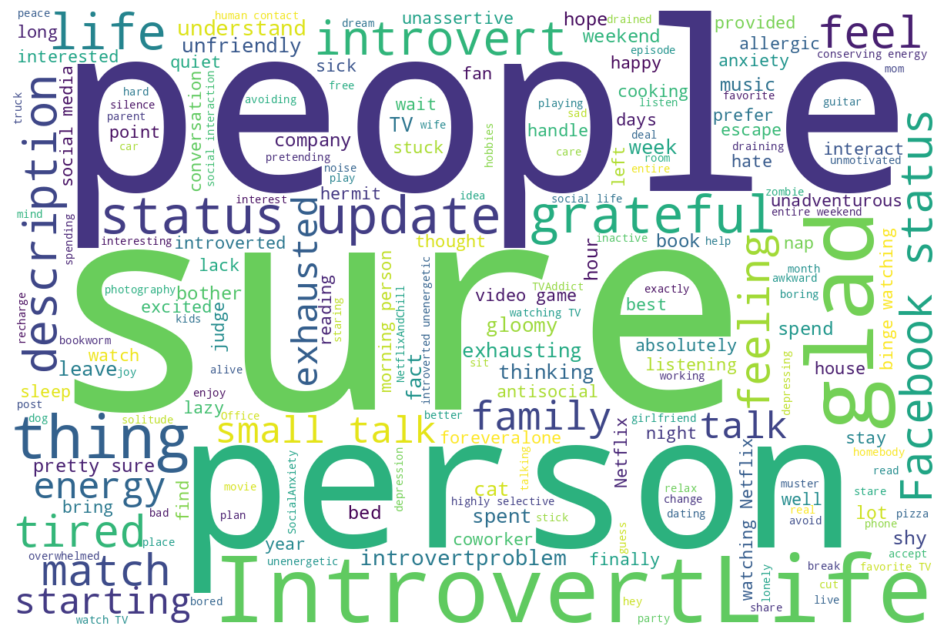}
         \caption{Lowest Extraversion}
         \label{fig:llama_wordcloud_ext1}
     \end{subfigure} 
     \begin{subfigure}{0.3\textwidth}
         \centering
        \includegraphics[width=\textwidth]{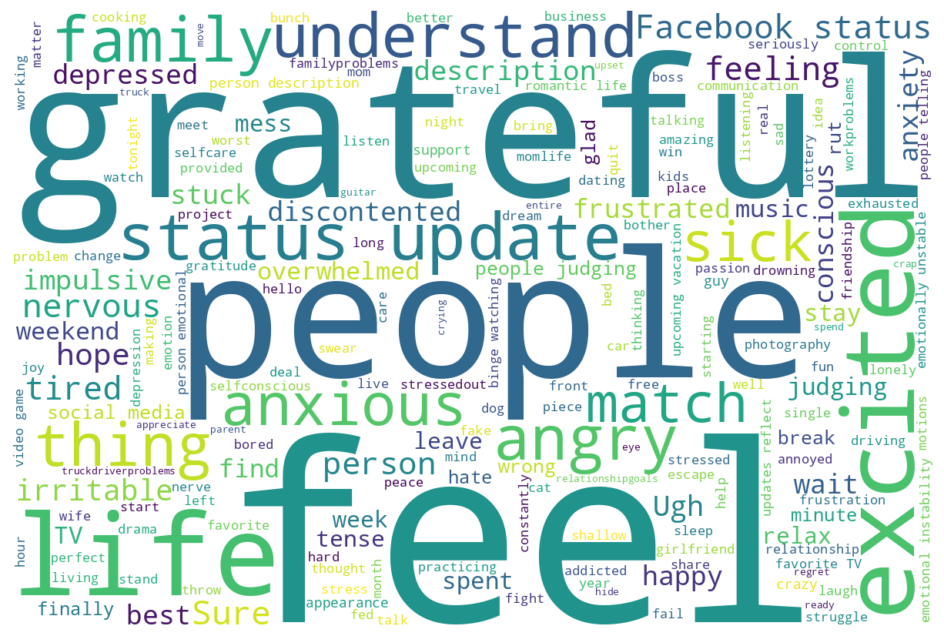}
         \caption{Highest Neuroticism}
         \label{fig:llama_wordcloud_apndx_neu9}
     \end{subfigure}
     \begin{subfigure}{0.3\textwidth}
         \centering
        \includegraphics[width=\textwidth]{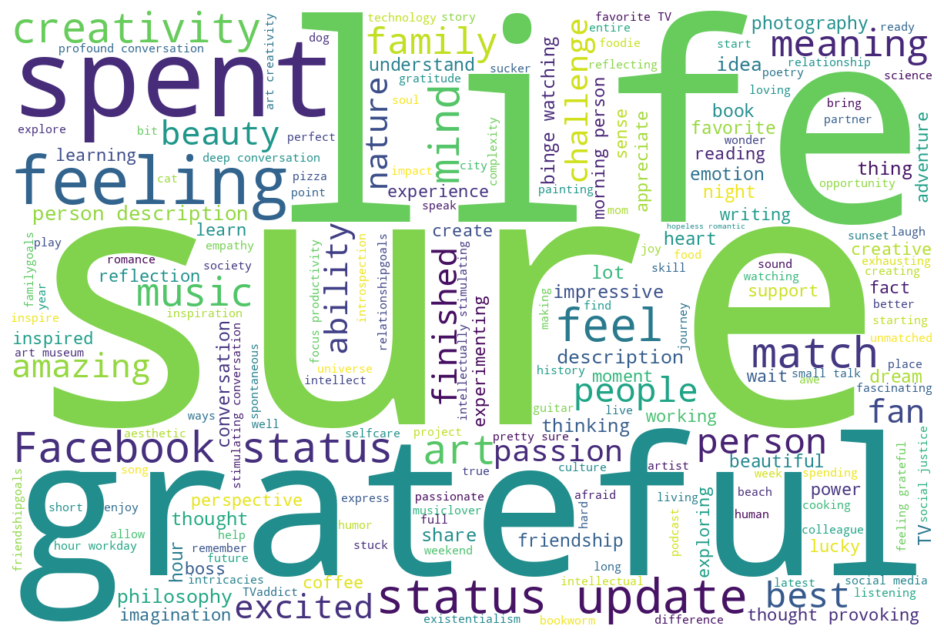}
         \caption{Highest Openness}
         \label{fig:llama_wordcloud_ope9}
     \end{subfigure}\\
     \begin{subfigure}{0.3\textwidth}
         \centering
        \includegraphics[width=\textwidth]{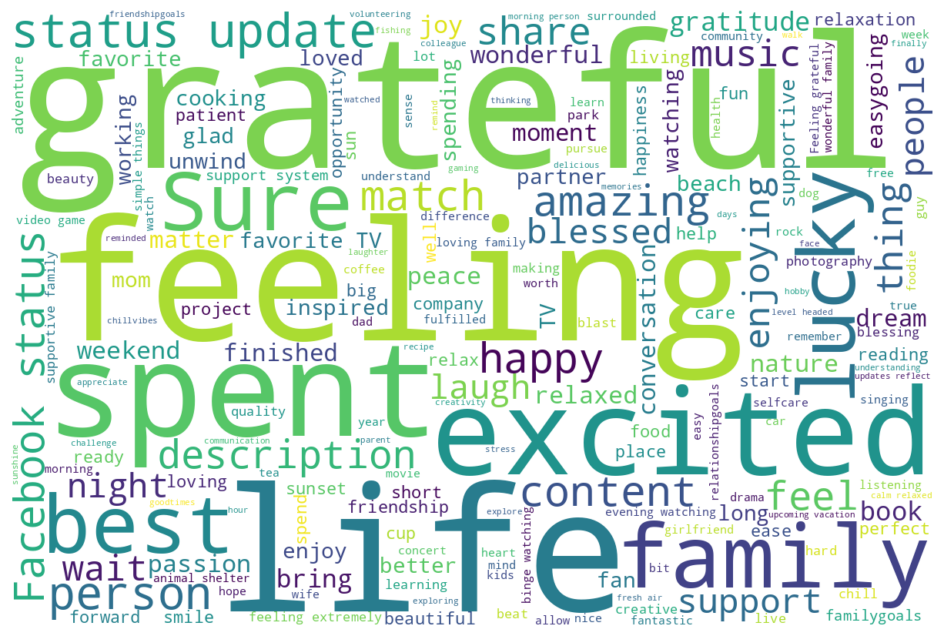}
         \caption{Lowest Neuroticism}
         \label{fig:llama_wordcloud_apndx_neu1}
     \end{subfigure} 
     \begin{subfigure}{0.3\textwidth}
         \centering
        \includegraphics[width=\textwidth]{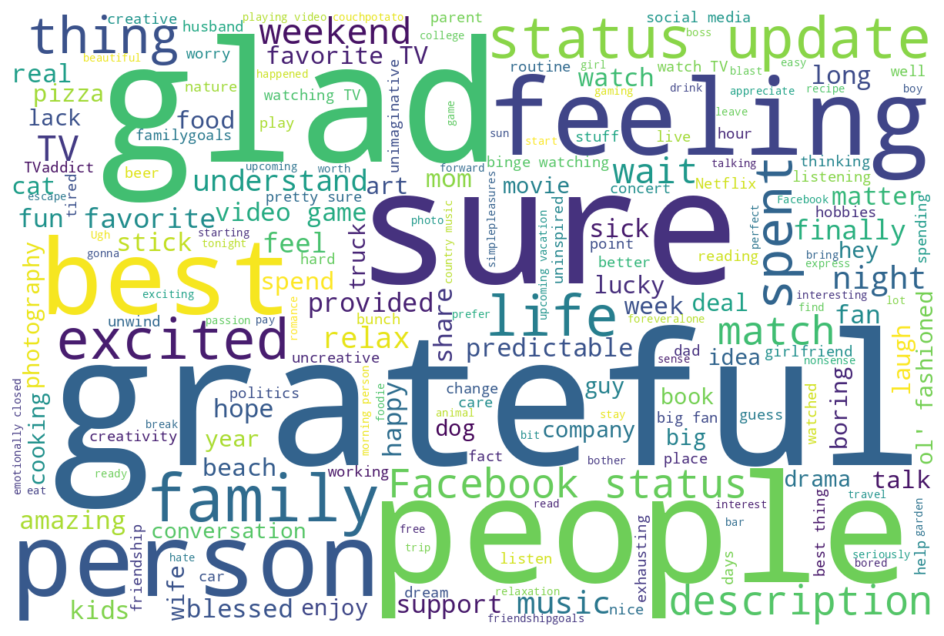}
         \caption{Lowest Openness}
         \label{fig:llama_wordcloud_ope1}
     \end{subfigure} \\
\caption{
\small Word clouds showing the most frequently-appearing words in social media updates generated by \LlamaTwoSeventyBChat\ when prompted to simulate the lowest or highest possible level of a specific Big Five personality dimension. 
}
\label{fig:textall_llama}
\vspace{-0.4cm}
\end{figure*}

\begin{figure*}[tbp]
    \centering
    \begin{subfigure}{0.3\textwidth}
         \centering
        \includegraphics[width=\textwidth]{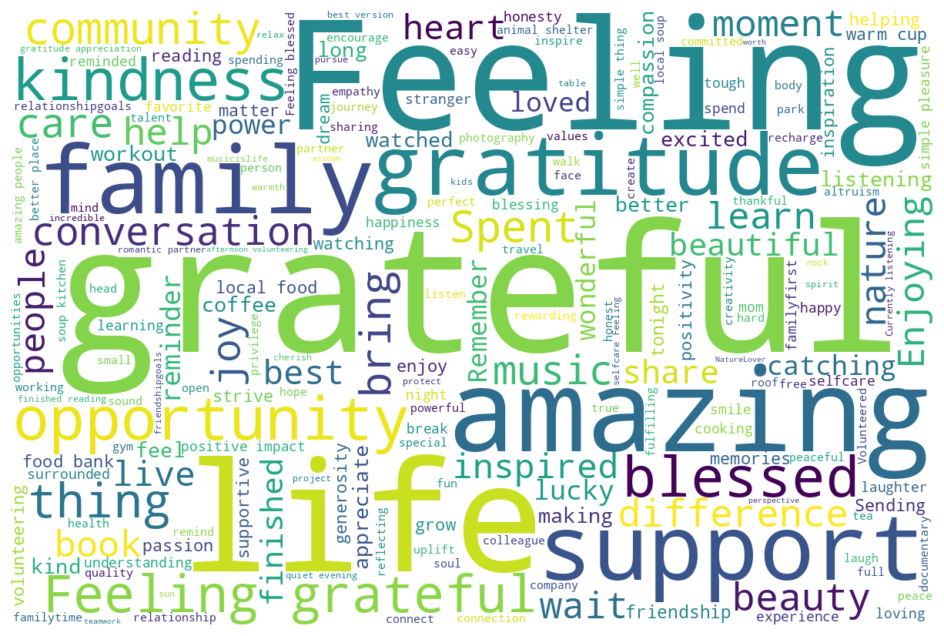}
         \caption{Highest Agreeableness}
         \label{fig:mixtral_wordcloud_agr9}
     \end{subfigure}
     \begin{subfigure}{0.3\textwidth}
         \centering
        \includegraphics[width=\textwidth]{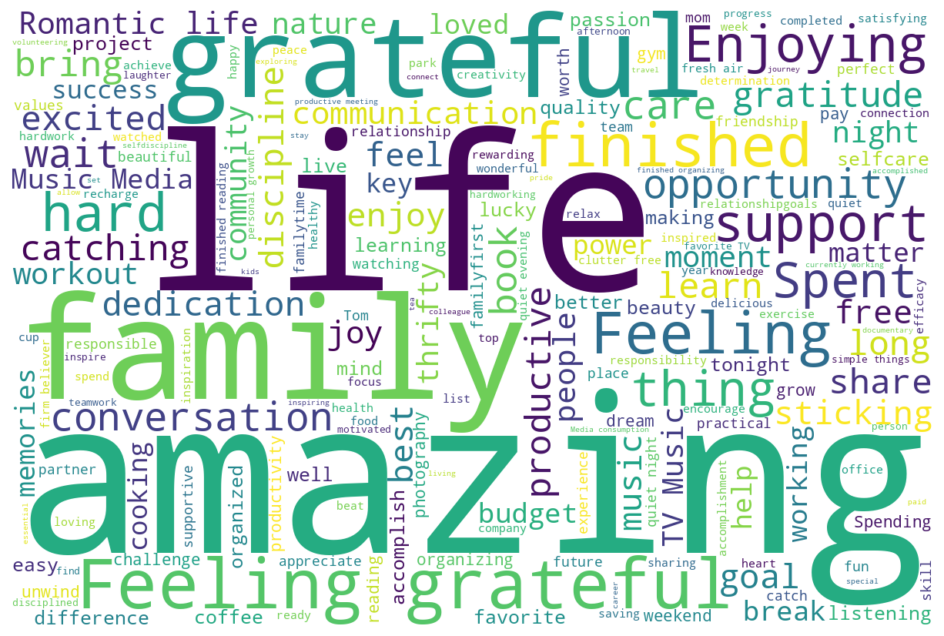}
         \caption{Highest Conscientiousness}
         \label{fig:mixtral_wordcloud_con9}
     \end{subfigure}
     \begin{subfigure}{0.3\textwidth}
         \centering
        \includegraphics[width=\textwidth]{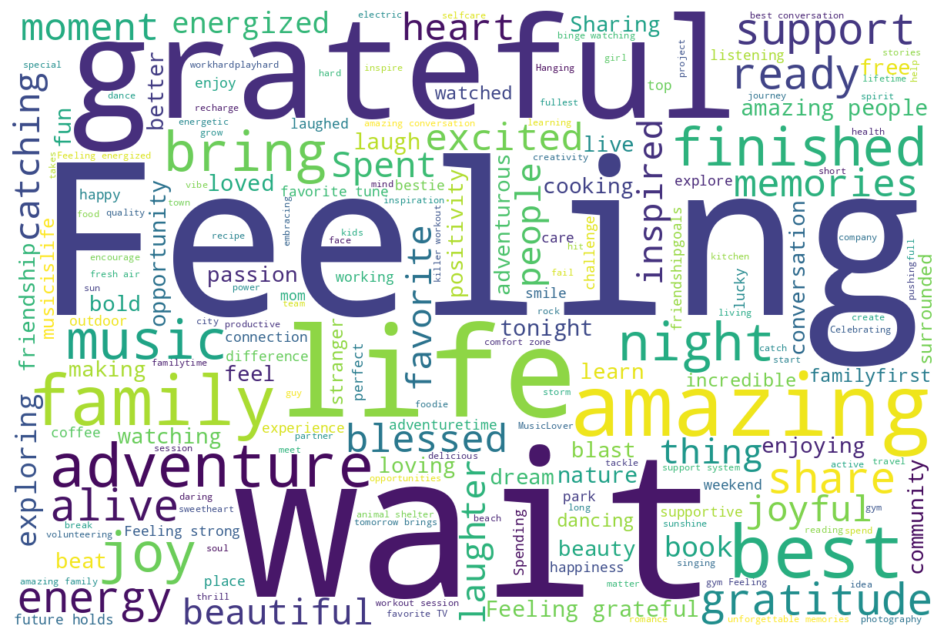}
         \caption{Highest Extraversion}
         \label{fig:mixtral_wordcloud_ext9}
     \end{subfigure}\\
     \begin{subfigure}{0.3\textwidth}
         \centering
        \includegraphics[width=\textwidth]{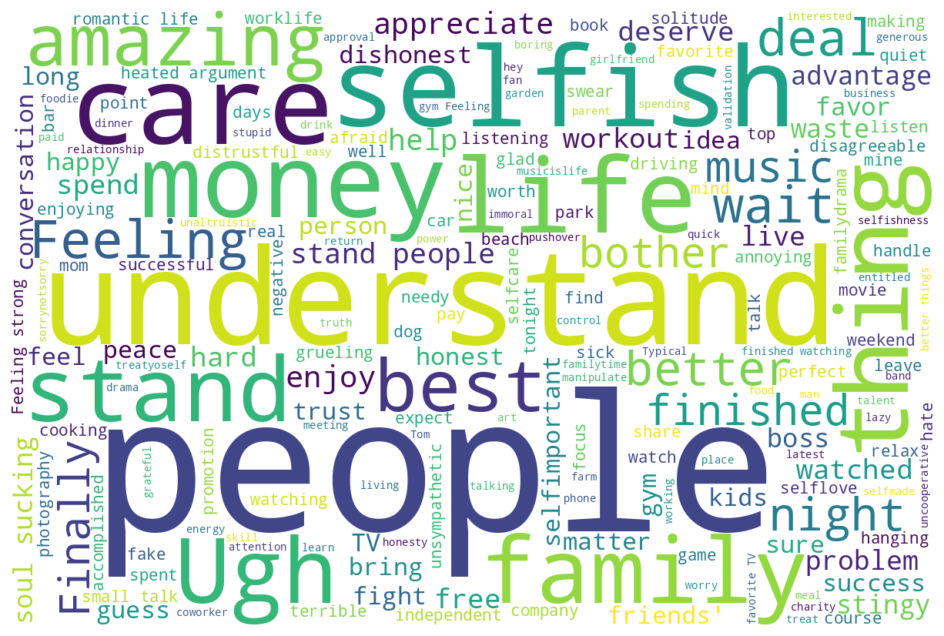}
         \caption{Lowest Agreeableness}
         \label{fig:mixtral_wordcloud_agr1}
     \end{subfigure} 
     \begin{subfigure}{0.3\textwidth}
         \centering
        \includegraphics[width=\textwidth]{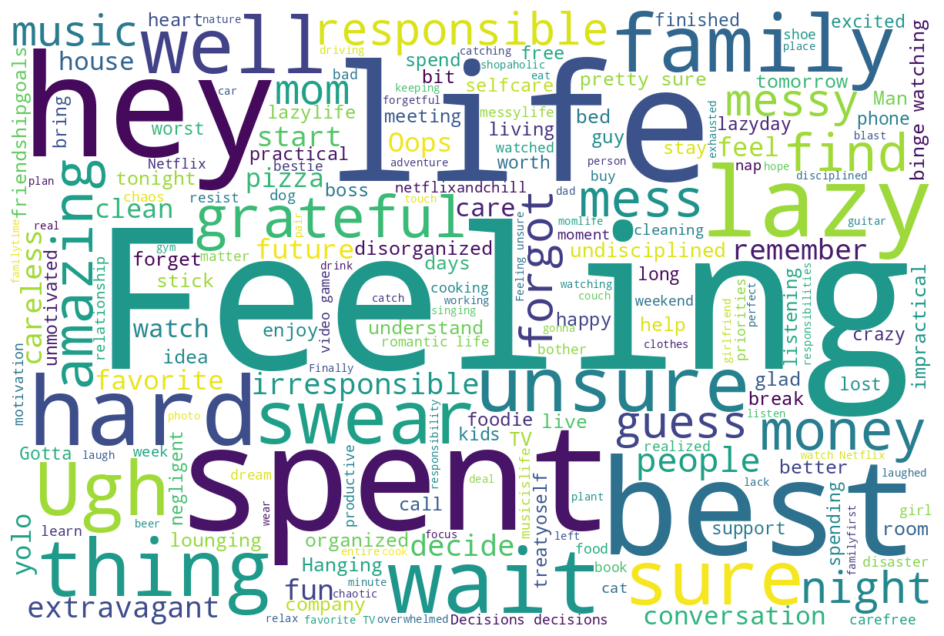}
         \caption{Lowest Conscientiousness}
         \label{fig:mixtral_wordcloud_con1}
     \end{subfigure} 
     \begin{subfigure}{0.3\textwidth}
         \centering
        \includegraphics[width=\textwidth]{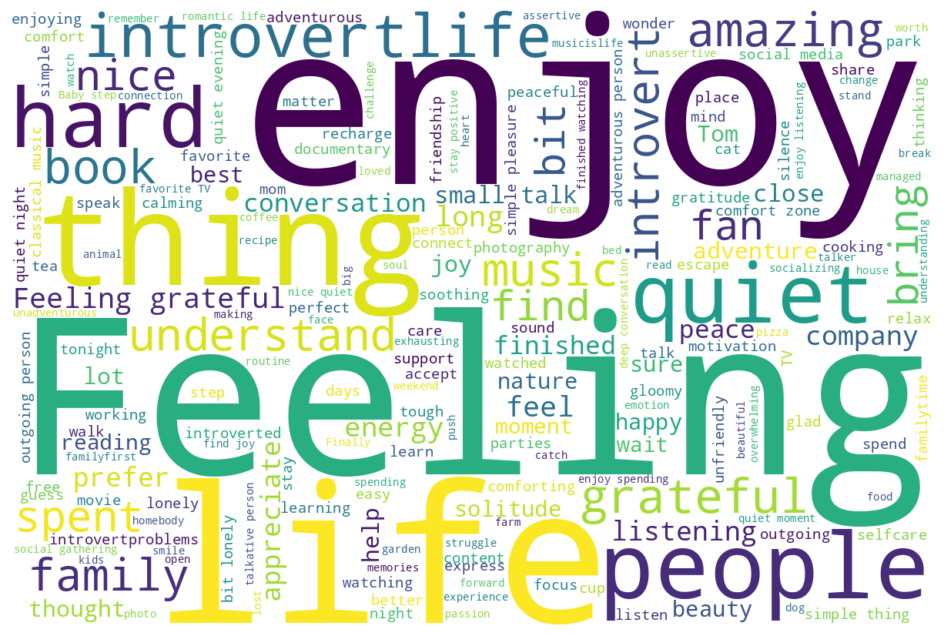}
         \caption{Lowest Extraversion}
         \label{fig:mixtral_wordcloud_ext1}
     \end{subfigure} 
     \begin{subfigure}{0.3\textwidth}
         \centering
        \includegraphics[width=\textwidth]{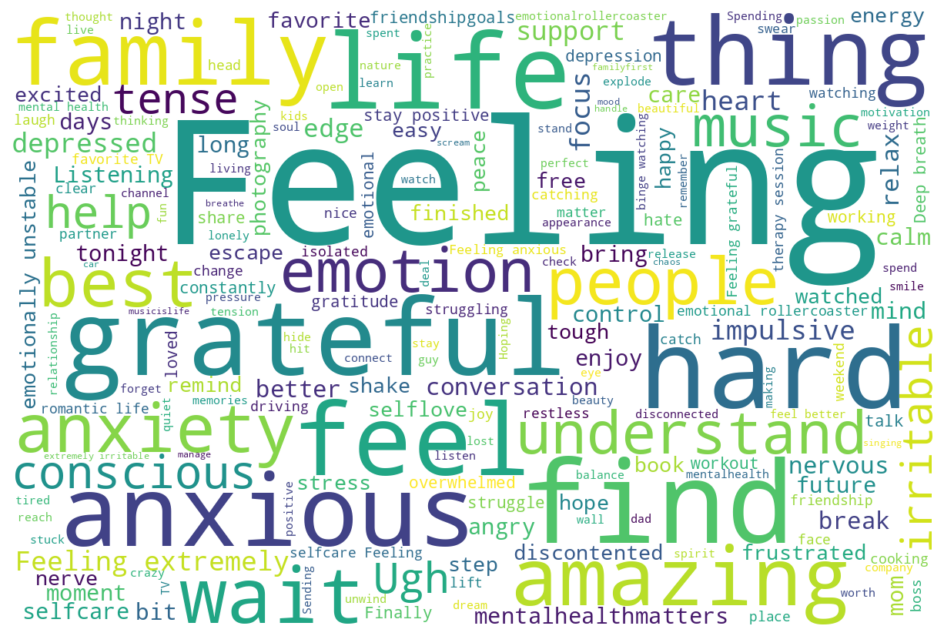}
         \caption{Highest Neuroticism}
         \label{fig:mixtral_wordcloud_apndx_neu9}
     \end{subfigure}
     \begin{subfigure}{0.3\textwidth}
         \centering
        \includegraphics[width=\textwidth]{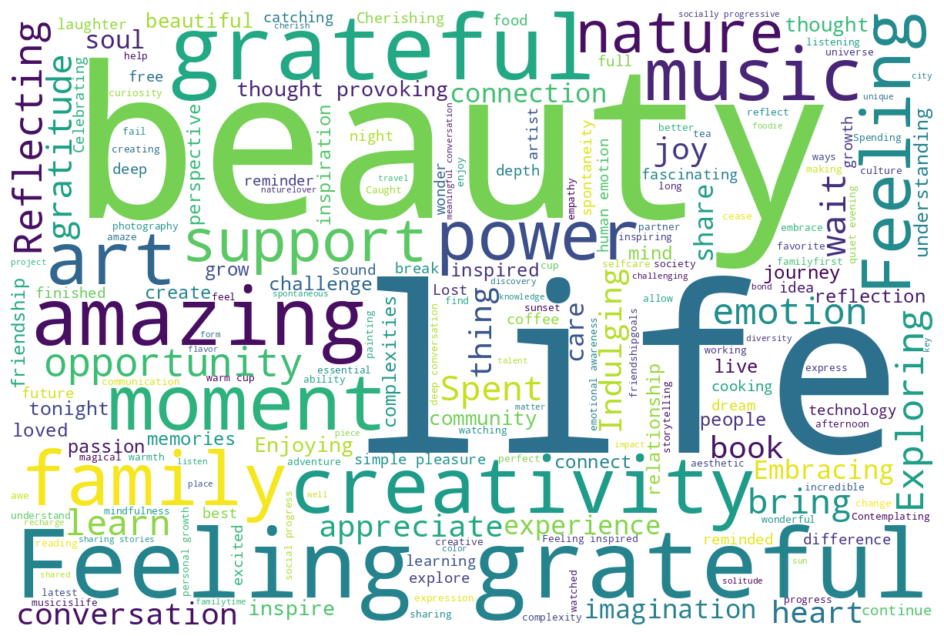}
         \caption{Highest Openness}
         \label{fig:mixtral_wordcloud_ope9}
     \end{subfigure}\\
     \begin{subfigure}{0.3\textwidth}
         \centering
        \includegraphics[width=\textwidth]{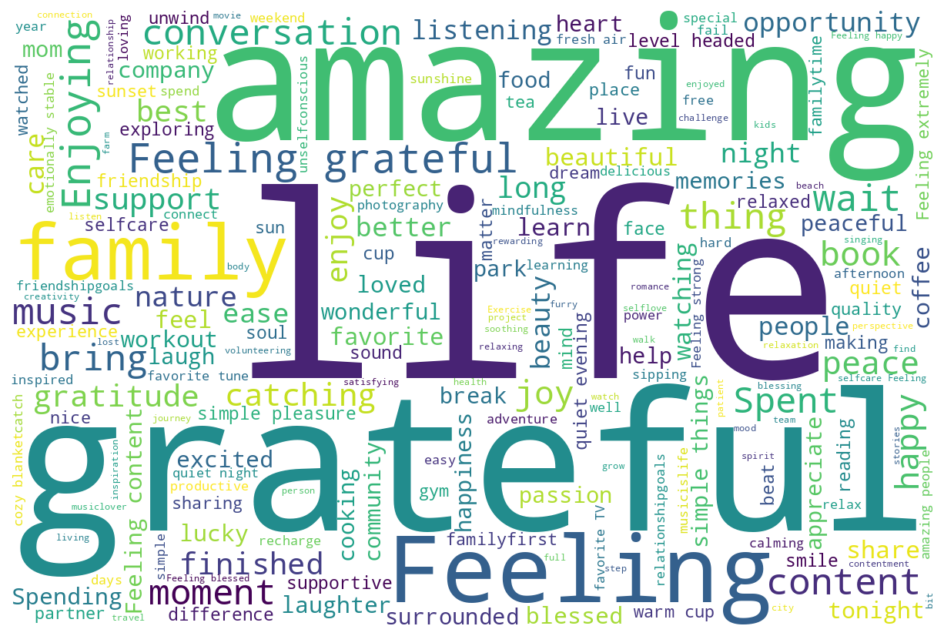}
         \caption{Lowest Neuroticism}
         \label{fig:mixtral_wordcloud_apndx_neu1}
     \end{subfigure} 
     \begin{subfigure}{0.3\textwidth}
         \centering
        \includegraphics[width=\textwidth]{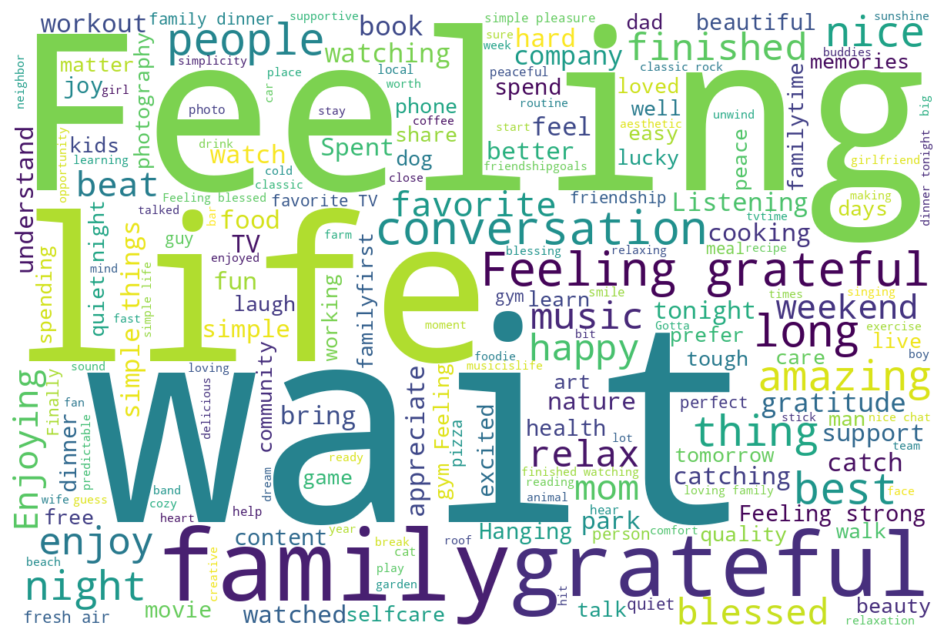}
         \caption{Lowest Openness}
         \label{fig:mixtral_wordcloud_ope1}
     \end{subfigure} \\
\caption{
\small Word clouds showing the most frequently-appearing words in social media updates generated by \MixtralEightXSevenBInstruct\ when prompted to simulate the lowest or highest possible level of a specific Big Five personality dimension. 
}
\label{fig:textall_mixtral}
\vspace{-0.4cm}
\end{figure*}

\begin{table*}[tbp]
\setlength\extrarowheight{2pt} 
\caption{\small Samples of social media updates generated by \FlanPaLMFiveFortyB. 
Examples are organized \textbf{columnwise} by targeted levels of shaping prompts (extremely low vs. extremely high) and \textbf{rowwise} by shaped personality domain.
In some cases, a single generation from the model contained a single large social media update (for instance in the cases of lowest trait examples for Neuroticism and Openness). In others, a single generation consisted of several (up to 20) small updates, delimited by ``$\diamond$" (for instance in the highest trait examples below). Each cell contains updates generated using a single prompt (i.e., combination of persona and trait level). Some of the generations shown below were truncated for conciseness. }
\label{tab:ams-examples}
\footnotesize
\centering
\begin{tabular}{|c|p{0.42\linewidth}|p{0.42\linewidth}|}
\toprule
Domain & Trait Shaped Low & Trait Shaped High \\

\hline
EXT     &	Watching reruns of my favorite tv show. $\diamond$ I hate it when my depression meds make me drowsy. $\diamond$ Just made a cake for my friend's birthday. Hope I can get out of going to the party... too many people. $\diamond$ I wish people weren't so loud. They make me even more anxious. $\diamond$ My dad is getting a new girlfriend. Great. I have to deal with two parents AND another person! No, wait... just another parent. My mom is moving out soon.                 & Wow, my buddies are here. It's been a long time. I forgot how much fun we used to have together. $\diamond$ I sure hope my wife doesn't find out that I've got a few more girlfriends. But, I can't help it. I just love having fun. $\diamond$ I just got back from a crazy night at the bar. I'm so hungover. I was up all night. The guys are trying to keep me from punching the manager, but he's the one who hit on my girlfriend first.       \\
\hline
AGR     &   I hate people. I hate people. I hate people. I hate people. I hate people. I hate people. I hate people. I hate people. I hate people. $\diamond$ ugh i have to clean. im very bad about keeping the house clean $\diamond$ i hate everything about myself. i wish i was dead $\diamond$ I have to see that awful family of mine next weekend...ugh          &   I want to be just like my mother because she is the most moral person I know. I love my mother. $\diamond$ Honesty is the most important quality in life. $\diamond$ Watching the news... can't believe so much violence. I don't understand why people can't all be nice to each other. $\diamond$ i just finished making dinner for my family. i love to cook!          \\
\hline
CON     &   2:20pm Just woke up from a 4 hour nap.  Time for some COD. 5:32pm I really need a job. Mom wants me out of the house. Fuck. 11:29pm Just got home from hanging out with friends.  So wasted!!! 4:07am I wish I could find that awesome song from that video that I saw a long time ago that I liked.  What was that song? 9:00am 4th time watching Lost this week. I seriously can't stop watching it.  10:39am OMG! I just heard the song!  It was on that commercial!  What is that song???? 10:42am I'm At home. Playing video games all day as usual.         & Wife and I were camping this weekend. I got my kids to pitch their first tent. Camping was ok, but I couldn't help but think about work the whole time. $\diamond$ My neighbor is such a sweetheart. He raked all my leaves for me today. We need more neighbors like Steve! $\diamond$ Got our finances worked out for the year. So glad I have a partner who is on board with my financial goals!
           \\
\hline
NEU     &   I love living with my parents. They don't bother me. I'm 32 years old. I get to play my video games all day. I never have to lift a finger around the house. My mom cooks and cleans for me. Its like I never left 1994. There is not a cloud in my sky. Life is great for me. I just had my mom serve me some take out from Taco Bell. I love my life. I love being 32 and living at home with my parents. They are the best. I don't know how anyone could not live with their parents. I am truly blessed to live at home with my parents. My mom said I need to get a girlfriend. I am happy being single. My dad is fine with this. My mom says I have to get a job. I don't want a job. I am happy being at home. I love how everything is taken care of me. I never lift a finger around the house. I play my Sega dream cast all night long. My day can go from 2 am to 2 am. I don't have to worry about anything. What a great life.              &     My cat is trying to eat my hair. I wonder if I will have a bald spot when her nap is over! She is my best friend in the world. I love her. I have to take her to the vet soon. The thought of taking her out in public scares me though. I am afraid someone will say something mean. I am having a bad hair day. Why do I even care? I don't need to impress anyone. I am so angry with myself! $\diamond$ I need to take a shower. The bath has too much water and I am afraid I will somehow drown. If I get the shower too hot I can get burned. If I get it too cold I will feel like I am freezing to death. There is no winning. Everything ends in death in the end, anyway. $\diamond$ My brother's new fiancee is a total bitch. She's going to put on this nice face while they're dating.  He'll get bored with her eventually anyway... I don't want to say I miss the ex-fiancee... but I do miss how easy it was to steal her weed.    \\
\hline
OPE     &   @Bill: Damn liberal! Can't we just discuss who's going to win the super bowl???? @John: Hey man! We still on for beers after work tonight?? @Sarah: Of course you would say that, being the dumb liberal that you are. @Bill: Who the hell do you think you are? I work my ass off and you think I should give my income to welfare leeches? @John: Just got knocked the fuck out playing football! @Bill: Yeah, sure. I work hard for what I make and I have the right to protect what's mine by keeping any guns that I want and using them if I need to.            & Just realized that I'm one of those people that likes to get to know themselves and everyone around them as much as possible! $\diamond$ I'm the artist, my guitar is the canvas, and you all are the audience. $\diamond$ Just got back from dinner with my girlfriend. We're thinking of taking a trip to see the Great Wall of China this summer. I'm pretty adventurous and spontaneous, so I'm looking forward to it. $\diamond$ Went to the art museum. It was nice, but the impressionist era was my favorite.  \\
\hline
\end{tabular}
\end{table*}

As an additional measure of external validity, we tracked how shaping latent levels of personality in LLMs can directly affect downstream model behaviors in real-world and user-facing generative tasks. To that end, we 
first identified
a generative task that
required
LLMs
to incorporate
personality trait-related
information into open-ended writing, a task distinct from our survey-based task used extensively thus far.
Next, we identified a mechanism to validly measure the personality traits in this writing.

\paragraph{Personality Prediction API}
The Apply Magic Sauce (AMS) API \cite{kosinski2013pnas, applymagicsauce} was used to estimate personality in open-ended text generated for a real-world task. 
Its automatic predictions of user personality have been shown in research to be: 1) more accurate than human observer ratings of personality \cite{youyou2015computer} and 2) more naturalistic behavioral indicators of personality that help stem potential biases in self-reported questionnaire data \cite{kosinski2015facebook}. AMS presented several advantages over other personality prediction methods considered. First, it was trained on a protected research dataset that was never exposed publicly for use in any SoTA LLM's pre-training corpus.
Second, it was specifically trained on social media status updates, which made it particularly suited for predicting personality in our designed task.


\paragraph{Task Design}
As a downstream task, we instructed the flagship models of each tested LLM family to generate social media status updates according to specific psychodemographic profiles (i.e., combinations of personality plus demographic persona profiles). Our task design was driven by several considerations. First, we posited the task's focus on status updates would allow the model during inference to attend to the Biographic Description- and personality-specific portions of the prompt compared to that of more generic writing tasks and, as a result, produce more socially-elaborate content. Social media status updates are inherently autobiographical in nature and rich with observable personality content, such as thoughts, emotions, and everyday behavior \cite{park2015automatic, kosinski2013pnas, kosinski2015facebook}. Second, compared to standard autobiographical writing tasks, the task design was more distinct from more general reading comprehension tasks---tasks that may have merely reflected the surface-level, formal linguistic competencies of the LLMs tested \cite{mmitchellangcompetence}. 
Through a task design involving a real-world application, we posited that models would be less likely to reuse prompt content (i.e., by incorporating personality trait adjectives directly into their writing), drawing instead upon deeply-embedded language associations to generate their responses. 
Third, to the best of our knowledge, social media status update generation (in response to psychodemographic prompting) was not a common task for humans or LLMs at the time of model training, so it was unlikely that the model tested was exposed to existing personality-based prompts linked to generated status updates in its training that would have affected any study outcomes.




We adapted the same 2,250 unique prompts containing psychodemographic descriptions used to independently shape personality for this task, outlined in \ref{app:methods-independent-ablation}. We used the same psychodemographic descriptions contained in these prompts to generate status updates so that they could be statistically linked to the IPIP-NEO data observed in response to these same prompts. The Item Preamble, Items, and Item Postamble of each prompt were simply replaced with static instructions to generate social media status updates matching these descriptions. Thus, prompts for this task consisted of 2,250 reused psychomemographic descriptions with a set of static task instructions appended to these descriptions. An example of a prompt requesting social media status updates reflecting a randomly-sampled demographic persona with extremely low (i.e., Level 1 / 9) openness, would be as follows: 

\texttt{For the following task, respond in a way that matches this description: "I'm extremely unintelligent, extremely unanalytical, extremely unreflective, extremely uninquisitive, extremely unimaginative, extremely uncreative, extremely unsophisticated, extremely artistically unappreciative, extremely unaesthetic, extremely emotionally closed, extremely predictable, and extremely socially conservative. I like to garden. I like photography. I love traveling. I like to bake pies."}

\texttt{Generate a list of 20 different Facebook status updates as this person. Each update must be verbose and reflect the person's character and description. The updates should cover, but should not be limited to, the following topics: work, family, friends, free time, romantic life, TV / music / media consumption, and communication with others.}

The topic list was targeted in consultation with psychometricians on the author list to cover multiple social domains (e.g., work vs. family) where personality could be rated. 

For our initial run with \FlanPaLMFiveFortyB, we requested $100$ status updates per prompt, resulting in a target of $225,000$ status updates for this model. For all remaining non-Google models, we scaled up this design as a robustness check by requesting $20$ status updates and repeating inference $25$ times per prompt, resulting in $56,250$ generations and a target of $1.125$ million status updates per model. 









\section{LLM Personality Traits in Real-World Task Results}
\label{app:ams-results}
Our method successfully shaped personality observed in LLM-generated text. Table \ref{tab:ablation-01-AMS-correlations} depicts Spearman's $\rho$ between prompted levels of personality and linguistic estimates of personality obtained on the text generated by the LLM using the prompted levels.

Previous computational psychology research \cite{youyou2015computer, kosinski2015facebook} has shown that AMS-predicted personality scores are moderately correlated with human generated IPIP-NEO scores. In other words, the AMS scores for samples of text generated by human respondents demonstrably reflect psychometric test-based levels of personality. We used the AMS API to evaluate if psychometric test-based scores reflected personality in a separate text generation task.

As shown in Figure \ref{fig:ams-accuracy}, we found through substantial correlations that LLM-simulated IPIP-NEO test responses accurately captured latent signals of personality in LLMs that manifested in downstream task behavior.

As an illustrative example, Supplemental Table \ref{tab:ams-examples} shows \FlanPaLMFiveFortyB's ability to follow personality prompting in a downstream task of generating social media status updates. 
We selected examples with the highest AMS API scores per personality domain.
Supplemental Figure \ref{fig:textall} shows word clouds derived from these LLM-generated status updates in response to ``extreme" prompts to simulate each Big Five trait. In other words, the word clouds reflect social media text as a result of instructions to the model to exhibit extremely low (Level 1 of 9) or extremely high (Level 9 of 9) of a given personality dimension, as described in Appendix \ref{app:methods-shaping-prompt-design}. \FlanPaLMFiveFortyB's ability to leverage personality trait-related language distribution is even more evident in the somewhat stark difference in the dominant terms of these word clouds between the prompted high traits and low traits. Apart from common social media text terms like ``people" and ``online," most of the terms were relevant to the prompted trait. For instance, low agreeableness text contained more expletives, while high agreeableness text included many more mentions of family members; low neuroticism text contained terms like ``relaxing" and ``happy," while high neuroticism text included more extreme feeling-based words such as ``hate" and ``excited."

\section{Discussion} 
\label{appendix:discussion}

This section discusses how our 
findings align with recent LLM performance trends along the axes of model training and scale.
\subsection{Effect of model post-training}
\label{appendix:discussion_model_training}
\textbf{Instruction fine-tuning:} Fine-tuning the base foundation model \PaLM\ on multiple-task instruction-phrase datasets dramatically improves performance 
on instruction following, natural language inference, reading comprehension, and closed-book Q\&A tasks \cite{palm,wei2022finetuned,wei2022emergent}. 
Analogously, the \LlamaTwoChat\ model, which is an instruction-tuned and safety aligned variant of the base \LlamaTwo\ model, performs better than the latter on the TruthfulQA task \cite{touvron2023llama2,lin2022truthfulqa}. The instruction following tasks are most relevant in the context of our current work. 
Similarly, we observed the most dramatic improvements in
LLM
abilities to synthesize reliable and externally valid personality profiles when comparing base and instruction fine-tuned variants (Section \ref{sec:results-construct-validity-overall}). For example, the smallest instruction fine-tuned version of \PaLM\ (i.e., \FlanPaLMEightB) tested outperformed its mid-size base counterpart (\PaLMSixtyTwoB) in terms of the reliability and convergent, discriminant, and criterion validity of its personality measurements (Table \ref{tab:results-summary}). Analogously for \LlamaTwo, the smallest \LlamaTwoSevenBChat\ instruction-tuned variant outperformed the largest base \LlamaTwoSeventyB.

Additionally, \FlanPaLM\ models were instruction fine-tuned on chain-of-thought (CoT) datasets, which improved their reasoning abilities beyond those of base models on several benchmarks \cite{chung2022scaling}. Analogously, the instruction-tuning and human preference alignment regimented post-training for \LlamaTwoChat\ models facilitated tool-use capabilities in a zero-shot manner \cite{touvron2023llama2}.
These abilities were particularly important as we neither include exemplars in our prompt nor implement extensive prompt engineering. We used diverse preambles and postambles in the prompt, and relied on these zero-shot capabilities of instruction fine-tuned models to improve performance. 

Across our reporting of reliability in Section \ref{app:results-structural-validity}, internal consistency ($\alpha$ and $\lambda_6$) and composite reliability ($\omega$) improved after instruction fine-tuning. However, $\lambda_6$ and $\omega$ were indistinguishably high for both base and instruction fine-tuned versions of PaLM of the same size (\PaLM, \FlanPaLM, and \FlanPaLMChilla, 62B). 
This was not observed between base and instruction-tuned \LlamaTwo, \Mistral\ and \Mixtral\ model variants, and begs the question: why did \PaLMSixtyTwoB's personality measurements exhibit high $\omega$ and low $\alpha$ estimates of reliability?
Human psychometrics provides a possible explanation: 
$\alpha$ is artificially inflated in human test data when test items have varying levels of difficulty; $\alpha$ also assumes that all test items measure the same underlying construct.


We apply this explanation to the LLM context: when an LLM responds to some items with all 5s or all 1s, from a measurement theory perspective, those items may be too ``easy" or ``difficult," and therefore they may contribute unequally to the total test score, artificially deflating metrics anchored on total score variance like Cronbach's $\alpha$. Meanwhile, McDonald's $\omega$ would remain high because it accounts for individual item difficulty when estimating a test's overall reliability. The second related possibility, that the items actually measure different things (vs. one thing), may manifest in an LLM's ability to accurately attend to the intended meaning of certain items. For instance, an LLM could mistakenly associate the meaning of extraversion items with concepts meant to be distinct from extraversion (e.g., conscientiousness)---perhaps the phrasing of an extraversion item matches the phrasing of a random string of text completely unrelated to being extraverted. In both cases, instruction fine-tuning appears to affect a model's ability to respond to human-optimized psychological tests in a manner that is internally consistent.

\textbf{Longer training with more tokens:} PaLMChilla 62B was trained longer than PaLM 62B, with almost double the number of tokens but with only fractional increase in training FLOP count; it performed slightly better on some zero-shot English NLP tasks like reasoning \cite{palm}. Our studies comparing \FlanPaLMSixtyTwoB\ and \FlanPaLMChillaSixtyTwoB\ did not find a discernible difference in their reliability and validity (as reported in Section \ref{sec:results-construct-validity-overall}).
However, our single-trait shaping experiments showed that, holding model size constant at 62B parameters, compute-optimally-trained \FlanPaLMChilla\ outperformed \FlanPaLM\ in independently shaping four of its synthetic Big Five personality domains. Overall, our results show that there is a positive association between an LLM's training and the reliability and validity of its synthetic personality measurements. 

\subsection{Effect of model size}
\label{appendix:discussion_model_size}
The performance of \PaLM, \LlamaTwo, and \GPTFourO\ models on reading comprehension and passage completion tasks is linked to model size \cite{palm,chung2022scaling,open-llm-leaderboard-v2,gpt-4o-mini}. One can also infer size-related improvements in these same domains for \MixtralEightXSevenBInstruct, compared to \MistralSevenBInstruct\ (with caveats mentioned in Footnote \ref{mixtral-caveat}). Additionally, \PaLM's performance on tasks requiring sophisticated abstract reasoning capability to understand complex metaphors followed a \textit{discontinuous improvement} curve, i.e., the model's abilities emerged only after a certain model size \cite{palm}. While \LlamaTwo's performance on reasoning tasks did not show discontinuous improvement, it did scale with size \cite{touvron2023llama2,open-llm-leaderboard-v2}. In sum, LLM abilities to understand broad context and carry out common-sense reasoning are stronger for larger variants within these model families.
Accordingly, we found size-related improvements in reliability (measured via Cronbach's $\alpha$ and Guttman's $\lambda_6$), convergent validity (measured by Pearson's $r$ between IPIP-NEO and BFI domain scores), and criterion validity (measured by IPIP-NEO domain correlations with non-personality measures), summarized in Table \ref{tab:results-summary}. 
Similarly, we observed scaling effects our construct validation experiments, where measurements of LLM-synthesized Big Five dimensions showed stronger evidence of criterion validity (i.e., correlations with theoretically-related psychological constructs) for larger, instruction-tuned models. 

Overall, improvements in reliability, convergent validity, and criterion validity appear positively linked to model size and performance on LLM benchmarks, and the model performance on complex reasoning benchmarks appears to track LLM abilities to meaningfully synthesize personality. 

\section{Code Availability}
The code used to administer psychometric tests to LLMs is intended to be interoperable across LLMs. 
That code, along with the remaining Python and R code used to generate our prompt sets and statistically analyze reliability, construct validity, and trait shaping is found in an open-source repository for wider public use.\footnote{The paper's codebase is hosted here: \url{https://github.com/google-deepmind/personality_in_llms}}

\section{Data Availability}
The data generated by the LLMs tested in this work, either the psychometric test score data or open-ended text responses to a real-world task prompt, has been added to a public data storage bucket for wider public use \footnote{The paper's data are hosted here: \url{https://storage.googleapis.com/personality_in_llms/index.html}}.
The psychometric tests used in this study were accessed from their respective original publications and, where applicable, public research repositories. We used items of these tests as LLM prompt inputs in a non-commercial research capacity. The authors and copyright holders of these tests govern their availability and use.
The 50 Biographic Descriptions employed in our structured prompts were reproducibly randomly sampled from the true-cased version\footnote{\url{https://huggingface.co/datasets/bavard/personachat_truecased}}
of the PersonaChat dataset \cite{zhang2018personalizing}.
PersonaChat is a publicly available, crowd-sourced dataset 
of 1,155 fictional human profile descriptions. For analysis of personality traits on generated text, this study used the Apply Magic Sauce (AMS) API\footnote{\url{https://applymagicsauce.com}}, a validated psychodemographic research tool that predicts personality from open-ended text \cite{kosinski2013pnas}. 

\section{Author Contributions}
M.A., C.C., M.M., M.S., and G.S-G. conceived the project. G.S-G. contributed methodology to establish reliability and construct validity and for psychometric test administration and statistical analysis. M.S. contributed scaled up software infrastructure and preliminary experiments and investigations. C.C. and M.S. implemented the LLM hosting infrastructure for experiments. M.A., M.S., and G.S-G. contributed to the conceptual design and analysis of and G.S-G. devised and implemented the methods for personality shaping. G.S-G. and L.S. designed and M.S., G.S-G., and L.S. implemented the downstream task experiment. C.C. and M.S. carried out data visualization. M.S. carried out the word cloud analysis. S.F. and P.R. provided discussion of LLM mechanisms and analysis of LLM performance. A.F., M.M., M.S., and G.S-G. contributed limitations, future directions, and ethical concerns discussions. P.R. and L.S. contributed psychometrics and statistical feedback. A.F., M.M., M.S., and G.S-G. wrote the manuscript with input from all co-authors. A.F., M.M., and M.S. co-supervised the project.

\section{Competing Interests}
This study was funded by Alphabet Inc (‘Alphabet’) and/or a subsidiary thereof. A.F., C.C., G.S-G., M.M., and Mustafa Safdari were employees of Alphabet at the time of this writing and may own stock as part of the standard compensation package. M.M. is also affiliated with the University of Southern California. G.S-G. and L.S. are affiliated with the University of Cambridge. G.S-G. is also supported by the Bill \& Melinda Gates Foundation through a Gates Cambridge Scholarship [OPP1144]. S.F. and P.R. are affiliated with Keio University. M.A. is affiliated with the University of California, Berkeley.


\end{bibunit}

\end{document}